\documentclass[lettersize,journal]{IEEEtran}
\usepackage{amsmath,amsfonts}
\usepackage{algorithmic}
\usepackage{algorithm}
\usepackage{array}
\usepackage{textcomp}
\usepackage{stfloats}
\usepackage{url}
\usepackage{verbatim}
\usepackage{graphicx}
\usepackage{natbib}
\usepackage{caption}        
\usepackage{subcaption}
\usepackage{booktabs}
\usepackage{hyperref}
\usepackage{xcolor}
\usepackage{soul}


\newcommand{\mpc}[1]{\textcolor{black}{#1}}

\begin{document}

\title{Deep Learning for Clouds and Cloud Shadow Segmentation in Methane Satellite and Airborne Imaging Spectroscopy}


\author{Manuel Pérez-Carrasco$^{1,2}$, Maya Nasr$^{3,4}$, Sébastien Roche$^{3,4}$, Chris Chan Miller$^{3,4}$, Zhan Zhang$^{3,4}$, Core Francisco Park$^{5}$, Eleanor Walker$^{4}$, Cecilia Garraffo$^{2}$, Douglas Finkbeiner$^{5}$, Sasha Ayvazov$^{3}$, Jonathan Franklin$^{4}$, Bingkun Luo$^{2}$, Xiong Liu$^{2}$, Ritesh Gautam$^{3}$, and Steven Wofsy$^{4}$}


\markboth{SUBMITTED TO IEEE TRANSACTIONS ON GEOSCIENCE AND REMOTE SENSING}
{Pérez-Carrasco \MakeLowercase{\textit{et al.}}: Deep Learning for Clouds and Cloud Shadow Segmentation in Methane Satellite and Airborne Imaging Spectroscopy}


\maketitle

\footnotetext[1]{Center for Data and AI, Universidad de Concepción, Edmundo Larenas 310, Concepción, Biobio 4030000}
\footnotetext[2]{Center for Astrophysics $|$ Harvard \& Smithsonian, 60 Garden st., Cambridge, MA 02138}
\footnotetext[3]{Environmental Defense Fund, 257 Park Avenue South New York, New York 10010}
\footnotetext[4]{Department of Earth and Planetary Sciences, Harvard University, 26 Oxford st., 
Cambridge, MA 02138}
\footnotetext[5]{Department of Physics, Harvard University, 17 Oxford st., 
Cambridge, MA 02138}

\begin{abstract}
Effective cloud and cloud shadow detection is a critical prerequisite for accurate retrieval of concentrations of atmospheric methane (CH$_4$) or other trace gases in hyperspectral remote sensing. This challenge is especially pertinent for MethaneSAT, a satellite mission launched in March 2024, to fill a significant data gap in terms of resolution, precision and swath between coarse-resolution global mappers and fine-scale point-source imagers of methane, and for its airborne companion mission, MethaneAIR. MethaneSAT delivers hyperspectral data at an intermediate spatial resolution ($\sim 100 \times 400$\, m), whereas MethaneAIR provides even finer resolution ($\sim 25$ m), enabling the development of highly detailed maps of concentrations that enable quantification of both the sources and rates of emissions. In this study, we use machine learning methods to address the cloud and cloud shadow detection problem for sensors with these high spatial resolutions. Cloud and cloud shadows in remote sensing data need to be effectively screened out as they bias methane retrievals in remote sensing imagery and impact the quantification of emissions. We deploy and evaluate conventional techniques—including Iterative Logistic Regression (ILR) and Multilayer Perceptron (MLP)—with advanced deep learning architectures, namely U-Net and a Spectral Channel Attention Network (SCAN) method. Our results show that conventional methods struggle with spatial coherence and boundary definition, affecting the detection of clouds and cloud shadows. Deep learning models substantially improve detection quality: U-Net performs best in preserving spatial structure, while SCAN excels at capturing fine boundary details. Notably, SCAN surpasses U-Net on MethaneSAT data, underscoring the benefits of incorporating spectral attention for satellite-specific features. 
Additionally, we combine the predictions of both U-Net and SCAN through a Convolutional Neural Network (CNN). 
\mpc{This ensemble method achieves the best performance on both MethaneAIR (F1: 78.50±3.08\%) and MethaneSAT (F1: 78.80±1.28\%) datasets, representing improvements of 2\% and 10\% over conventional methods (U-Net: 78.50±3.08\% and 68.56±0.36\% F1, respectively), while maintaining efficient inference (4.1 ms per 1,000 km²).}
This in-depth assessment of various disparate machine learning techniques, as applied to MethaneSAT and MethaneAIR imaging spectroscopic data at varying spatial resolutions, demonstrates the strengths and effectiveness of advanced deep learning architectures in providing robust, scalable solutions for clouds and cloud shadow screening towards enhancing methane emission quantification capacity of existing and next-generation hyperspectral missions.

Our data and code is publicly available at:  \url{https://doi.org/10.7910/DVN/IKLZOJ}

\end{abstract}


\thanks{This work has been submitted to the IEEE for possible publication. Copyright may be transferred without notice, after which this version may no longer be accessible.}

\section{Introduction}
\IEEEPARstart{R}{emote}  sensing has rapidly evolved as a key tool for quantifying emissions of critically-important greenhouse gases, including carbon dioxide (CO$_2$) and methane (CH$_4$). Among these, methane is especially critical: although it has a relatively short atmospheric lifetime of about twelve years, CH$_4$ exhibits over 80 times the warming potential of CO$_2$ during the first two decades after emission \cite{myhre_2013, Etminan_2016}. This makes methane an attractive target for near-term climate mitigation efforts \cite{Shindell_2012}.

Methane mitigation efforts have advanced in the past few years with the Global Methane Pledge \cite{weforum2024methane} signed by over 150 countries aiming to reduce anthropogenic methane emissions by 30\% by year 2030. Sectorally, the Oil and Gas Decarbonization Charter (OGDC) has pledged to reduce methane emissions intensity down to 0.2\% of production by 2030, which includes over 50 oil and gas companies accounting for more than 40\% of global production\cite{ogdc_2023}. It is important to track the performance of critical methane mitigation targets, and remote sensing has emerged as an effective tool for globally monitoring methane emissions at scale.

Two primary remote sensing strategies have emerged for monitoring methane \cite{Chauhan_2025, Mehrdad_2025}: high-resolution imaging spectrometers, such as AVIRIS-NG, GHGSat, \mpc{PRISMA, EMIT, EnMAP}, Carbon Mapper, and Sentinel-2, which excel at identifying point sources \cite{Frankenberg_2016, Varon_2021, Guanter_2021, Zhang_2022, Roger_2025, Xu_2025, Smith_2025, Bue_2025, Mancoridis_2025}, and low-resolution global mappers like TROPOMI and GOSAT, which offer daily or near-real-time regional coverage \cite{Veefkind_2012, Watine-Guiu_2023}. These hyperspectral missions have provided important data on global scale atmospheric methane concentrations and information on high-emitting point sources in targeted domains. However, there exists an observing and data gap at the scales of individual areas where emission quantification information is not readily available for assessing emissions at high spatial resolution, high precision and wider coverage -- limiting our overall understanding of methane emission sources and magnitude.

\begin{figure*}[ht!]
  \centering
  \includegraphics[width=1\linewidth]{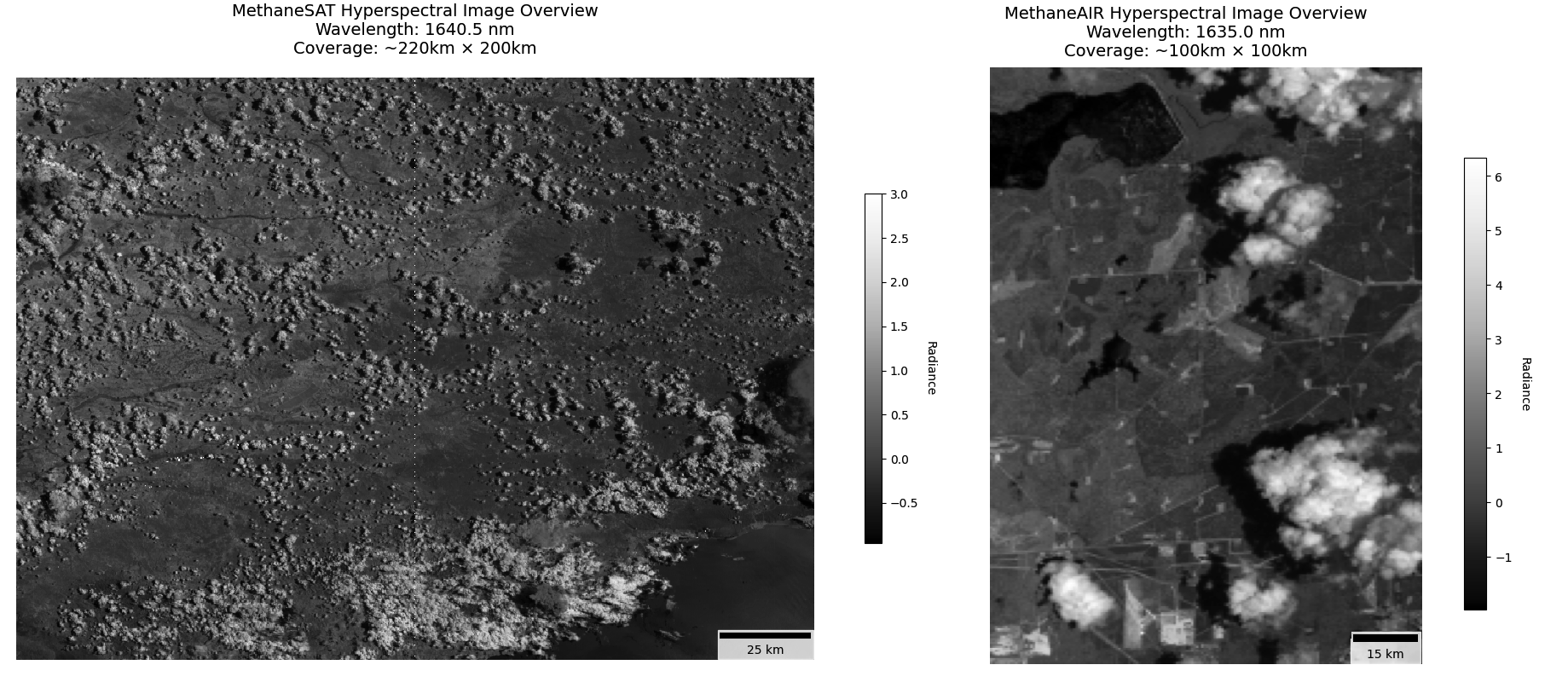}
  \caption{Hyperspectral methane observations from MethaneSAT and MethaneAIR platforms showing typical mapping coverage and spatial extent. \textbf{Left}: MethaneSAT image at 1640.5 nm wavelength, captured on September 4 of 2024, covering approximately 220 km × 200 km area. \textbf{Right}: MethaneAIR image at 1635.0 nm wavelength, captured on June 2 of 2023, covering approximately 100 km × 100 km area.}
  \label{fig:msat_mair}
\end{figure*}

The MethaneSAT mission \cite{Rohrschneider}, launched in March 2024, directly addresses this gap by combining fine-scale resolution, high-precision measurements with broad swath coverage of 220 km at nadir (extendable to over 400 km swath at greater off-nadir viewing geometries). The satellite features two imaging spectrometers: one targeting CH$_4$ (1589–1686 nm) and another measuring O$_2$ (1249–1305 nm), enabling precise retrieval of methane concentrations over oil and gas production basins and agricultural regions distributed globally. MethaneAIR, its airborne companion \cite{Conway_2024}, provides critical data for algorithm development and instrument validation, as well as ongoing mapping of methane emissions. Figure \ref{fig:msat_mair} illustrates the contrasting spatial coverage capabilities of MethaneSAT and MethaneAIR.

A key obstacle in retrieving accurate methane concentrations from hyperspectral imagery is the presence of clouds and cloud shadows, which introduce significant artifacts. When a pixel includes a cloud, surface reflectance is occluded, leading to signal loss that is required for retrieving methane concentrations in cloud-free column atmosphere. In contrast, cloud shadows obscure the solar illumination path, making the optical path length ambiguous. These artifacts impact methane retrievals differently across wavelengths due to variable scattering and absorption in the shortwave infrared (SWIR), resulting in class-specific spectral distortions. As a result, both the spectral and spatial domains encode rich, complementary information about the presence of clouds and shadows—making this problem well-suited to machine learning approaches that can leverage high-dimensional contextual features. 

\mpc{Deep learning has driven substantial progress in cloud and shadow detection for satellite imagery, evolving from pixel-wise classifiers using multilayer perceptrons and random forests \cite{gomezchova_2017, Taravat_2015} to fully convolutional architectures that exploit spatial context. U-Net \cite{Ronneberger2015} became a dominant backbone due to its encoder-decoder structure with skip connections, with Jeppesen et al. \cite{JEPPESEN_2019} demonstrating robust cloud masking across diverse Landsat scenes and Wieland et al. \cite{Wieland_2019} showing strong generalization across Landsat-8 and Sentinel-2. More recently, attention mechanisms have been explored to capture spectral relationships in high-dimensional data \cite{Sun_2020, wright_2024, Li_2022}, while semantic segmentation architectures have been widely adopted for related tasks including land cover mapping and change detection \cite{MA_2019, Chen_2021, Chen_2025}. Despite these advances, most existing methods target broadband multispectral sensors, leaving their applicability to imaging spectrometers with hundreds of contiguous channels largely unexplored. Furthermore, cloud \textit{shadow} detection remains comparatively understudied, as shadows exhibit subtler spectral signatures that are difficult to distinguish from dark surfaces.}

In this work, we investigate a range of machine learning algorithms for the detection of clouds and shadows in MethaneSAT and MethaneAIR hyperspectral imagery. We evaluate conventional models such as logistic regression \cite{Edisanter_2016, park_2023, Gautam_2023} and multilayer perceptrons (MLPs) \cite{BENEDIKTSSON_1993, Yang_1999, Tian_1999, Taravat_2015, gomezchova_2017}, as well as deep learning approaches including U-Net \cite{Tushar_2024, Ronneberger2015, Jiao_2020, Wieland_2019, Miao_2022, wright_2024, Li_2022, Tan_2023} and Spectral Channel Attention Network (SCAN), the latter designed to apply channel-wise attention mechanisms for hyperspectral band selection. Our results demonstrate that while conventional models offer computational efficiency, they struggle to resolve spatially complex or spectrally ambiguous regions. Deep networks, on the other hand, yield higher accuracy—U-Net excels in spatial consistency, while SCAN improves spectral boundary delineation, particularly for shadow detection.

To further improve performance, we develop ensemble architectures that fuse predictions from U-Net and SCAN. We explore both MLP- and CNN-based fusion strategies, with the latter achieving up to 10\% improvements in F1 score over baseline models. These ensembles effectively combine spatial coherence and spectral attention to outperform individual models, especially in regions with complex topography or partial cloud cover.

\mpc{In summary, this study presents the first benchmark of machine learning methods for cloud and shadow segmentation in MethaneSAT and MethaneAIR hyperspectral imagery. We develop SCAN (Spectral Channel Attention Network), a novel architecture employing channel-wise attention mechanisms for hyperspectral band selection, and a CNN-based ensemble that fuses spatial and spectral information to achieve the best overall performance. The motivation of this work is to enable operational efficiency with high accuracy: because our models operate directly on Level-1 radiance data, they can identify and filter cloud and shadow-contaminated pixels prior to Level-2 retrievals, bypassing the computationally expensive full retrieval pipeline for these pixels. This has significant implications for MethaneSAT's
mission to progress towards global methane reduction and is readily transferable to other hyperspectral platforms facing similar preprocessing bottlenecks. We release publicly annotated datasets for both MethaneSAT (262 scenes) and MethaneAIR (508 scenes), along with reproducible code for all evaluated models.}

\section{Background}



Hyperspectral semantic segmentation (HSS), also known as pixel-wise classification, aims to assign distinct labels to each pixel based on their spatial and spectral characteristics \cite{Ahmad_2021, Grewal_2022, Kumar_2024}. The rich spectral information contained in hyperspectral data, spanning multiple contiguous narrow bands, enables precise discrimination of materials and atmospheric features that may be indistinguishable in conventional imaging.

Conventional approaches for HSS primarily relied on spectral-based classification, where each pixel is classified independently based on its spectral signature \cite{PAOLETTI2019279, BENEDIKTSSON_1993, Yang_1999, park_2023}. Methods like logistic regression \cite{Li_2010, park_2023, Gautam_2023} and multilayer perceptrons (MLPs) \cite{BENEDIKTSSON_1993, Yang_1999, Tian_1999, Taravat_2015, gomezchova_2017} operate directly on the spectral vectors, learning the complex relationships between spectral bands and target classes. While these spectral-only approaches can effectively leverage the rich spectral information, they fail to capture important spatial context and patterns that could significantly improve segmentation accuracy.

To address this limitation, modern deep learning architectures have been developed to jointly exploit both spectral and spatial information \cite{PAOLETTI2019279, munishama_2015, Hu_2015}. The U-Net \cite{Ronneberger2015, Tushar_2024} architecture, originally designed for biomedical image segmentation \cite{Ronneberger2015}, has been adapted for multispectral cloud and shadow segmentation using Landsat 8 \cite{Jiao_2020}, Sentinel-2 \cite{Wieland_2019, Miao_2022, wright_2024}, MODIS \cite{Li_2022}, and Gaofen-1 \cite{Tan_2023} multisperspectral data. The architecture incorporates 2D convolutions that process both spatial neighborhoods and spectral bands simultaneously, while its encoder-decoder structure with skip connections enables multi-scale feature capture while preserving fine spatial details, making it particularly effective for semantic segmentation tasks.

More recently, Vision Transformer-based architectures have emerged as a powerful alternative for HSS analysis \cite{Dosovitskiy_2020}. These models employ self-attention mechanisms to capture long-range dependencies in both spatial and spectral dimensions. Transformers have been also applied to cloud detection using Landsat 8 \cite{Li_2024}, Sentinel-2 \cite{Zhang_2023b, Yan_2023}, MODIS \cite{Ge_2024}, and Gaofen-1 \cite{Tan_2023} multispectral images. Although Vision Transformers have shown promising results in various hyperspectral applications \cite{Hong_2021}, their computational complexity and large data requirements can be challenging for operational scenarios \cite{ahmad_2025}.

Atmospheric monitoring of greenhouse gases using hyperspectral remote sensing has significantly advanced with missions such as AVIRIS-NG, EMIT, GHGSat, Carbon Mapper, TROPOMI, and MethaneSAT. These instruments detect and quantify greenhouse gases such as CH$_4$ by leveraging spectral absorption features, primarily in the 1.6 micron and 2.3 micron methane absorption bands. 
Conventional hyperspectral imaging techniques, including matched filtering \cite{fuhrmann1992}, band-pass filtering  and differential optical absorption spectroscopy (DOAS; \cite{Platt_2008}), rely on spectral signatures to enhance detection (e.g. CH$_4$ plumes against background variability \cite{Thompson_2015, Foote_2020, Ehret_2022, kumar_2023}). Therefore, these methods often require expert-driven parameter tuning and manual inspection to reduce false positives \cite{Frankenberg_2016, ruzicka_2023}. 

Cloud and shadow segmentation in hyperspectral imagery presents a different set of challenges from CH$_4$ detection, which benefits from well-characterized spectral absorption features. These atmospheric features exhibit high spectral variability across illumination conditions, surface reflectance, and sensor factors, complicating detection efforts \cite{zhu_2015, Wieland_2019, Chai_2019}. Cloud shadow detection is particularly difficult because dark surfaces like water bodies resemble shadows spectrally, shadowed areas contain diverse land cover types with broad spectral ranges, and thin clouds with high transmittance create shadows that mix with clear pixel characteristics \cite{zhai_2018}. Consequently, conventional spectral-based classification methods struggle to generalize across diverse scenes \cite{Wang_2024}. Influence of cloud shadows in the measured top-of-atmosphere radiances can lead to significant biases in atmospehric methane retrievals in turn affecting emission quantification, and therefore correcting for such artifacts is essential for accurately characterizing emission patterns.

\section{Data}

\begin{figure*}[t]
    \centering
    \begin{minipage}[t]{0.48\textwidth}
        \centering
        \includegraphics[width=\textwidth]{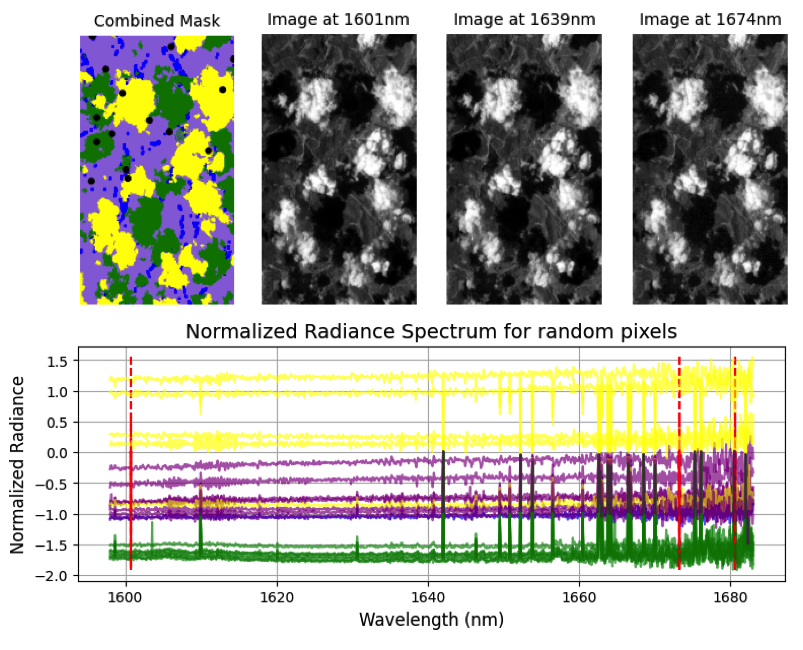}
        \caption{MethaneAIR data example showing classification mask and spectral analysis. The top panel shows a classification mask (purple: background, yellow: clouds, green: shadows, blue: dark surfaces) alongside three images from randomly selected spectral wavelengths. The bottom panel displays normalized radiance spectra from 10 randomly sampled spatial soundings per each class, with colors corresponding to their classification in the mask above. Spectral normalization was performed by computing mean and standard deviation for each spectral band across the entire dataset, then standardizing each spectrum by subtracting the mean and dividing by the standard deviation. This image was captured on September 5, 2023.}
        \label{fig:methaneair_data}
    \end{minipage}
    \hfill
    \begin{minipage}[t]{0.48\textwidth}
        \centering
        \includegraphics[width=\textwidth]{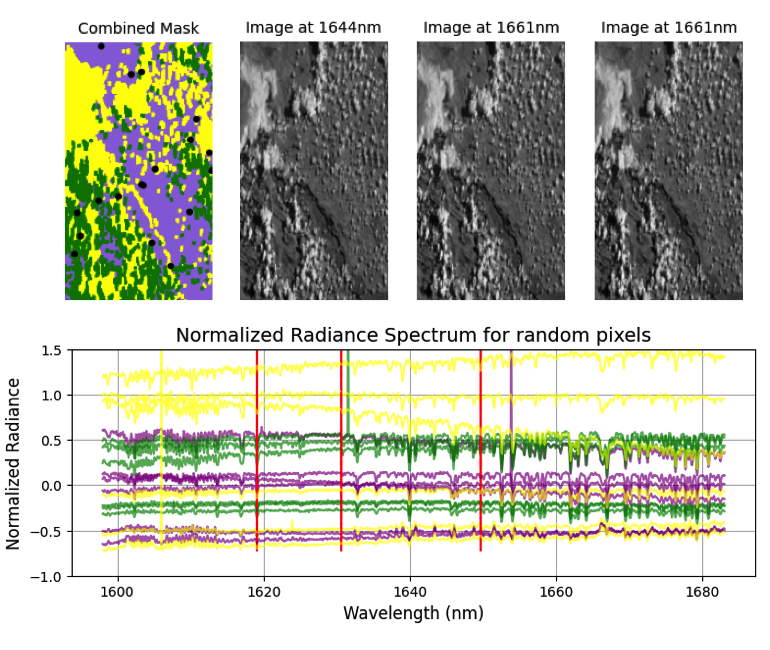}
        \caption{MethaneSAT data example showing classification mask and spectral analysis. The top panel presents the classification mask using the same color scheme as MethaneAIR (purple: background, yellow: clouds, green: shadows) and three representative spectral band images. The bottom panel shows the spectral signatures of 10 randomly selected spatial soundings for each class, with colors indicating their classification status. Spectral normalization was performed by computing mean and standard deviation for each spectral band across the entire dataset, then standardizing each spectrum by subtracting the mean and dividing by the standard deviation. This image was captured on November 18, 2024.}
        \label{fig:methanesat_data}
    \end{minipage}
\end{figure*}

This study utilizes calibrated and georeferenced Level 1B (L1B) hyperspectral data from MethaneAIR and MethaneSAT \cite{Conway_2024}, with cloud labels derived from Level2 (L2) derived quantities\footnote{To download the data and code  visit \url{https://doi.org/10.7910/DVN/IKLZOJ}} \cite{ccm2024methaneair}. The L2 products consist of the retrieved fitted parameters obtained from the best fit of a radiative transfer model to the measured L1B spectra. The main products from the CH$_{4}$ spectrometer are the retrieved CO$_{2}$ and CH$_{4}$ vertical column densities (VCDs), and the main product from the O$_{2}$ spectrometer is the retrieved surface pressure.

MethaneSAT is a satellite mission developed by MethaneSAT LLC, which is a wholly-owned subsidiary of Environmental Defense Fund. Both MethaneAIR and MethaneSAT data products are available in the public domain.

\subsection{MethaneAIR}

MethaneAIR \cite{ccm2024methaneair, chulakadabba2023methane, Guanter_2024, Warren_2024} is an airborne simulator for the MethaneSAT satellite mission, designed to capture high-resolution spectral imagery for atmospheric composition analysis. MethaneAIR flights are conducted at typically 40,000 ft above ground. MethaneAIR measurements have been conducted on both NCAR GV and Learjet-35 platforms, and over 100 flight hours have been completed across the US, observing major oil and gas production areas. Each individual MethaneAIR campaign maps roughly 100 km x 100 km regions in over 2 hours, which are resolved at high precision and high spatial resolution.

The L1B dataset comprises 508 hyperspectral cubes, each consisting of 1024 spectral bins. MethaneAIR L1B spectra are aggregated by a factor 5 in the across-track direction. We center cropped our images, leading to \textasciitilde300×178 spatial soundings (along-track × across-track). The spectral range covers key absorption features for CH$_{4}$ and CO$_{2}$ (1592-1678 nm). In order to reduce computational costs of processing, the O$_{2}$ band L1B dataset was not used in this work.

Accompanying each hyperspectral image is a corresponding mask that delineates four distinct categories: clouds, cloud shadows, dark surfaces, and background. These mask labels were created during Level 2 post-processing \cite{ccm2024methaneair} using a cloud screening algorithm. This algorithm uses the following L2 derived quantities as features in a naive Bayes classifier: the retrieved minus apriori surface pressure ($\Delta$P) from the O$_{2}$ band, the O$_{2}$-band surface albedo, the ratio of the O$_{2}$-band surface albedo with its median, the relative difference between the apriori and the retrieved CO$_{2}$, O$_{2}$-band H$_{2}$O, and CO$_{2}$-band H$_{2}$O VCDs, and a terrain shadow indicator. Each feature has an associated probability density function (PDF) for clouds, shadows, and terrain shadows obtained from a set of manually labeled MethaneAIR scenes. These PDFs are used to derive cloud, shadow, and dark surface probabilities from each L2 feature before combining them with the naive Bayes classifier to yield the final cloud, shadow, and dark surface flags. This approach allows cloud flags to be derived even in the absence of O$_2$-band data and provides some robustness to systematic changes in the L2 products between different flights, mostly caused by temperature variations in the aircraft, that prevent the use of simple thresholding on the L2 fields for cloud screening.

\mpc{While the L2-derived cloud flags provide a physically consistent and operationally relevant basis for training labels, we recognize the potential for propagating systematic biases from the screening algorithm. To mitigate this, we implemented a rigorous manual curation process. All 508 scenes were visually inspected to verify label quality and consistency with the observed radiance patterns. We excluded scenes exhibiting: (1) ambiguous cloud/shadow boundaries where L2 flags showed inconsistencies with visual inspection, (2) systematic artifacts such as dark surface misclassification, and (3) full cloud coverage or completely clear conditions that provide limited training diversity. This curation ensures that our training data represents high-confidence examples where L2-derived labels align with true atmospheric conditions.}

Figure \ref{fig:methaneair_data} presents a visual representation of the data, showcasing three sample images generated from randomly selected spectral bins. These examples illustrate the diverse spectral information captured across different wavelengths and highlight the spatial resolution of the MethaneAIR instrument.

\subsection{MethaneSAT}

For MethaneSAT, the CH$_{4}$ spectrometer operates between 1598-1683 nm. The satellite does not observe continuously, it instead collects data over a discrete list of targets chosen to cover \textasciitilde80\% of global oil and gas production. Each target is acquired in \textasciitilde30 seconds and covers a \textasciitilde220x200 km$^{2}$ area with a spatial resolution of \textasciitilde100x400 m$^{2}$ (across-track x along-track, when looking at nadir). Our dataset contains a total of 262 hyperspectral samples. Since MethaneSAT is an agile observing system with more than 20 degree pointing ability, individual scenes can frequently be mapped with wider swaths that are in the vicinity of 400-450 km widths.  

For training our cloud and shadow detection algorithms, we utilize MethaneSAT's cloud screening procedure, which generates masks identifying clouds and shadows in the imagery. Unlike on the aircraft, the satellite instrument is not affected by temperature variability so a simpler cloud screening algorithm can be used effectively. The MethaneSAT cloud screening only uses $\Delta$P, and the background-corrected relative difference between the retrieved and a priori CO$_{2}$ VCDs ($\delta CO_{2}$). Cloud flags correspond to $\Delta$P$<$-20hPa and $\delta CO_{2}$$<$-2\%. Shadow flags correspond to $\Delta$P$>$20hPa and $\delta CO_{2}$$>$2\%. A potential caveat of the resulting set of flags when using them as a training set is that a pointing error over strong topography can also lead to large $\Delta$P that would be flagged as clouds/shadows. To ensure the high quality of our training set, we discard all images that contain water bodies and full cloudy areas. 

\mpc{Similar to MethaneAIR, we manually curated the MethaneSAT dataset through visual inspection of all available scenes. Specifically, we excluded: (1) scenes containing water bodies, (2) images with full cloud coverage that provide limited training value, and (3) any cases where visual inspection revealed systematic inconsistencies between the L2-derived labels and the apparent cloud/shadow patterns in the radiance imagery. This quality control process retained 262 high-quality scenes spanning diverse geographic regions, surface types, and atmospheric conditions, ensuring our training data represents reliably labeled examples rather than systematic artifacts of the L2 screening algorithm.}

Figure \ref{fig:methanesat_data} presents a visual representation of MethaneSAT's hyperspectral data. The figure displays three sample images from different spectral wavelengths alongside their corresponding classification mask. The bottom panel shows normalized radiance spectra from randomly selected spatial soundings, demonstrating the distinctive spectral signatures captured across the satellite's 220x200 km² footprint. These examples illustrate both the spatial coverage and spectral resolution of the MethaneSAT instrument.

\section{Methodology}

\begin{figure*}[ht!]
  \centering
\includegraphics[width=.9\linewidth]{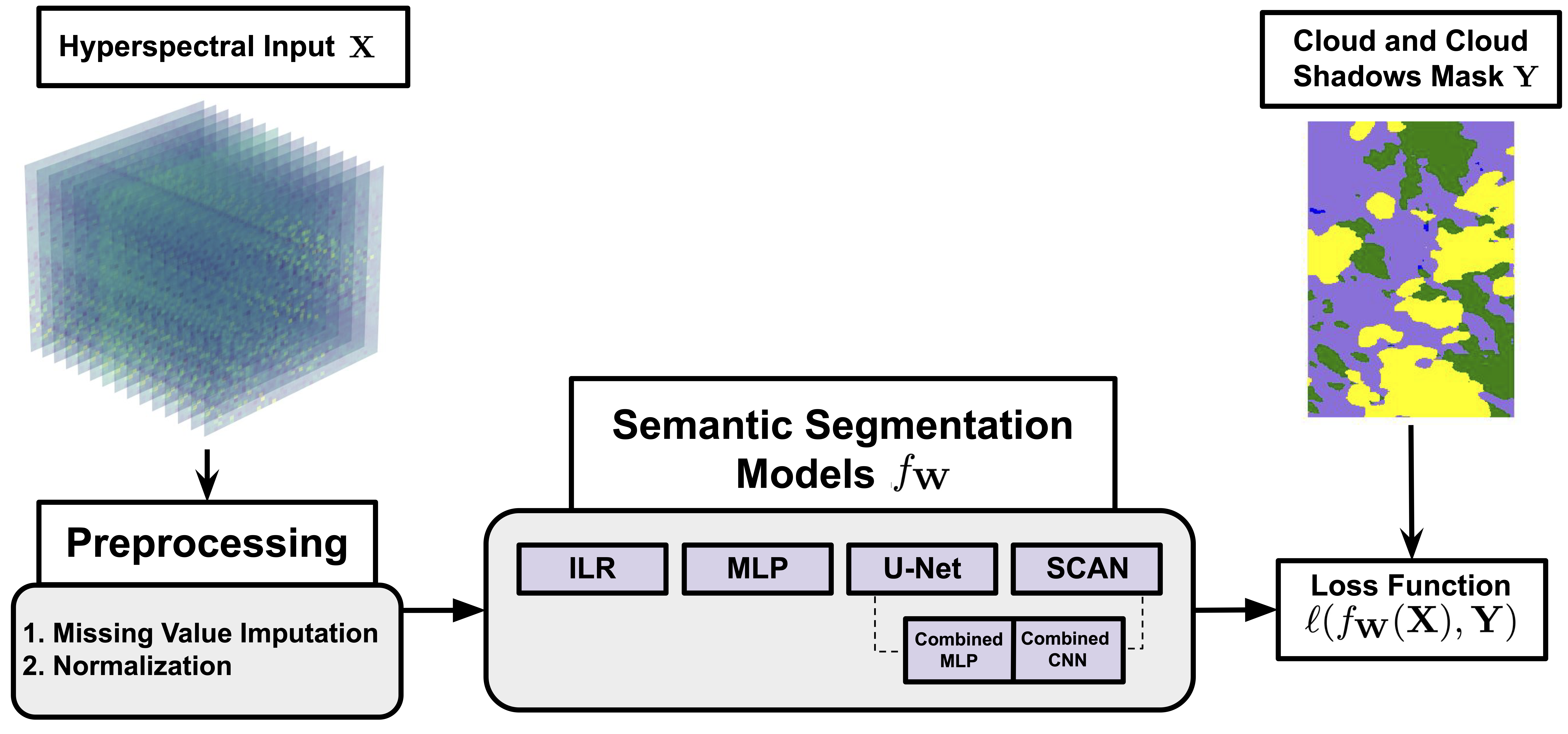}
\caption{Methodology overview for cloud and shadow detection in hyperspectral methane monitoring data. The pipeline processes input hyperspectral cubes \textbf{X} $\in \mathbb{R}^{H \times W \times C}$ through preprocessing steps (missing value imputation, spectral normalization) before semantic segmentation. We evaluate six model architectures: spectral-only methods (ILR, MLP), spatial convolution (U-Net), spectral attention (SCAN), and ensemble fusion approaches (Combined MLP/CNN) that integrate predictions from frozen U-Net and SCAN models (dashed arrows). All models are optimized using weighted cross-entropy loss $\ell(f_{\mathcal{W}}(\mathbf{X}), \mathbf{Y})$ to produce pixel-wise cloud and shadow masks \textbf{Y} $\in \{0,1\}^{H \times W \times K}$ with $K=4$ classes (MethaneAIR) or $K=3$ classes (MethaneSAT).}
  \label{fig:methdology_architecture}
\end{figure*}

Our approach for cloud and shadow detection in hyperspectral satellite imagery comprises three main components: preprocessing, semantic segmentation, and a training procedure. Additionally, an evaluation framework is developed to compare model performances. For MethaneSAT, we also developed a post-processing component that handles variable spatial dimensions through patch-based inference. \mpc{An overview of our approach is displayed in Figure \ref{fig:methdology_architecture}}

\mpc{We first establish a formal problem statement, then systematically evaluate diverse architectures spanning spectral-only methods (ILR, MLP), spatial methods (U-Net), and spectral channel attention mechanism (SCAN). This comprehensive comparison reveals the complementary strengths of different approaches and motivates our best-performing ensemble architectures (Combined MLP/CNN).}

\subsection{Problem Formulation}
\label{subsec:problem_formulation}

\mpc{Given a hyperspectral data cube $\mathbf{X} \in \mathbb{R}^{H \times W \times C}$, where $H$ and $W$ represent spatial dimensions and $C$ denotes spectral channels ($C=1024$ for MethaneAIR, $C=1080$ for MethaneSAT), our objective is to learn a mapping function $f: \mathbb{R}^{H \times W \times C} \rightarrow \mathbb{R}^{H \times W \times K}$ that assigns each spatial sounding to one of $K$ classes (background, cloud, shadow, and optionally dark surface for MethaneAIR, where $K=4$; background, cloud, shadow for MethaneSAT, where $K=3$).}

Formally, we seek to minimize the expected loss:
\begin{equation}
\mathcal{L}(\mathbf{W}) = \mathbb{E}_{(\mathbf{X},\mathbf{Y}) \sim \mathcal{D}} \left[ \ell(f_\mathbf{W}(\mathbf{X}), \mathbf{Y}) \right]
\label{eq:expected_loss}
\end{equation}
\mpc{where $\mathbf{W}$ represents learnable parameters, $\mathbf{Y} \in \{0,1\}^{H \times W \times K}$ is the one-hot encoded ground truth segmentation mask, $\ell$ is the weighted cross-entropy loss (detailed in Section~\ref{sec:results}), and $\mathcal{D}$ is the distribution of hyperspectral scenes.}


\subsection{Preprocessing Steps} \label{sec:prep}
To ensure high-quality input data and optimize model performance, we designed a rigorous preprocessing pipeline. The first step addresses missing values in the hyperspectral data. Any missing values (NaNs) within the dataset are imputed using the mean value along the spectral dimension for each affected spatial sounding, ensuring continuity in the spectral signatures. Missing values may arise, for example, over very dark or very bright features on the landscape.

For spatial standardization, we crop all MethaneAIR L1B files to 300×178 spatial soundings, creating a consistent input size across the dataset. Given the varying input size of MethaneSAT data, we randomly crop patches of 224×224 spatial soundings from the complete images during training, providing a standardized input dimension while increasing effective sample size through data augmentation.

We implement a two-step spectral bin normalization process across all 1024 and 1080 spectral bands of MethaneAIR and MethaneSAT respectively. Specifically, values are clipped to the 1st and 99th percentile to mitigate the impact of extreme outliers that could skew the model training. After clipping, each spectral bin undergoes standardization by subtracting the per-bin mean radiance value and dividing by the per-bin standard deviation. This normalization step ensures that each spectral band contributes equally to the model's learning process.

Before passing the input data to any machine learning model, a batch-wise normalization is performed across all dimensions. In this step, each sample in the batch is normalized by subtracting the mean value calculated across all spectral bands and spatial dimensions (height and width) and dividing by the standard deviation computed across these same dimensions. 
This batch-sample normalization helps reduce the influence of overall intensity variations between different images while maintaining the relative relationships between spectral bands and spatial structures that are crucial to distinguish between clouds, shadows, and clear sky. 

\subsection{Semantic Segmentation Algorithms} 
We evaluate six distinct architectures for cloud and shadow segmentation, ranging from conventional spectral-based methods to advanced deep learning approaches: Iterative Logistic Regression (LR; \cite{Edisanter_2016, park_2023, Gautam_2023}), Multilayer Perceptron (MLP; \cite{BENEDIKTSSON_1993, Yang_1999, Tian_1999, Taravat_2015, gomezchova_2017}), U-Net \cite{Ronneberger2015, Jiao_2020, Wieland_2019, Miao_2022, wright_2024, Li_2022, Tan_2023}, and the Spectral Attention Network (SCAN), along with the combined models. This systematic comparison reveals the progression of capabilities and the complementary strengths of spatial versus spectral modeling.




\subsubsection{Iterative Logistic Regression (ILR)}

We implement the Iterative Logistic Regression (ILR) approach introduced by \cite{park_2023} as a baseline model for our hyperspectral shadow detection system. This established method learns a compact spectral basis that effectively distinguishes between shadowed and non-shadowed spatial soundings through an iterative dimensionality reduction process.

Given an input spectrum $\mathbf{x}_i \in \mathbb{R}^C$ for spatial sounding $i$, we first compute its mean radiance $m_i = \frac{1}{C}\sum_{c=1}^C x_{i,c}$ and apply a logarithmic normalization transformation:
\begin{equation}
\mathbf{s}_i = \log\left(\mathbf{x}_i / m_i\right)
\label{eq:ilr_normalization}
\end{equation}
where $\log$ denotes element-wise natural logarithm. This preprocessing step isolates spectral shape features from overall brightness variations, making the representation more robust to illumination changes. The core algorithm iteratively extracts spectral components most relevant for cloud and shadow detection:

1. Using log spectra $s_i$ and their associated labels $y_i \in \mathcal{R}^{K}$, where K is the number of classes, the algorithm trains a logistic regression classifier to find spectral weights $\mathbf{W}$ that separate each of the K classes.

2. After identifying these discriminative weights, their contribution is projected out from all spectra.

3. This process repeats $z$ times until the classification performance (measured by F1-score) converges or falls below a predefined threshold.

Through this iterative process, we compute a low-dimensional representation $\boldsymbol{\beta}_i \in \mathbb{R}^z$, with $z=23$ components (compared to the original 1024 and 1080 spectral bands for MethaneAIR and MethaneSAT respectively), which captures the most salient spectral features for shadow detection.

The learned low-dimensional representations $\boldsymbol{\beta}_i$ are used for to obtain output probabilities for each spatial sounding:
\begin{equation}
P(\mathbf{y}_i | \mathbf{x}_i) = \text{softmax}\left(\mathbf{W}_{\text{class}}^T \boldsymbol{\beta}_i\right)
\label{eq:ilr_classification}
\end{equation}
where $\mathbf{W}_{\text{class}} \in \mathbb{R}^{z \times K}$ is trained via multinomial logistic regression on the compressed features.

\subsubsection{Multilayer Perceptron (MLP)}

Building upon the spectral feature extraction approach of ILR, we implement a Multilayer Perceptron (MLP; \cite{werbos_1970}) model that processes hyperspectral data on a pixel-wise basis. \mpc{The MLP treats each spatial sounding $\mathbf{x}_i \in \mathbb{R}^C$ independently, learning a direct non-linear mapping from the full spectral signature to class probabilities without explicit dimensionality reduction.}

Our MLP implementation processes each spatial sounding's complete spectral signature independently through a neural network architecture. The network architecture consists of an input layer with $C$ nodes, followed by two hidden layers with with $d_1 = d_2 = 20$ nodes each, and an output layer with $K$ nodes corresponding to our target classes. Mathematically, for each spatial sounding $x_i$, the probability of belonging to each class can be computed as:

\begin{align}
\mathbf{h}^{(1)}_i &= \phi\left(\mathbf{W}^{(1)}\mathbf{x}_i + \mathbf{b}^{(1)}\right) \label{eq:mlp_layer1} \\
\mathbf{h}^{(2)}_i &= \phi\left(\mathbf{W}^{(2)}\mathbf{h}^{(1)}_i + \mathbf{b}^{(2)}\right) \label{eq:mlp_layer2} \\
P(\mathbf{y}_i | \mathbf{x}_i) &= \text{softmax}\left(\mathbf{W}^{(3)}\mathbf{h}^{(2)}_i + \mathbf{b}^{(3)}\right) \label{eq:mlp_output}
\end{align}

where $\mathbf{W}^{(1)} \in \mathbb{R}^{20 \times C}$, $\mathbf{W}^{(2)} \in \mathbb{R}^{20 \times 20}$, and $\mathbf{W}^{(3)} \in \mathbb{R}^{K \times 20}$ are weight matrices; $\mathbf{b}^{(1)} \in \mathbb{R}^{20}$, $\mathbf{b}^{(2)} \in \mathbb{R}^{20}$, and $\mathbf{b}^{(3)} \in \mathbb{R}^K$ are bias vectors; and $\phi(\cdot) = \max(0, \cdot)$ is the Rectified Linear Unit (ReLU; \cite{Agarap_2018}) activation function.

\mpc{Unlike ILR, which explicitly reduces dimensionality through iterative feature extraction, the MLP learns a direct non-linear mapping from the full spectral signature to class probabilities. This allows the network to capture complex spectral patterns without the need for explicit feature engineering, potentially discovering discriminative spectral relationships not captured by linear projections.}

\subsubsection{U-Net Architecture}
\label{subsubsec:unet}


We adopt the U-Net architecture \citep{Ronneberger2015} for its proven effectiveness in dense prediction tasks through its multi-scale spatial feature extraction. Unlike pixel-wise methods (ILR, MLP), U-Net captures hierarchical spatial features through its symmetric encoder-decoder structure with skip connections, enabling accurate boundary delineation essential for precise cloud and shadow detection \citep{Wei_2023, ruzicka_2023}.


\textbf{Contracting Path (Encoder)}: The encoder comprises three stages, each consisting of two convolutional layers followed by a downsampling operation. The convolutional layers use $3 \times 3$ kernels with padding to preserve spatial dimensions and a stride of $1$. Each convolution is followed by batch normalization and a ReLU activation function. After the convolutions, a $2 \times 2$ max pooling operation with stride 2 reduces the spatial resolution by half.

The feature progression for each stage is: Input ($C$ channels) $\rightarrow$ 8 channels $\rightarrow$ 16 channels $\rightarrow$ 32 channels. This systematic increase in feature representation capacity while reducing spatial dimensions enables the network to encode progressively higher-level contextual information at each stage, transitioning from fine-grained spectral patterns to abstract semantic concepts.

This progression increases the feature representation capacity while reducing spatial dimensions, enabling the network to encode higher-level contextual information.

\textbf{Expanding Path (Decoder)}: The decoder symmetrically reconstructs the spatial resolution using three upsampling stages. Each stage begins with a transposed convolution (also known as deconvolution) with a kernel size of $3 \times 3$, stride $2$, and padding to double the spatial resolution. This is followed by the concatenation of the corresponding feature map from the encoder (via skip connections), and two $3 \times 3$ convolutions with batch normalization and ReLU activations, similar to the encoder.

The channel progression of each upsampling stage reverses the encoder: 32 channels $\rightarrow$ 16 channels $\rightarrow$ 8 channels $\rightarrow$ $K$ output classes, where each reduction maintains spatial detail while focusing the representation toward final class predictions.


\textbf{Skip Connections}: To retain spatial information lost during downsampling, we employ skip connections that directly pass feature maps from the encoder stages to their corresponding decoder stages. These connections concatenate encoder features with upsampled decoder features along the channel dimension. For example, features from encoder stage 2 (16 channels) are concatenated with upsampled features from decoder stage 2 (16 channels), resulting in a 32-channel input to the following convolutional layers. This fusion of high-resolution spatial detail and abstract semantic information enables accurate boundary delineation and class prediction at the spatial sounding level.


\textbf{Output Layer}: The final layer produces a tensor with dimensions $ K \times H \times W$, where $K$ is the number of classes. A softmax function is applied to generate a probability distribution over classes for each spatial sounding:

$$P(\mathbf{Y} | \mathbf{X}) = \text{softmax}(f_{U-Net}(\mathbf{X}))$$

This formulation allows the model to produce a dense segmentation mask with per-pixel class labels, suitable for hyperspectral semantic segmentation tasks.

\subsubsection{Spectral Channel Attention Network}
\label{subsubsec:scan}

While U-Net architectures have become standard in semantic segmentation tasks, they often struggle with accurate delineation of cloud and shadow boundaries due to their reliance on fixed receptive fields and limited incorporation of spectral information. Specifically, the transition zones between clouds/shadows and land cover features present challenges where spectral confusion leads to misclassification.

Attention mechanisms have been widely adopted in computer vision, but their application to spectral band selection in hyperspectral cloud/shadow segmentation presents special challenges and opportunities. Unlike natural images where spatial attention focuses on semantic objects, spectral attention specifically addresses the band selection problem inherent in hyperspectral analysis. The key insight is that different spectral regions provide varying levels of discrimination between clouds/shadows and surface materials, particularly at boundary regions where spectral mixing occurs.

We reformulate the classical attention mechanism from \cite{bahdanau_2014} as a Spectral Channel Attention Network (SCAN), adapting channel-wise attention for hyperspectral band selection in cloud and shadow detection tasks. \mpc{An overview of this model is displayed in Appendix \ref{appB} Figure  \ref{fig:scan_architecture}}. This approach directly addresses the boundary delineation problem by dynamically weighting the importance of different spectral bands based on their discriminative power for cloud/shadow detection. Rather than treating all spectral bands equally as in conventional approaches, our method learns to emphasize wavelengths that are most informative for distinguishing clouds/shadows from underlying surfaces, particularly in spectrally ambiguous transition zones. The model consists of two main components: a spectral attention module and a pixel-wise classification framework. 

\textbf{Spectral Attention Module.} Given input $\mathbf{X} \in \mathbb{R}^{H \times W \times C}$, \mpc{we first compute spatially-pooled features that summarize global spectral characteristics:}
\begin{equation}
\bar{\mathbf{x}} = \text{GlobalAveragePooling}(\mathbf{X}) \in \mathbb{R}^{C}
\label{eq:scan_pooling}
\end{equation}

\mpc{This spatial aggregation is motivated by the observation that spectral discriminability is a global property (e.g., the 1640 nm band is informative for cloud detection everywhere in the scene), not a local, position-dependent characteristic. The pooled features are then passed through a multilayer perceptron (MLP) to generate attention weights}:
\begin{align}
\mathbf{z} &= \text{ReLU}\left(\mathbf{W}_1 \bar{\mathbf{x}} + \mathbf{b}_1\right) \quad \text{where } \mathbf{W}_1 \in \mathbb{R}^{\frac{C}{r} \times C} \label{eq:scan_bottleneck} \\
\boldsymbol{\alpha} &= \sigma\left(\mathbf{W}_2 \mathbf{z} + \mathbf{b}_2\right) \quad \text{where } \mathbf{W}_2 \in \mathbb{R}^{C \times \frac{C}{r}} \label{eq:scan_attention}
\end{align}
\mpc{where $\sigma(\cdot)$ is the sigmoid function, $r=16$ is the reduction ratio controlling bottleneck dimensionality, and $\boldsymbol{\alpha} \in [0,1]^{B \times C}$ represents learned channel importance weights. Values near 1 amplify discriminative bands; values near 0 suppress uninformative bands.}

The attended features are obtained via channel-wise multiplication ($\boldsymbol{\alpha}$ across spatial dimensions):
\begin{equation}
\mathbf{X}_{\text{att}} = \mathbf{X} \odot \alpha
\label{eq:scan_weighting}
\end{equation}


\textbf{Classification Framework}:
The attended features are passed through a fully-connected neural network with $N$ layers as follows:
$$
f_{MLP}(\mathbf{X}) = \mathbf{W'}_N(\phi(\mathbf{W'}_{N-1}(...\phi(\mathbf{W'}_1X_{att})))), 
$$

\mpc{where $\mathbf{W'}_i$ are the classifier learnable weight matrices}, $\phi$ denotes ReLU activation function. The final probabilities are obtained through a softmax activation function:
$$
P(\mathbf{Y}|\mathbf{X}) = \text{Softmax}(f_{MLP}(\mathbf{X}_{att})).
$$

\mpc{Note that the spectral attention module parameters $\mathbf{W}_1$ and $\mathbf{W}_2$, along with the classifier parameters $\mathbf{W'}_i$, are jointly optimized during training.}

\subsubsection{Combined Models}
\mpc{To leverage the complementary strengths of spatial (U-Net) and spectral (SCAN) approaches, we develop ensemble architectures that fuse their predictions. We explore two fusion strategies: MLP-based fusion (pixel-wise combination) and CNN-based fusion (spatially-aware combination).}

\textbf{Combined MLP Architecture}: 

 This model fuses model predictions using a multilayer perceptron (MLP). The architecture employs two pre-trained base models (U-Net and SCAN) with frozen weights to maintain their individual predictive capabilities. Given the input hyperspectral image $\mathbf{X}$, we first obtain predictions from frozen pre-trained models:

 \begin{align}
\mathbf{P}_{\text{UNet}} &= f_{\text{UNet}}(\mathbf{X}; \mathbf{W}_{\text{UNet}}^*) \in \mathbb{R}^{H \times W \times K} \label{eq:comb_pred_unet}, \\
\mathbf{P}_{\text{SCAN}} &= f_{\text{SCAN}}(\mathbf{X}; \mathbf{W}_{\text{SCAN}}^*) \in \mathbb{R}^{H \times W \times K}, \label{eq:comb_pred_scan}
\end{align}
 
\mpc{where $\mathbf{W}^*$ denotes fixed weights trained independently as described in Sections~\ref{subsubsec:unet} and \ref{subsubsec:scan}. Freezing base model weights prevents degradation during fusion training.}

We concatenate predictions channel-wise:
\begin{equation}
\mathbf{P}_{\text{combined}} = [\mathbf{P}_{\text{UNet}} \,||\, \mathbf{P}_{\text{SCAN}}] \in \mathbb{R}^{H \times W \times 2K}
\label{eq:comb_concat}
\end{equation}

The MLP fusion network with $M$ hidden layers processes each spatial sounding independently:

$$
f_{fusion}(\mathbf{X}) = \mathbf{W}_M(\phi(\mathbf{W}_{M-1}(...\phi(\mathbf{W}_1P_{combined})))), 
$$

where $\mathbf{W}_i$ are learnable weight matrices, and $\phi$ is the ReLU activation function. Dropout with rate $\delta=0.2$ is applied after each hidden layer for regularization. The final output provides the fused class probabilities as follows:

$$
P(\mathbf{Y}|\mathbf{X}) = \text{Softmax}(f_{fusion}(P_{combined}))
$$

\textbf{Combined CNN Architecture}: 

\mpc{While MLP fusion treats each spatial sounding independently, CNN fusion preserves spatial relationships during combination, enabling the fusion network to leverage spatial context when resolving disagreements between base models}. Using the same pre-trained U-Net and SCAN models with frozen weights, the concatenated predictions $P_{combined} \in \mathbb{R}^{B \times H \times W \times 2K}$ are processed through a series of convolutional layers:

$$F_1 = \phi(\text{Conv}_1(P_{combined})) \in \mathbb{R}^{B \times H \times W \times C_1}$$
\begin{align}
F_i &= \phi(\text{Conv}_i(F_{i-1})) \in \mathbb{R}^{B \times H \times W \times C_i}, \nonumber \\
&\quad i \in \{2,...,N-1\}
\end{align}
$$F_N = \text{Conv}_N(F_{N-1}) \in \mathbb{R}^{B \times H \times W \times K}$$

where $\text{Conv}_i$ represents a 2D convolutional layer with kernel size $3 \times 3$ and padding size 1 (except for the final layer which uses $1 \times 1$ convolution), preserving original spatial dimensions. $\phi$ is the ReLU activation function, and $C_i$ are the channel dimensions of the intermediate feature maps (e.g., $C_1=64$, $C_2=32$, $C_3=16$ in our work). Dropout with rate $\delta=0.2$ is applied after each intermediate convolutional layer. The final class probabilities are obtained through:

$$
P(\mathbf{Y}|\mathbf{X}) = \text{Softmax}(F_N)
$$

Each convolutional layer captures local relationships between the predictions of both models, allowing the network to learn spatial patterns in prediction agreement or disagreement. This spatial-aware fusion can be particularly beneficial for complex scenes where the performance of individual models varies spatially.

\subsection{Training}
\label{subsec:training}

We train all models using a weighted cross-entropy loss function that accounts for class imbalance in our dataset. For a batch of N samples, the loss function is defined as:

$$
\mathcal{L} = -\sum_{n=1}^N \sum_{k=1}^K w_k y_{n,k} \log(\hat{y}_{n,k})
$$

where $K$ represents the number of classes, $w_k$ denotes the class weight assigned to class $k$, $y_{n,k}$ is the ground truth binary indicator (1 if sample $n$ belongs to class $k$, and 0 otherwise), and $\hat{y}_{n,k}$ represents the model's predicted probability that sample $n$ belongs to class $k$. The class weights $w_k$ are computed as the inverse of the class frequencies in the training set, helping to address the inherent imbalance between cloud, shadow, and clear sky spatial soundings.

For optimization, we employ the Adam optimizer with a learning rate determined through cross-validation experiments detailed in Appendix \ref{appA}. To enhance model generalization, we implement data augmentation techniques including random horizontal and vertical flips, and rotations at multiples of 90 degrees. These augmentations help the models learn invariance to common geometric transformations while preserving the physical meaning of the spectral signatures. Training proceeds over a maximum of 100 epochs with randomly shuffled batches of size 32. We employ an early stopping strategy to prevent overfitting, where training is terminated if the validation loss does not improve for 20 consecutive epochs (patience period), and we save the best-performing model checkpoint based on validation performance for final evaluation.

\subsection{Post-processing and Evaluation}

\subsubsection{Post-processing for MethaneSAT}

To accommodate the varying input dimensions of MethaneSAT hyperspectral imagery, we implemented a patch-based evaluation strategy. This approach enables consistent model application across diverse acquisition scenarios while maintaining spatial context. Specifically, each variable-sized hyperspectral image is systematically partitioned into overlapping patches of 224×224 spatial soundings with a stride of 112 spatial soundings, ensuring 50\% overlap between adjacent regions. The segmentation model, trained exclusively on fixed-dimension patches, processes each segment independently. Subsequently, we employ a weighted averaging scheme in overlapping regions, where model predictions from multiple patches contribute proportionally to the final segmentation map. We show that this simple methodology effectively addresses the dimensional heterogeneity in MethaneSAT data.

\subsubsection{Evaluation}

To comprehensively evaluate the performance of our cloud and shadow detection models, we employed four key metrics:
\begin{itemize}
    \item \textbf{Accuracy}: The overall correctness of the model, calculated as the proportion of correctly classified spatial soundings across all classes. Although informative, accuracy alone can be misleading for imbalanced datasets where one class significantly outnumbers others (e.g., when clear land spatial soundings vastly outnumber cloud and shadow spatial soundings in the imagery).
    \item \textbf{Precision}: Represents the model's exactness in identifying clouds and shadows, calculated as the ratio of correctly predicted cloud/shadow spatial soundings to all spatial soundings predicted as cloud/shadow by the model. High precision indicates a low false positive rate, meaning when the model predicts a spatial sounding as cloud or shadow, it is likely to be correct. Precision = $\frac{\text{True Positives}}{\text{True Positives + False Positives}}$
    \item \textbf{Recall}: Measures the completeness of the model in detecting actual clouds and shadows, calculated as the ratio of correctly predicted cloud/shadow spatial soundings to all ground truth cloud/shadow in the dataset. High recall indicates the model's ability to identify most of the actual clouds and shadows.
Recall = $\frac{\text{True Positives}}{\text{True Positives + False Negatives}}$
 \item \textbf{F1-Score}: The harmonic mean of precision and recall, providing a balanced measure of the model's performance that accounts for both false positives and false negatives. It is particularly useful when you want to seek a balance between precision and recall and when there is an uneven class distribution. F1-Score = $2 \times \frac{\text{Precision} \times \text{Recall}}{\text{Precision + Recall}}$. 
\end{itemize}

 We use macro-averaging, computing the precision, recall, and F1-score (i.e. we compute the metrics for each class independently and average them to report final results). The reported values include the test set mean performance and standard deviation in 3 cross-validated folds, ensuring a robust evaluation of model performance. See Appendix \ref{appA} for details of the hyperparameter selection.

\section{Results} \label{sec:results}

\subsection{MethaneAIR}

We evaluated six different architectures for cloud and shadow segmentation: Iterative Logistic Regression (ILR), Multi-Layer Perceptron (MLP), U-Net, the Spectral Attention Network (SCAN), and two ensemble approaches—Combined MLP and Combined CNN. Table \ref{tab:model_comparison_methaneair} presents the quantitative comparison of the performance metrics of these models.

\begin{table}[h]
\centering
\begin{tabular}{|l|c|c|c|c|}
\hline
Model & Accuracy & F1 & Precision & Recall \\
\hline
ILR & 73.81+-4.05 & 62.07+-0.86 & 61.33+-0.67 & 72.59+-1.46 \\
MLP & 82.49±2.24 & 71.29±1.02 & 68.24±1.04 & 81.42±0.85 \\
U-Net & 88.26±0.45 & 76.24±1.90 & 72.59±2.13 & 83.65±1.03
\\
SCAN & 86.51±2.90 & 74.96±0.96 & 72.17±1.60 & 83.46±3.13
\\
Comb. MLP & 88.92±1.80 & 76.99±6.78 & 72.79±6.38 & 86.34±6.32
\\
Comb. CNN & \textbf{89.42±1.20} & \textbf{78.50±3.08} &  \textbf{74.44±1.89} &  \textbf{88.97±2.77}
\\
\hline
\end{tabular}
\caption{Performance comparison of different models for the MethaneAIR dataset.}
\label{tab:model_comparison_methaneair}
\end{table}

\begin{figure*}[h!]
  \centering
\includegraphics[width=.9\linewidth]{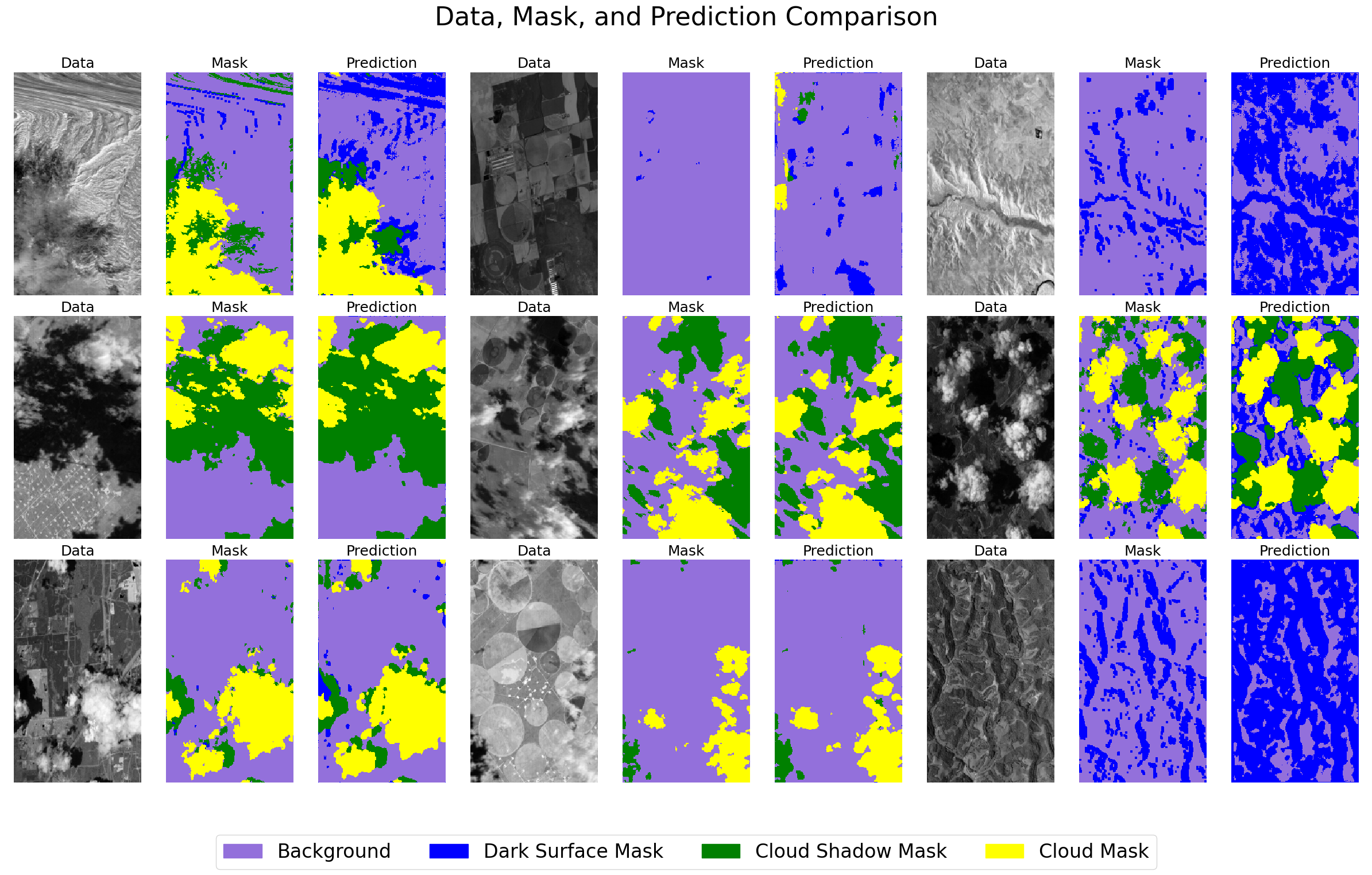}
  \caption{Radiance at 1592nm, labels and predictions of the Combined CNN model. All figures correspond to the test set from the first cross-validation fold (where the dataset was split into 3 parts, with one part held out for testing)}.
  \label{fig:preds_combined_cnn}
\end{figure*}

\begin{figure*}[h!]
\centering
\includegraphics[width=.9\linewidth]{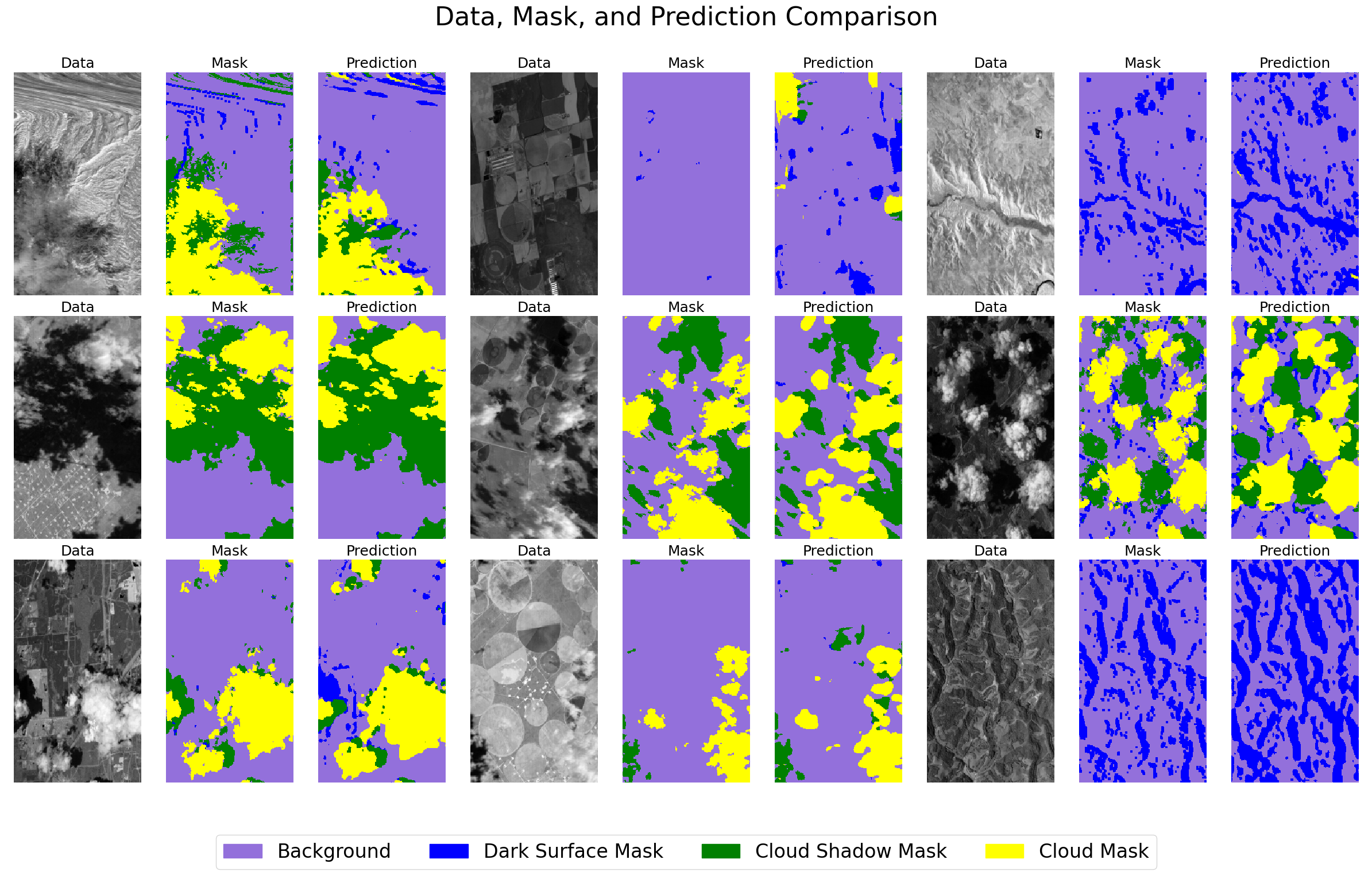}
\caption{Radiance at 1592nm, ground truth labels, and predictions of the U-Net model for MethaneAIR data. The model demonstrates significantly improved spatial coherence but exhibits over-smoothed boundaries.}
\label{fig:preds_unet}
\end{figure*}

\begin{figure*}[hb!]
\centering
\begin{subfigure}{0.30\textwidth}
\includegraphics[width=\textwidth]{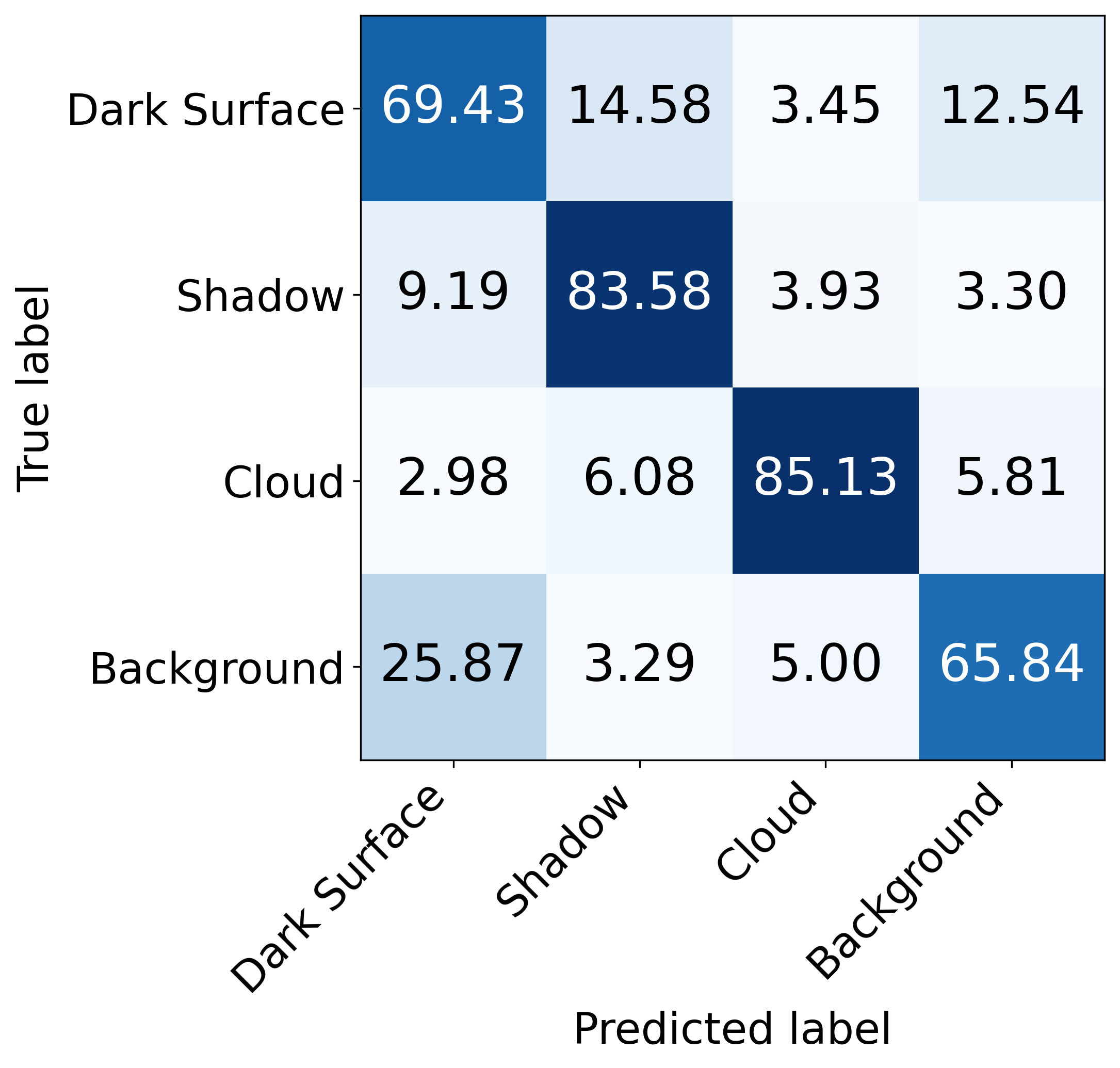}
\caption{ILR}
\end{subfigure}
\begin{subfigure}{0.30\textwidth}
\includegraphics[width=\textwidth]{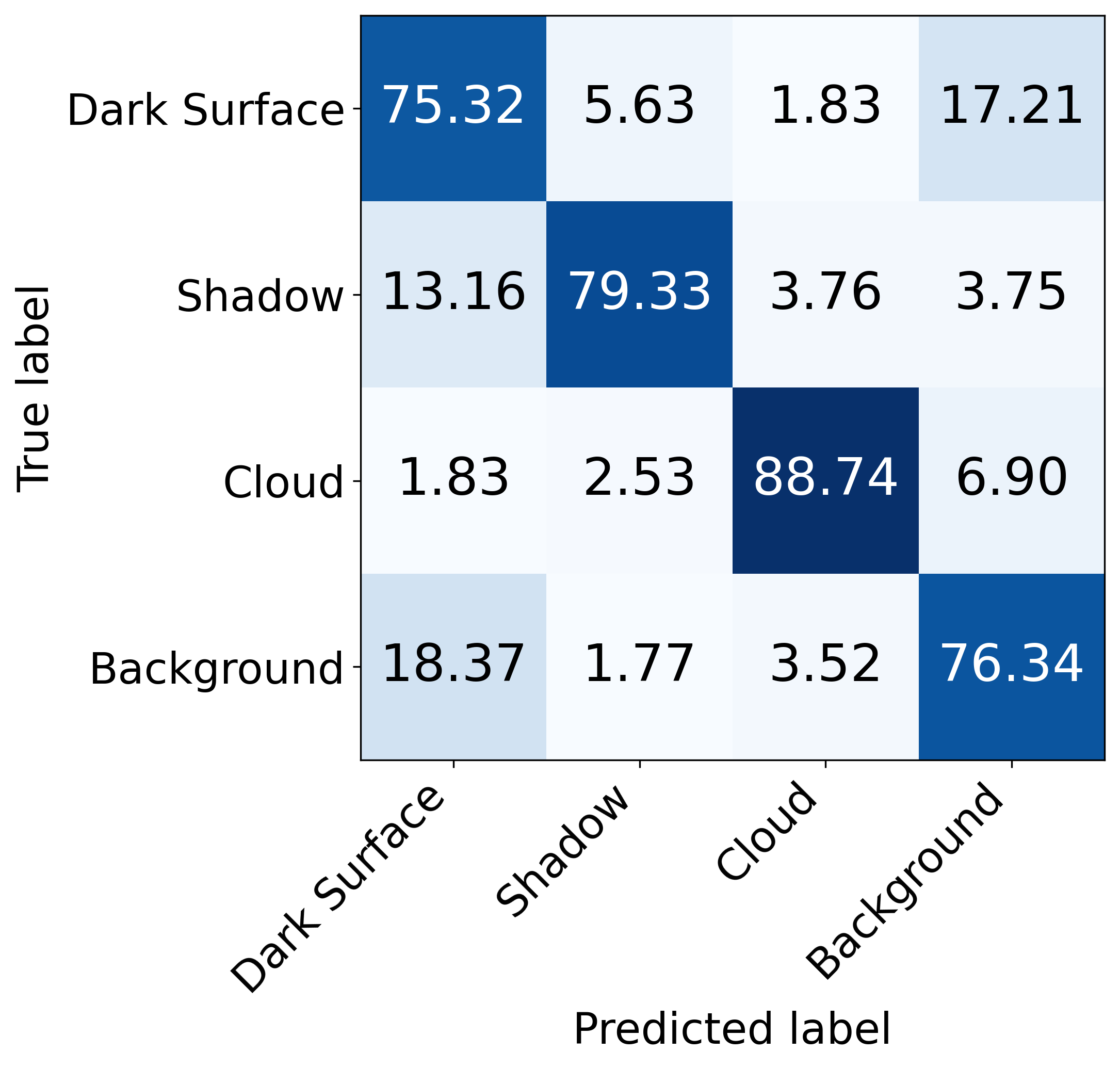}
\caption{MLP}
\end{subfigure}
\begin{subfigure}{0.30\textwidth}
\includegraphics[width=\textwidth]{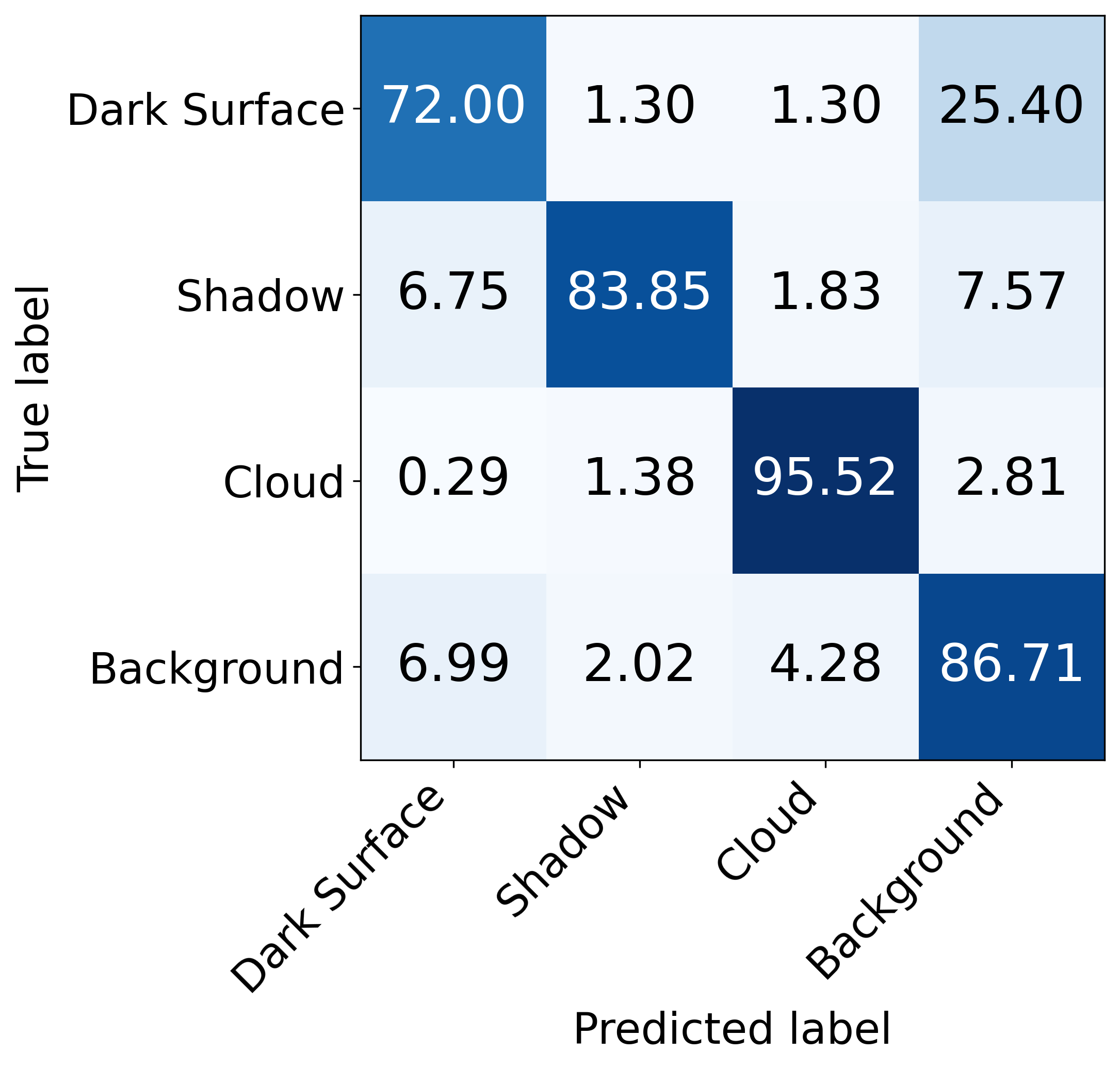}
\caption{U-Net}
\end{subfigure}
\begin{subfigure}{0.30\textwidth}
\includegraphics[width=\textwidth]{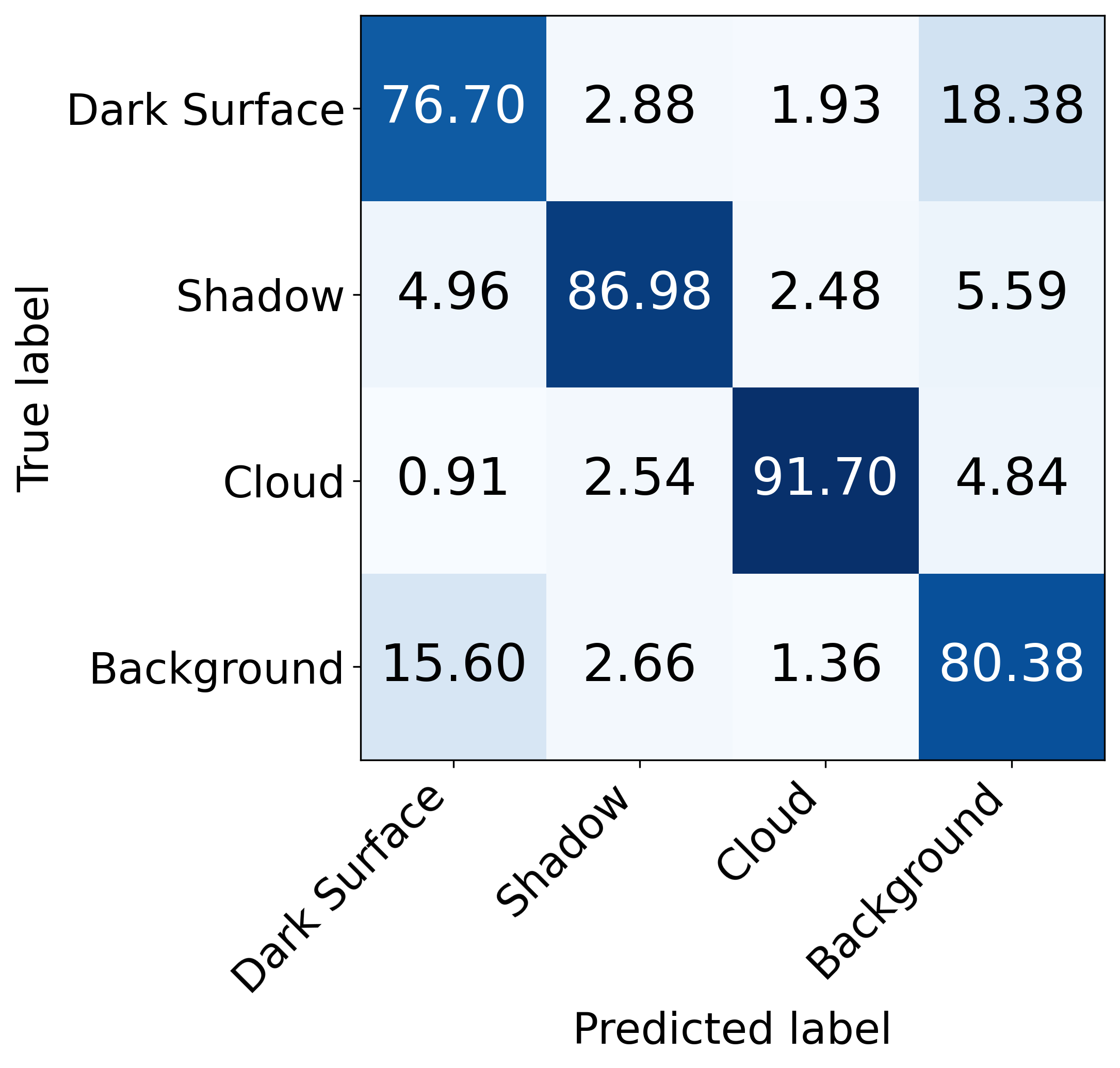}
\caption{SCAN}
\end{subfigure}
\begin{subfigure}{0.30\textwidth}
\includegraphics[width=\textwidth]{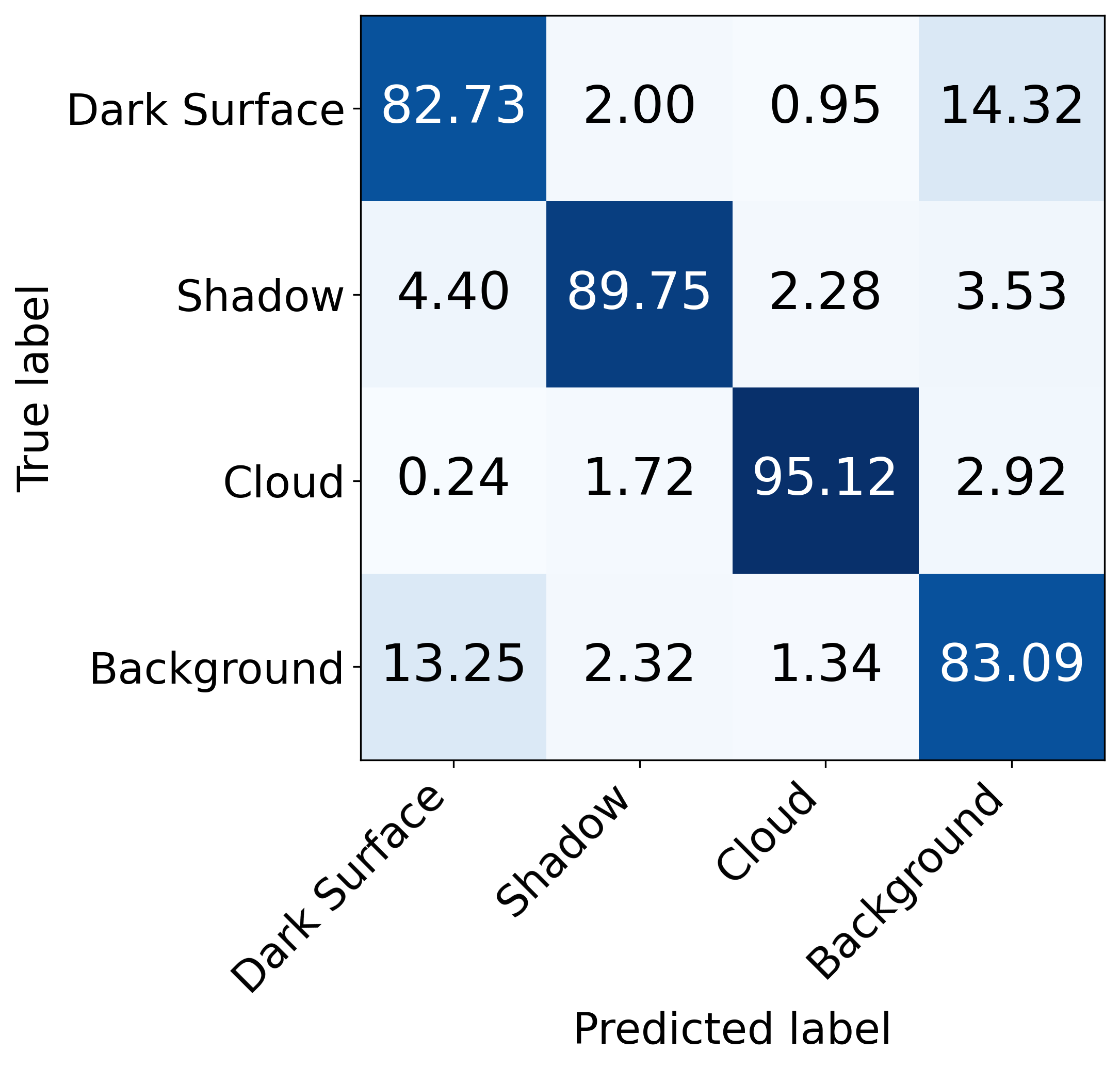}
\caption{Combined MLP}
\end{subfigure}
\begin{subfigure}{0.30\textwidth}
\includegraphics[width=\textwidth]{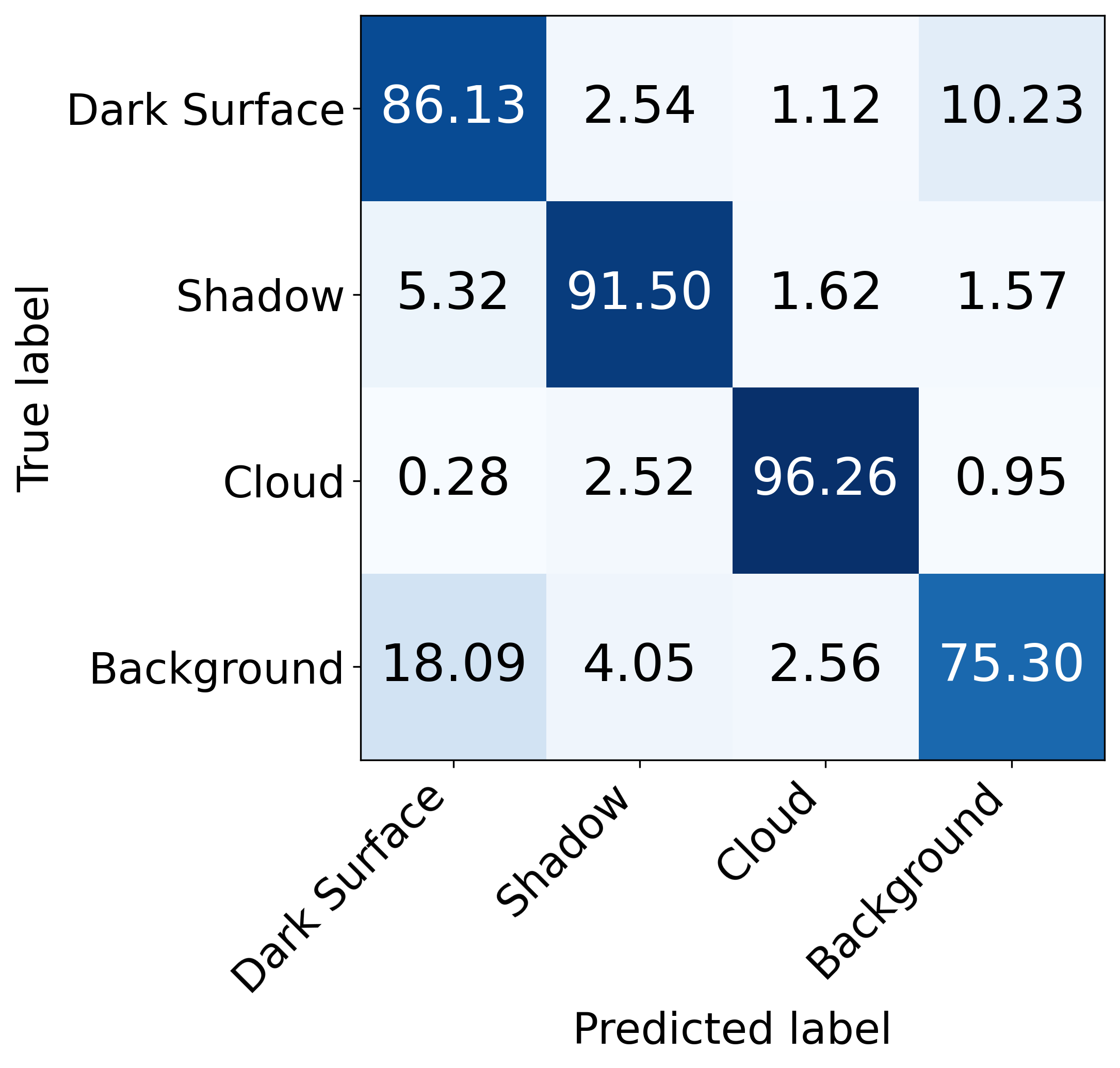}
\caption{Combined CNN}
\end{subfigure}
\caption{Confusion matrices for all evaluated models on MethaneAIR data. All the results were computed using the test set over the first cross-validated fold.}
\label{fig:all_conf_m_mair}
\end{figure*}

\begin{figure*}[h!]
  \centering
  \includegraphics[width=.9\linewidth]{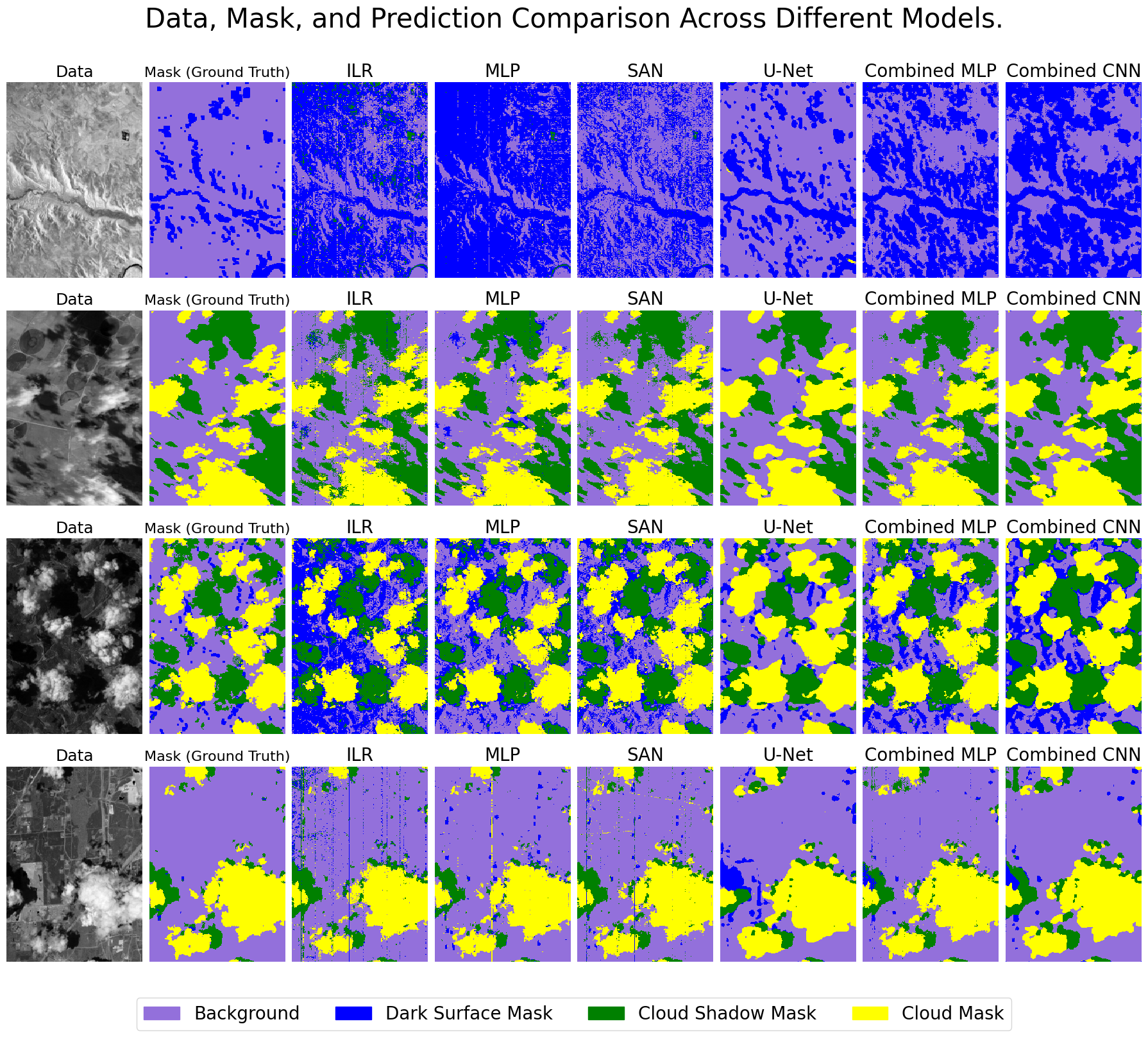}
  \caption{Prediction comparison across all evaluated models for MethaneAIR scenes. All images belongs to the test set and were computed on the first cross-validated fold.}
  \label{fig:preds_overall_mair}
\end{figure*}

The Combined CNN architecture achieved the best overall performance with an accuracy of 89.42±1.20\% and an F1-score of 78.50±3.08\%. Figure \ref{fig:preds_combined_cnn} illustrates predictions for 9 random samples from our dataset. As can be seen, the model successfully captures complex cloud formations and their associated shadows across diverse landscapes. The model demonstrates particularly strong performance in identifying cloud patterns (yellow regions) with high precision, as evidenced in the second row of samples where the predicted cloud boundaries closely align with the ground truth masks. Shadow detection (green regions) also shows high accuracy, particularly visible in the middle row images where the model correctly identifies the areas affected by cloud shadows despite varying surface reflectance properties underneath. While not achieving the highest overall metrics, the U-Net architecture emerges as the second best-performing model due to its remarkable capacity for detecting dark surfaces. The performance of this model can be visualized in Figure \ref{fig:preds_unet}, which demonstrates its effectiveness in challenging detection scenarios.

Figure \ref{fig:all_conf_m_mair} presents a comparative view of confusion matrices for all evaluated models on the MethaneAIR dataset. This side-by-side comparison reveals a clear progression in classification performance across model architectures. The ILR model shows significant confusion between background and dark surface classes (18.09\% misclassification), while the MLP improves but still struggles with shadow detection. The SCAN model demonstrate substantially higher diagonal values, indicating improved class-specific accuracy, particularly for cloud detection (91.20\%). However, it still exhibits confusion between shadows and dark surfaces. The U-Net model achieves best accuracy for background detection (86.71\%), and notably reduces the dark surface's false positives. This can be seen in the first row of Figure \ref{fig:all_conf_m_mair}. The Combined MLP approach reduces cloud and shadow misclassifications, while the Combined CNN achieves the highest performance across all classes, with notably improved shadow detection (91.50\% accuracy) and significantly reduced confusion between spectrally similar categories.



Figure \ref{fig:preds_overall_mair} reveals the distinctive characteristics of each approach across four representative scenes. The ILR and MLP models produce noisy, fragmented predictions due to their pixel-wise processing, particularly evident in the first row where terrain features are inconsistently classified with scattered dark surfaces. In contrast, the SCAN model shows notable improvement in capturing accurate boundaries of clouds and shadows through its spectral attention mechanism, though it still exhibits some noise in spectrally ambiguous regions. Interestingly, while U-Net does not achieve the highest quantitative metrics, visual inspection shows it produces remarkably balanced predictions with significantly less overprediction of dark surfaces compared to other models. This suggests that its lower classification metrics may be attributed to its tendency to generate overly smoothed boundaries around cloud and shadow regions rather than to fundamental misclassifications.

The choice of optimal method should ultimately be determined by domain experts who understand the specific application requirements and downstream processing needs. For instance, in applications where subsequent processing steps include boundary smoothing or where 3-D cloud effects not captured in training masks lead to edge artifacts, U-Net's tendency toward smoother boundaries may actually be advantageous rather than detrimental. Oversmoothing is not necessarily problematic in scenarios where precise edge delineation is less critical than avoiding false positive detections, or where post-processing workflows are designed to handle boundary refinement. Therefore, the "best" model depends on whether the application prioritizes sharp boundary accuracy.

Detailed hyperparameter experiments can be found in Appendix \ref{appA}, and comprehensive predictions for all models can be found in Appendix \ref{appC}.


\subsection{MethaneSAT}
For the MethaneSAT dataset, we observed similar performance trends across models as with MethaneAIR, though with some notable differences in the relative performance of spectral versus spatial approaches. Table \ref{tab:model_comparison_methanesat} presents the performance metrics for each model, again showing the Combined CNN approach achieving the highest scores across all metrics, with an accuracy of 81.96±1.45\% and F1-score of 78.80±1.28\%.

\begin{table}[htbp]
\centering
\begin{tabular}{|l|c|c|c|c| }
\hline
Model & Accuracy & F1 & Precision & Recall \\
\hline
ILR & 71.82±4.02 & 64.35±3.56 & 70.25±2.57 & 65.68±1.98 \\
MLP & 74.03±3.72 & 67.11±2.06 & 69.54±2.86 & 68.79±0.97 \\
U-Net & 78.73±3.23 & 68.56±0.36 & 67.87±0.26 & 71.90±1.76  \\
SCAN & 80.33±3.43 & 71.53±0.75 & 70.53±0.11 & 74.73±0.95 \\
Comb. MLP & 81.32±1.28 & 78.10±1.72 & 78.30±1.02 & 80.35±1.49\\
Comb. CNN & \textbf{81.96±1.45} & \textbf{78.80±1.28} & \textbf{78.85±0.86} & \textbf{81.09±1.23} \\
\hline
\end{tabular}
\caption{Performance metrics comparison across different models for MethaneSAT data.}
\label{tab:model_comparison_methanesat}
\end{table}

\begin{figure*}[h!]
  \centering  \includegraphics[width=.7\linewidth]{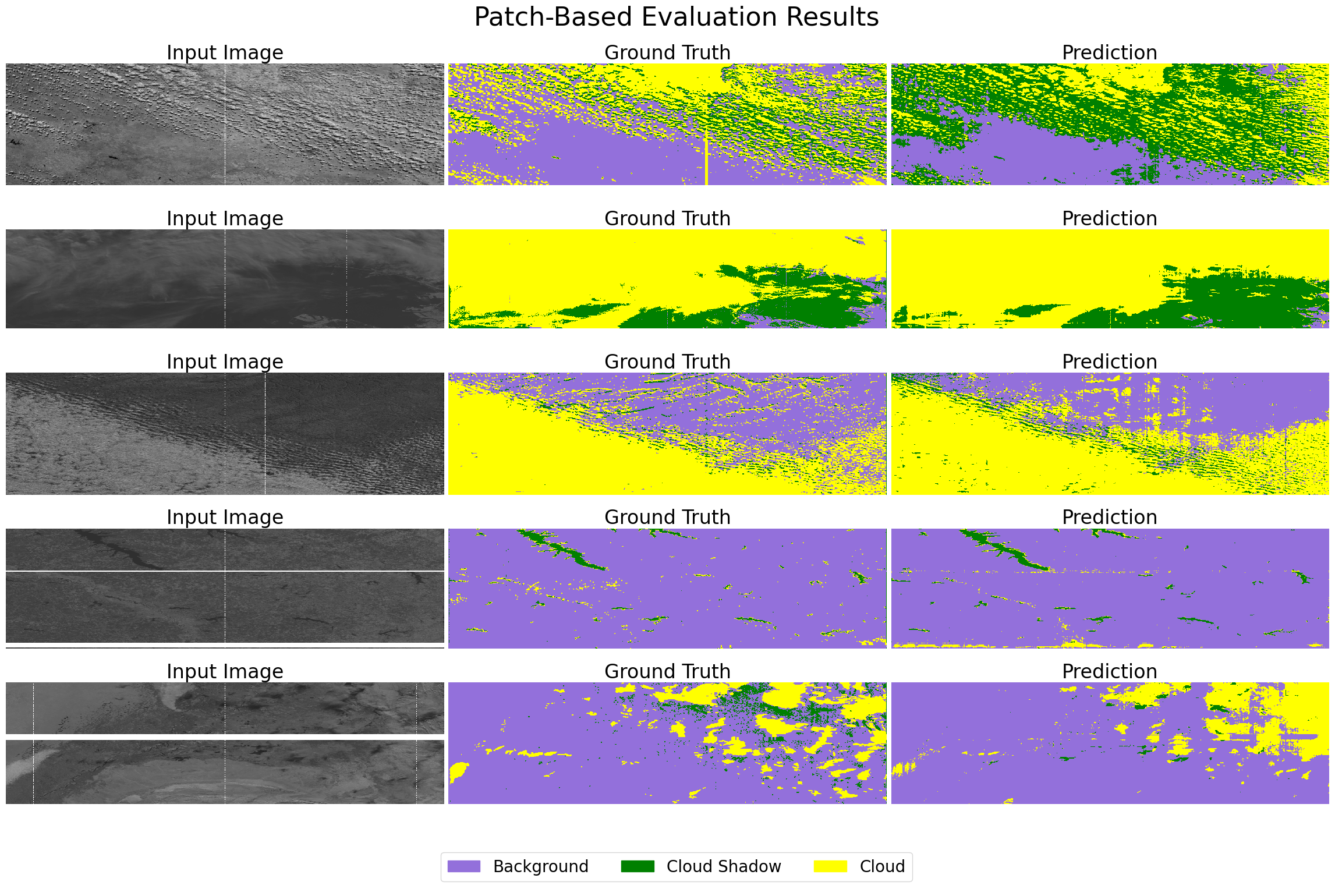}
  \caption{Examples of Combined CNN model predictions on MethaneSAT data showing input images (left), ground truth masks (middle), and model predictions (right) across diverse scenes. Results were computed over the test set using the first cross-validated fold.}
  \label{fig:preds_combined_cnn_msat}
\end{figure*}

Interestingly, for the MethaneSAT data, the SCAN model (accuracy: 80.33±3.43\%, F1-score: 71.53±0.75\%) outperformed the U-Net model (accuracy: 78.73±3.23\%, F1-score: 68.56±0.36\%), suggesting that spectral attention mechanisms may be particularly valuable for the unique spectral characteristics of MethaneSAT observations.  Figure \ref{fig:preds_combined_cnn_msat} shows representative examples of the Combined CNN model predictions across varying scenes, demonstrating its effectiveness in accurately identifying clouds and shadows across diverse terrain and atmospheric conditions. 

\begin{figure*}[h!]
\centering
\begin{subfigure}{0.288\textwidth}  
\includegraphics[width=\textwidth]{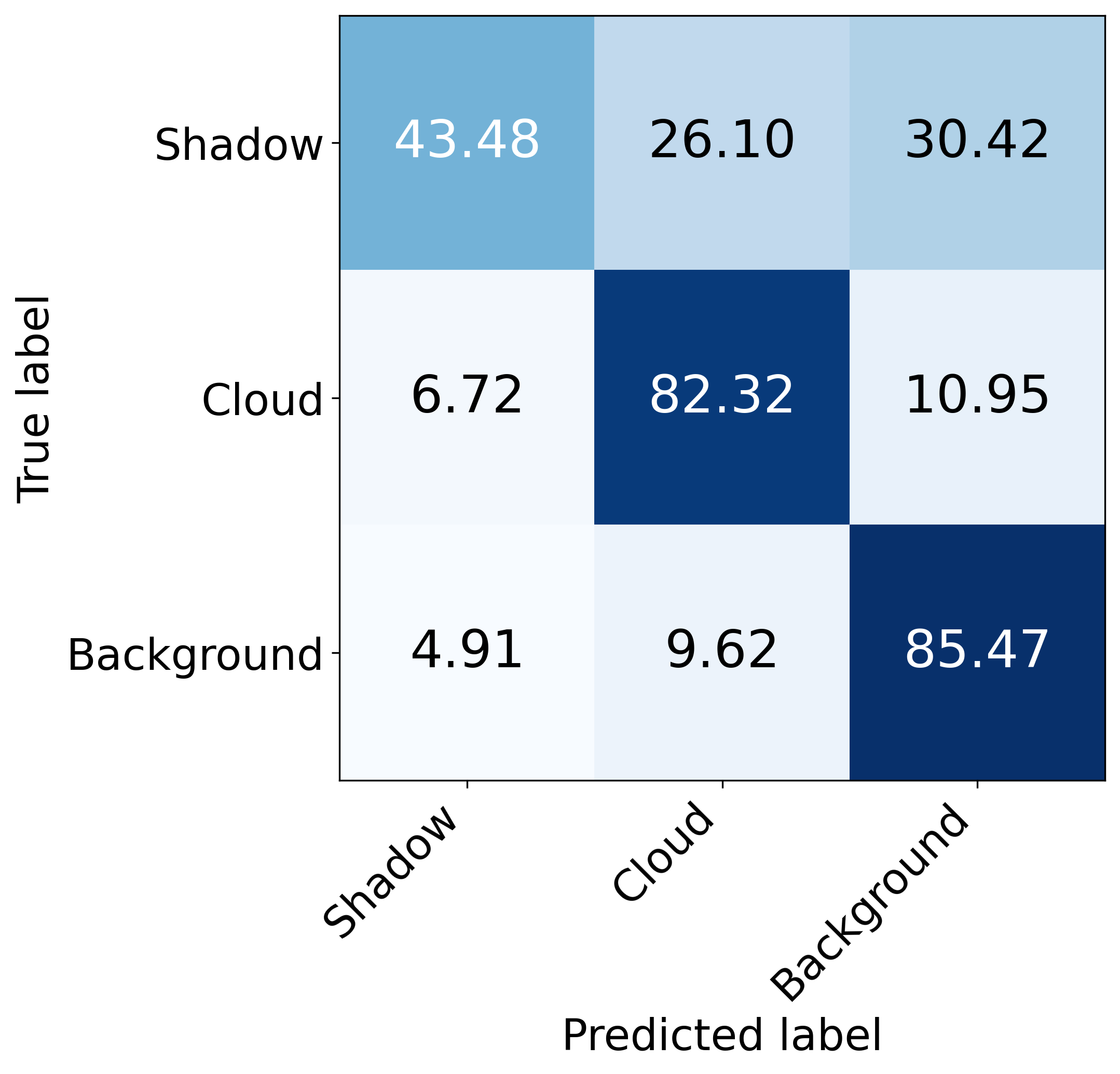}
\caption{ILR}
\end{subfigure}
\begin{subfigure}{0.3\textwidth}  
\includegraphics[width=\textwidth]{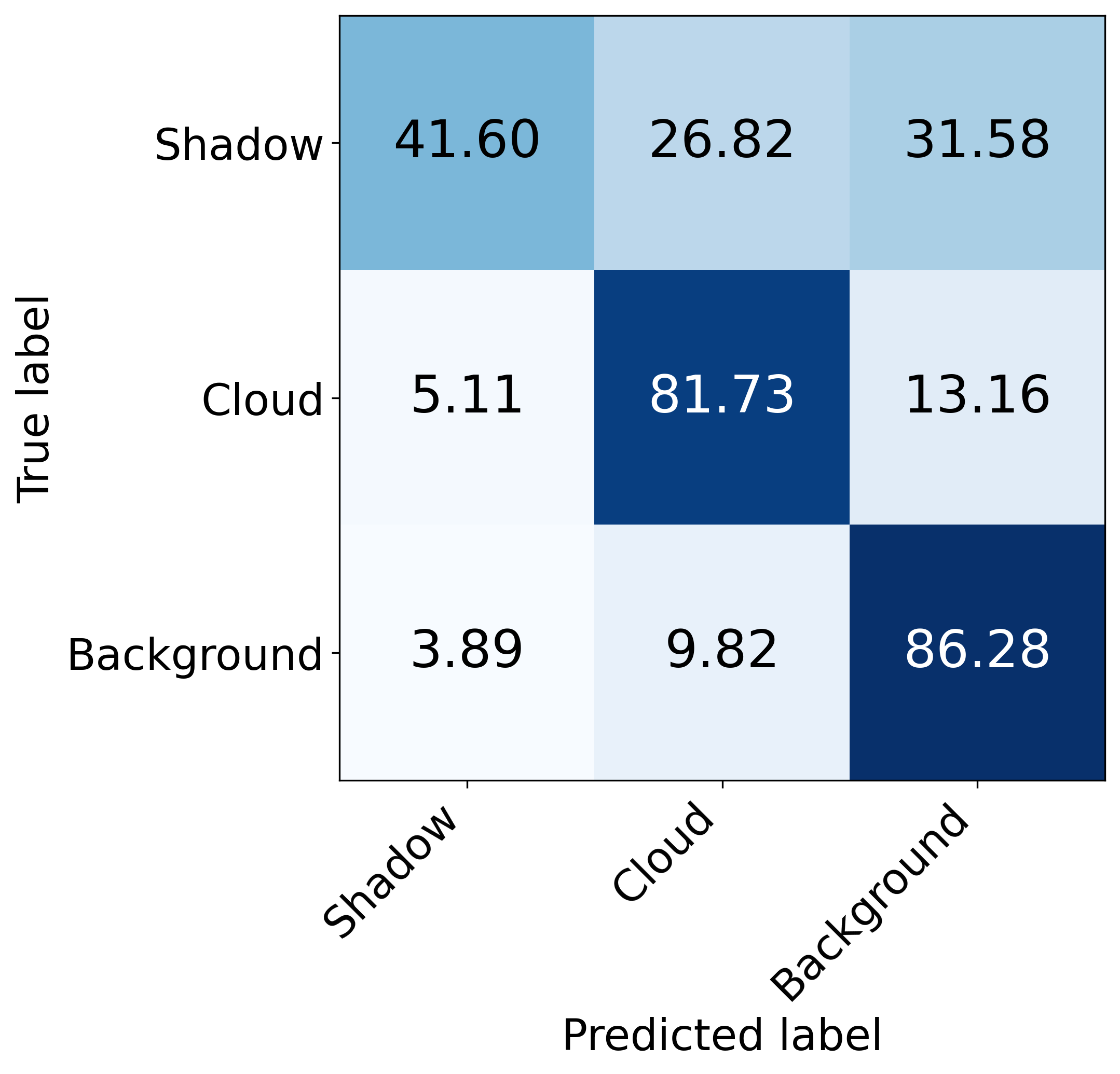}
\caption{MLP}
\end{subfigure}
\begin{subfigure}{0.3\textwidth}  
\includegraphics[width=\textwidth]{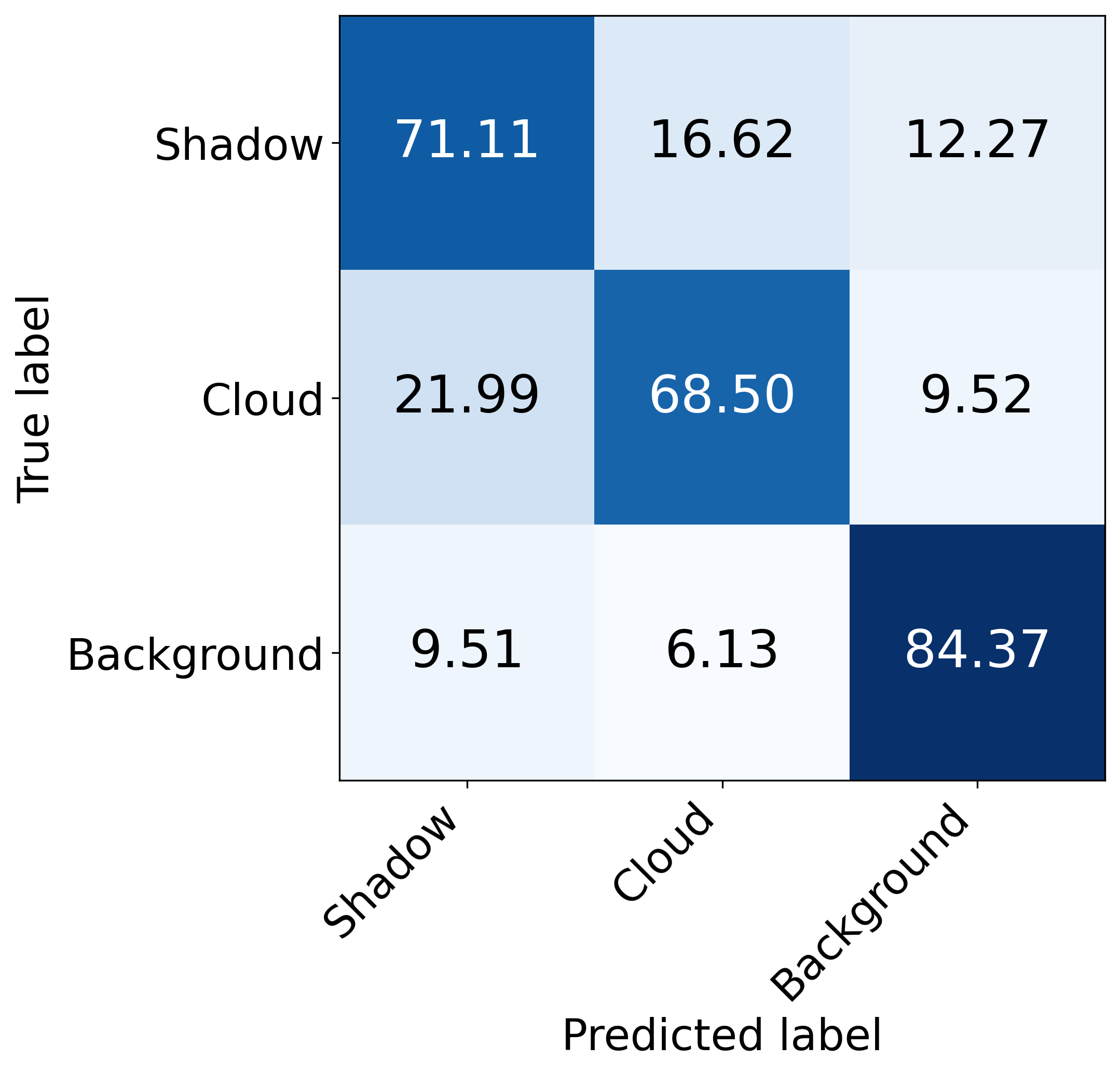}
\caption{U-Net}
\end{subfigure}
\begin{subfigure}{0.3\textwidth}  
\includegraphics[width=\textwidth]{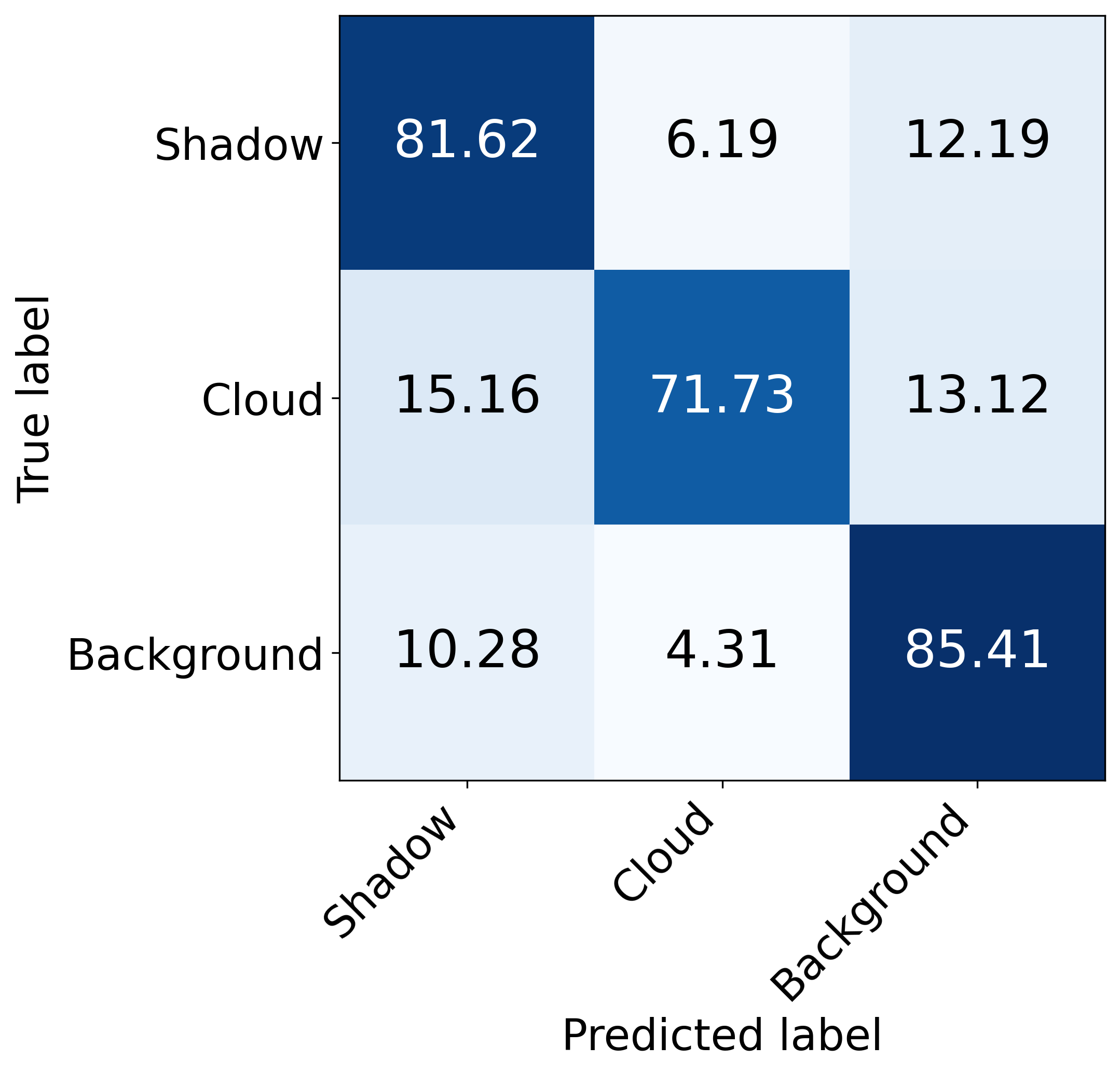}
\caption{SCAN}
\end{subfigure}
\begin{subfigure}{0.3\textwidth}  
\includegraphics[width=\textwidth]{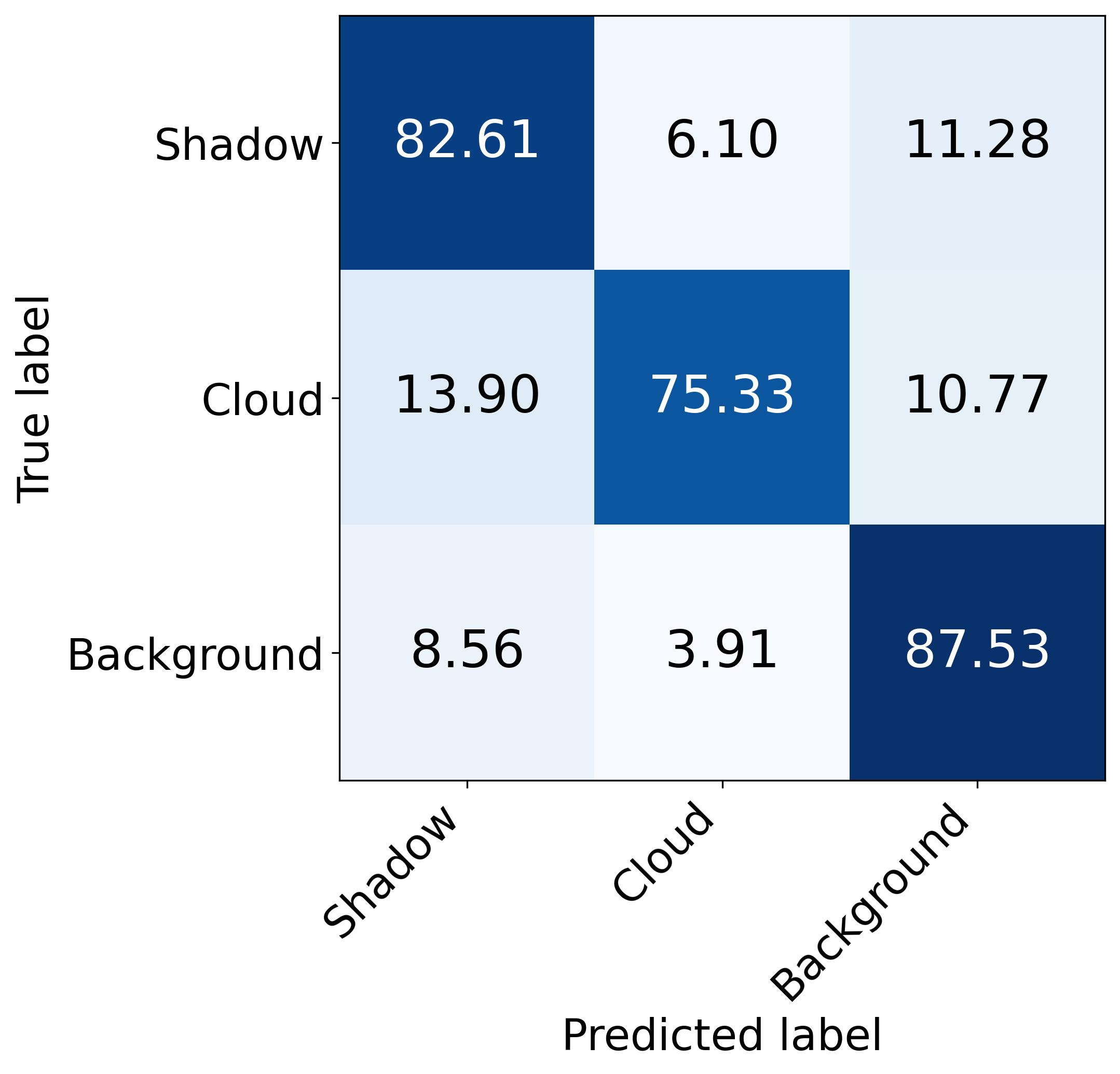}
\caption{Combined MLP}
\end{subfigure}
\begin{subfigure}{0.3\textwidth}  
\includegraphics[width=\textwidth]{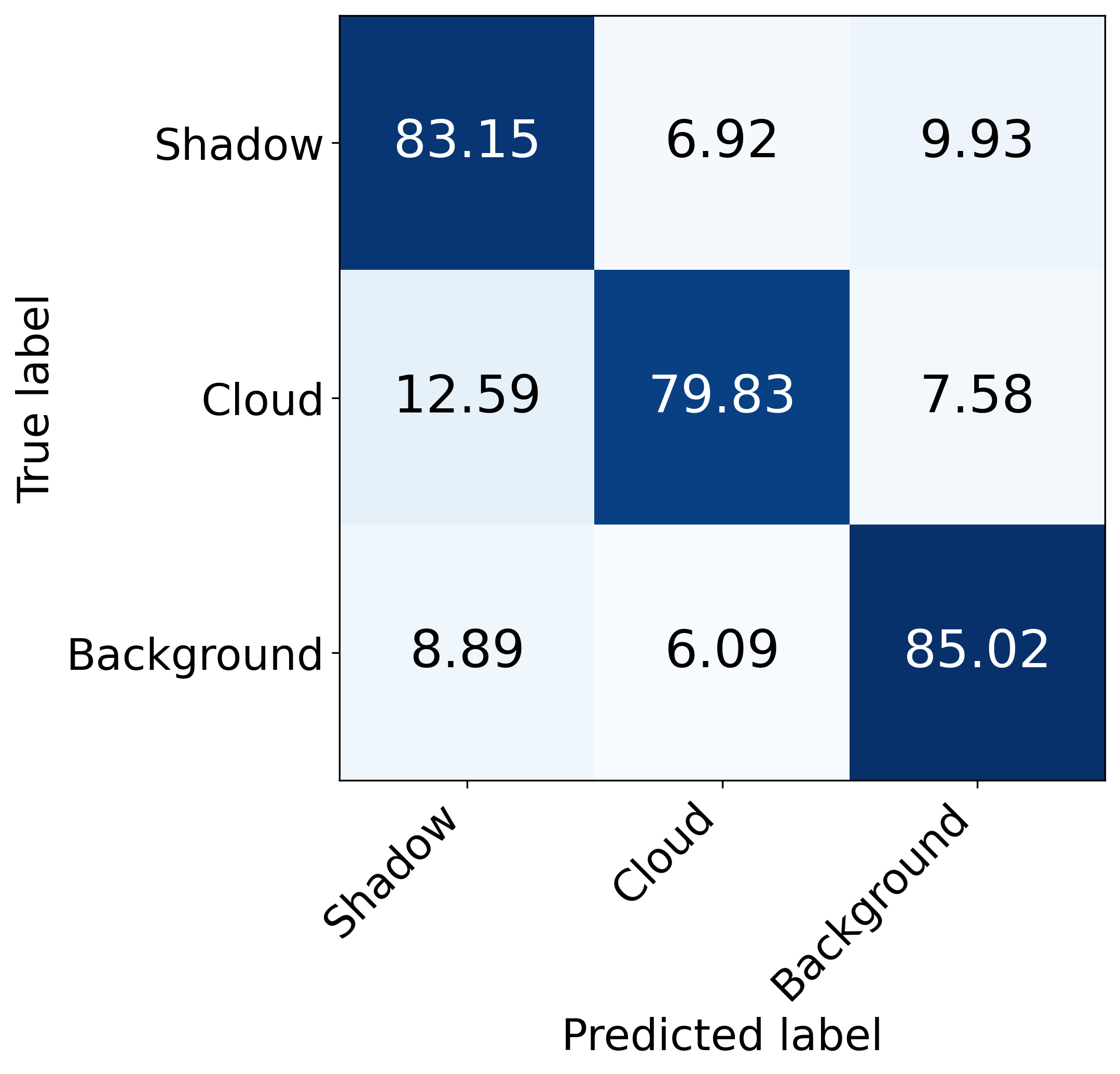}
\caption{Combined CNN}
\end{subfigure}
\caption{Confusion matrices for all evaluated models on MethaneSAT data. All matrices show results over the test set using the first cross-validated fold.}
\label{fig:conf_m_all_msat}
\end{figure*}

The confusion matrices for all models on the MethaneSAT dataset (Figure \ref{fig:conf_m_all_msat}) reveal distinctive classification challenges compared to MethaneAIR. Notably, all models demonstrate increased confusion between cloud and shadow classes, with even the Combined CNN showing 12.59\% misclassification of cloud spatial soundings as shadows. This pattern suggests greater spectral similarity between these features in the MethaneSAT data. The SCAN model (shadow accuracy: 81.62\%) outperforms the U-Net (shadow accuracy: 71.11\%) in shadow detection, reversing the trend observed in MethaneAIR data. This reversal highlights the value of spectral attention mechanisms for MethaneSAT spectral characteristics. The Combined MLP substantially improves cloud-shadow discrimination, while the Combined CNN achieves the highest overall performance with class-specific accuracies of 85.02\% for background, 79.83\% for clouds, and 83.15\% for shadows. 

\begin{figure*}[h!]
  \centering
  \includegraphics[width=.7\linewidth]{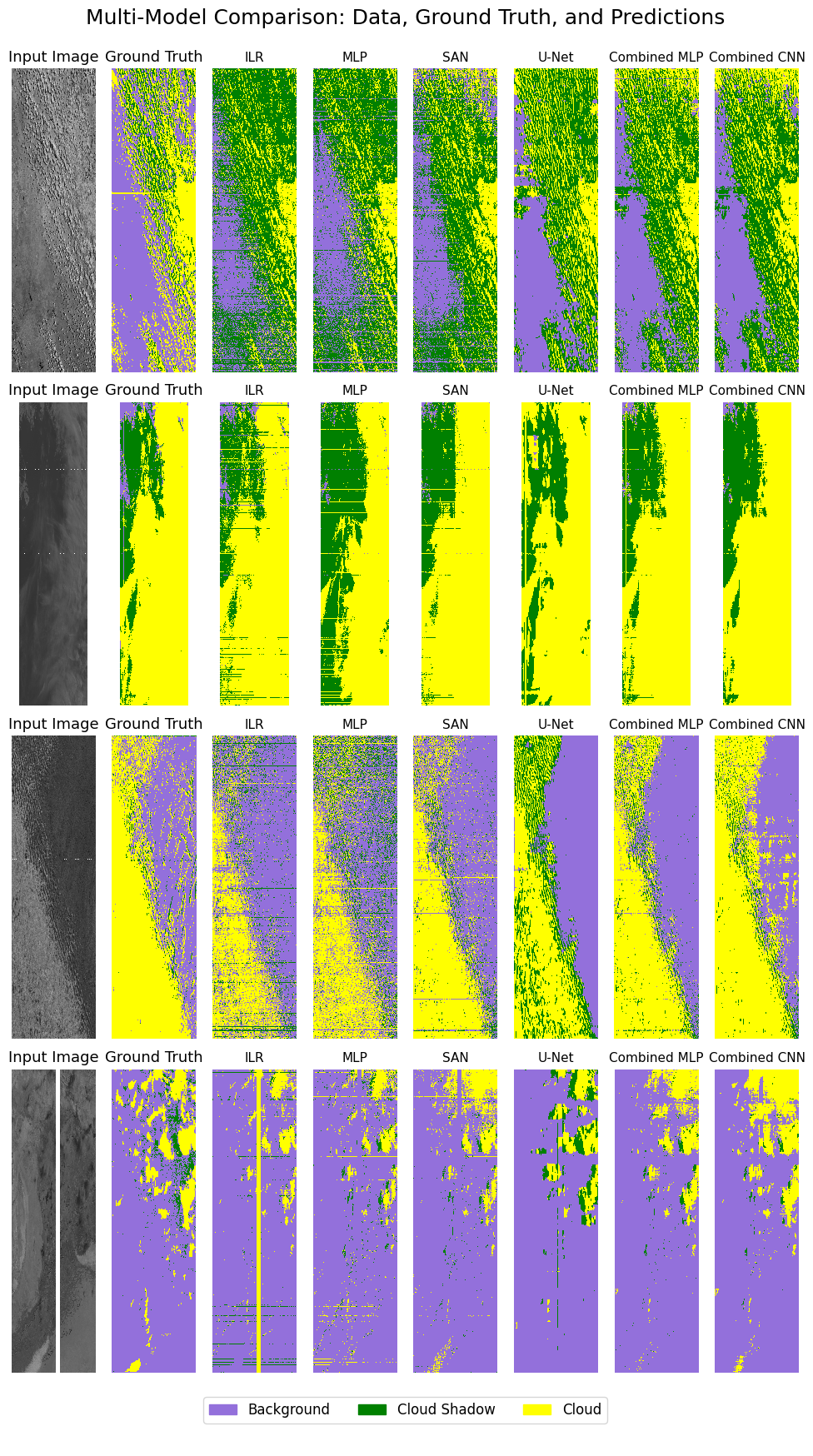}
  \caption{Prediction comparison across all evaluated models for representative MethaneSAT scenes. All the scenes were transposed for better visualization. Displayed images belong to the test set, and predictions were computed using a model trained with the first cross validated fold.}
  \label{fig:preds_overall_msat}
\end{figure*}

The cross-model comparison in Figure \ref{fig:preds_overall_msat} further validates the findings from the MethaneAIR analysis. The MLP model produces noisy, fragmented predictions with horizontal striping artifacts particularly visible in the third and fourth rows. The SCAN model demonstrates improved boundary detection but still exhibits noise in spectrally complex regions. The U-Net shows smoother predictions but as can be seen in the first and third row of Figure \ref{fig:preds_overall_msat}, it tends to overestimate shadows and create more smooth boundaries around the clouds. Both combined models show enhanced performance, with the Combined CNN approach yielding the most balanced results, preserving cloud structure detail while maintaining spatial coherence in shadow regions.

Similar to MethaneAIR, the selection of the most appropriate method for MethaneSAT data should be guided by the specific operational requirements and downstream atmospheric retrieval algorithms. Given MethaneSAT's role in quantitative methane monitoring, applications may benefit from U-Net's conservative approach to boundary delineation, particularly when subsequent atmospheric correction processes can accommodate smoothed cloud edges or when avoiding false cloud detections is more critical than capturing precise cloud boundaries.

\subsection{Computational Performance Analysis}

To provide a comprehensive understanding of the practical considerations for deploying these models in operational settings, we conducted a detailed evaluation of computational efficiency across all architectures. This analysis is particularly relevant for the MethaneSAT mission, which requires near real-time processing capabilities to support timely methane monitoring.

\subsubsection{Experimental Setup}

We conducted computational performance measurements on a workstation equipped with an NVIDIA RTX A6000 GPU (48GB VRAM) running CUDA 11.7. All experiments utilized PyTorch 1.12.0 with CUDA acceleration. For each model, we measured:

\begin{itemize}
   \item Parameters count (millions): Number of parameters to be fitted during training.
    \item Memory consumption during inference (MB): Memory consumption was evaluated by tracking peak GPU memory allocation during both training and inference processes using PyTorch's native memory profiling utilities. We recorded the maximum memory footprint across complete forward and backward passes to capture the full computational requirements.
    \item Training time per epoch (seconds): Training performance was assessed by measuring the time required to complete one full epoch across the MethaneSAT training dataset with consistent batch size (8 samples) and optimization parameters. These measurements were averaged over 10 complete training epochs to account for system variability. All models utilized identical loss functions and optimizer configurations.
    \item Inference time per area (ms/1,000 km²): For inference time measurements, each model processed identical input samples with dimensions matching the MethaneSAT dataset (samples with spatial resolution of $224 \times 224$, across 1080 spectral bands). We executed 100 consecutive forward passes for each model after 10 initial warm-up iterations to eliminate cold-start variability. Synchronization barriers were implemented between measurements to ensure accurate timing of GPU operations. Given the spatial resolution of $\sim100\times400$ m$^2$ (across-track × along-track), each $224 \times 224$ sample covers approximately 2,007 km$^2$, allowing conversion from per-image timing to per-area metrics.
\end{itemize}

Measurements were averaged over 100 runs for inference and 10 complete training epochs on the MethaneSAT dataset to ensure statistical robustness.

\subsubsection{Performance Results}

Table \ref{tab:computational_performance} presents the computational performance metrics across all evaluated models.

\begin{table}[ht]
\centering
\scalebox{0.85}{
\begin{tabular}{lcccc}
\toprule
Model & Parameters & Training Time & Inference Time \\
& (M) & (s/epoch) & (ms/1,000 km²) \\
\midrule
MLP & 0.022 & $245.4 \pm 1.0$ & $1.2 \pm 0.0$ \\
U-Net & 0.113 & $255.8 \pm 7.2$ & $2.1 \pm 0.0$  \\
SCAN & 0.168 & $290.7 \pm 24.2$ & $1.7 \pm 0.0$  \\
Combined MLP & 0.035  & $ 296.7 \pm $ 12.4 & $ 4.2 \pm 0.1 $ \\
Combined CNN & 0.026  & $ 326.6 \pm 10.3 $ & $ 4.1 \pm 0.0 $  \\
\bottomrule
\end{tabular}}
\caption{Computational Performance of the MLP Model}
\label{tab:computational_performance}
\end{table}

Our analysis reveals significant differences in computational efficiency across the evaluated models. The lightest model, MLP, demonstrates the fastest inference time (1.2 ms per 1,000 km²), making it well-suited for resource-constrained environments. However, this computational efficiency comes at the cost of significantly lower segmentation accuracy, as demonstrated in our performance evaluation.

The U-Net and SCAN models represent an effective middle ground, with moderate parameter counts (0.113M and 0.168M, respectively) and efficient inference times (2.1 ms and 1.7 ms per 1,000 km²). Despite having more parameters, the SCAN model achieves faster inference, likely due to its efficient spectral attention mechanism that reduces the computational overhead compared to U-Net's multi-scale convolutional operations.


The combined models offer an interesting trade-off. While they require approximately twice the inference time of their individual components (4.2 ms for Combined MLP and 4.1 ms per 1,000 km² for Combined CNN), their parameter counts and memory consumption do not scale linearly with the addition of two base models. This efficiency is achieved through parameter freezing of the base models during training, which allows the fusion components to remain lightweight. Notably, the Combined CNN is marginally faster than the Combined MLP during inference.

\section{Discussion}

Our comprehensive evaluation of cloud and shadow detection models for hyperspectral satellite imagery reveals critical insights for methane monitoring missions. The performance analysis across diverse model architectures demonstrates a clear progression of capabilities, from the baseline spectral-only approaches to ensemble methods that effectively combine spatial and spectral information.

The baseline ILR model (accuracy: 73.81±4.05\ for MethaneAIR) captures basic spectral features but generates highly fragmented predictions with significant noise. This purely spectral approach, while computationally efficient, fails to leverage the spatial relationships crucial for distinguishing between spectrally similar features like shadows and dark surfaces. The MLP model (accuracy: 82.49±2.24\%) shows improved performance but still suffers from noise and fragmentation due to its pixel-wise classification approach, as clearly visible in Figures \ref{fig:preds_overall_mair} and \ref{fig:preds_overall_msat}.

The U-Net architecture (accuracy: 88.26±0.45\% for MethaneAIR, 78.73±3.23\% for MethaneSAT) represents a substantial advancement, leveraging its encoder-decoder structure with skip connections to produce spatially coherent predictions. While U-Net exhibits a tendency toward over-smoothed boundaries around cloud formations and their shadows, this characteristic can be advantageous depending on the specific application requirements. On the other hand, our proposed SCAN model (accuracy: 86.51±2.90\% for MethaneAIR, 80.33±3.43\% for MethaneSAT) demonstrates superior ability in detecting accurate boundaries through its spectral attention mechanism, though it still exhibits some noise in spectrally complex regions. 

\mpc{Notably, SCAN outperforms U-Net on MethaneSAT despite lower MethaneAIR performance. We attribute this reversal to MethaneSAT's coarser spatial resolution ($\sim 100 \times 400$ m vs. $\sim 25$ m) creating more spectrally mixed boundary spatial soundings where wavelength-specific attention provides better discrimination than spatial convolution. The 6-7\% cross-dataset performance gap underscores generalization challenges between sensors due to spatial resolution differences, scene complexity variations, and platform-specific characteristics. To validate SCAN's design, we benchmarked against state-of-the-art attention mechanisms (Appendix \ref{appB}). SCAN achieves 8.88-10.5 percentage point F1 improvements over SE-UNet, ViT-SegFormer, and SpectralFormer, demonstrating that spectral attention outperforms complex self-attention architectures when training data is limited (262 scenes).}

The combined models represent the most promising approach, effectively addressing the limitations of individual architectures. The Combined CNN model achieves the highest performance across all metrics for both datasets (accuracy: 89.42±1.20\% for MethaneAIR, 82.99\% for MethaneSAT), demonstrating how preserving spatial context during the fusion process yields predictions that combine the detailed boundary sensitivity of SCAN with the spatial coherence of U-Net. This integrated approach proves particularly effective for challenging scenes with complex cloud formations and varying surface reflectance.

Despite models for MethaneAIR and MethaneSAT being each developed, trained, and tested independently, the performance gap in similar architectures (approximately 6-7\% in accuracy across models) underscores the challenges in generalizing algorithms performance between different hyperspectral sensors. This discrepancy likely stems from differences in spatial resolution, spectral characteristics, and the complexity of the scenes captured by each instrument. The reduced performance on MethaneSAT data is particularly evident in the confusion matrices, which show increased misclassification between clouds and shadows (12.59\% of cloud spatial soundings misclassified as shadows), reflecting the difficulty in distinguishing these spectrally similar features in certain lighting conditions.

From a computational perspective, our analysis reveals important trade-offs between model performance and efficiency. While the MLP offers the fastest inference time (1.2 ms), its limited accuracy makes it unsuitable for operational use. The Combined CNN, despite requiring approximately twice the inference time of individual models (4.1 ms), delivers substantially improved performance without proportional increases in memory consumption or parameter count. This computational efficiency, achieved through parameter freezing and efficient fusion architectures, makes the Combined CNN viable for operational deployment in satellite missions with reasonable computational constraints.

These findings have direct implications for the MethaneSAT mission, which requires accurate identification of atmospheric artifacts to ensure reliable methane retrievals. The superiority of the Combined CNN approach in handling complex scenes with varying surface reflectance is particularly relevant for global methane monitoring, where diverse terrain and atmospheric conditions can significantly impact detection accuracy. \mpc{Our methods should generalize to other hyperspectral missions facing similar atmospheric preprocessing challenges,  improving the reliability of methane plumes mapping and supporting global efforts to identify and quantify methane emissions across different platforms.}

\section{Conclusions}

This study addresses a critical challenge in atmospheric methane monitoring: the accurate detection of clouds and shadows in hyperspectral satellite imagery, which is essential for reliable retrieval of methane concentrations, especially when a sensor combines a wide swath, fine spatial resolution, and high precision, as achieved by MethaneSAT and MethaneAIR. These sensors are designed to provide holistic assessments of point sources, area sources, and regional totals, from a single image, and interference by clouds and by cloud and terrain shadows can introduce significant errors. Highly reliable cloud and shadow masking algorithms are critical for next-generation methane monitoring missions that aim to track and mitigate greenhouse gas emissions as part of global climate change efforts.

Our comprehensive evaluation of semantic segmentation models for MethaneSAT and MethaneAIR reveals a clear progression in performance from traditional spectral-based methods (ILR, MLP) to advanced deep learning architectures (U-Net, SCAN), with each approach demonstrating distinct strengths and limitations. While spectral-based methods lack spatial context, U-Net produces spatially coherent but over-smoothed predictions, and SCAN excels at boundary detection but exhibits noise in complex regions.
The Combined CNN model emerges as the superior approach, achieving the highest performance metrics for both MethaneAIR (accuracy: 89.42±1.20\%, F1-score: 78.50±3.08\%) and MethaneSAT data (accuracy: 82.99\%, F1-score: 80.45\%). By preserving spatial context during fusion, this model successfully integrates the precise boundary detection capabilities of spectral attention mechanisms with the spatial coherence of convolutional architectures. This is the critical element needed to attain performance that spans over wide areas down to individual facilities. The performance differences between datasets highlight the importance of tailoring algorithmic approaches to specific sensor characteristics and use cases, with SCAN notably outperforming U-Net on MethaneSAT data despite showing slightly lower performance on MethaneAIR.

From a computational perspective, our analysis demonstrates that the Combined CNN model offers an effective balance between performance and efficiency. While requiring moderately increased computational resources compared to individual models (4.1 ms inference time versus 1.7-2.1 ms per 1,000 km²), its significantly improved segmentation accuracy justifies this overhead for operational deployment in satellite missions where accurate atmospheric artifact detection is critical for reliable methane retrievals and subsequent emission quantification. Furthermore, the patch-based evaluation strategy demonstrated for MethaneSAT's variable-sized inputs enhances the practical applicability of these methods in operational contexts. 

\mpc{Our findings have direct and urgent implications for global methane monitoring and climate change mitigation efforts. Accurate cloud and shadow masking is not merely a technical preprocessing: it is a critical prerequisite for reliable atmospheric CH$_4$ retrievals that directly impact our ability to track progress toward international commitments in addressing climate change. Furthermore, our results offer immediately applicable insights and methodologies to other operational and planned hyperspectral missions facing similar atmospheric challenges in remote sensing-based greenhouse gas monitoring, amplifying the impact of this work across the growing constellation of methane monitoring capabilities.}

Future research should focus on developing more efficient fusion architectures, and investigating transfer learning techniques to improve the model's performance. The Combined CNN approach represents a significant advancement in hyperspectral image segmentation for atmospheric artifact detection, balancing high performance with reasonable computational requirements for operational satellite-based methane monitoring.


\section{Acknowledgments}
\mpc{Funding for MethaneSAT and MethaneAIR activities was provided in part by Anonymous, Arnold Ventures, The Audacious Project, Ballmer Group, Bezos Earth Fund, The Children’s Investment Fund Foundation, Heising-Simons Family Fund, King Philanthropies, Robertson Foundation, Skyline Foundation and Valhalla Foundation. For a more complete list of funders, please visit \url{www.methanesat.org}. We express our gratitude to the MethaneAIR field team for their contributions to MethaneAIR data collection. We thank the AstroAI and EarthAI institutes at the Center for Astrophyiscs $|$ Harvard \& Smithsonian for useful discussions and guidance. CG and MPC were supported by AstroAI at the Center for Astrophysics $|$ Harvard and Smithsonian.}

\bibliographystyle{unsrt}
\bibliography{references}


\vskip -2\baselineskip plus -1fil
\begin{IEEEbiography}[{\includegraphics[width=1in,height=1.25in,clip,keepaspectratio]{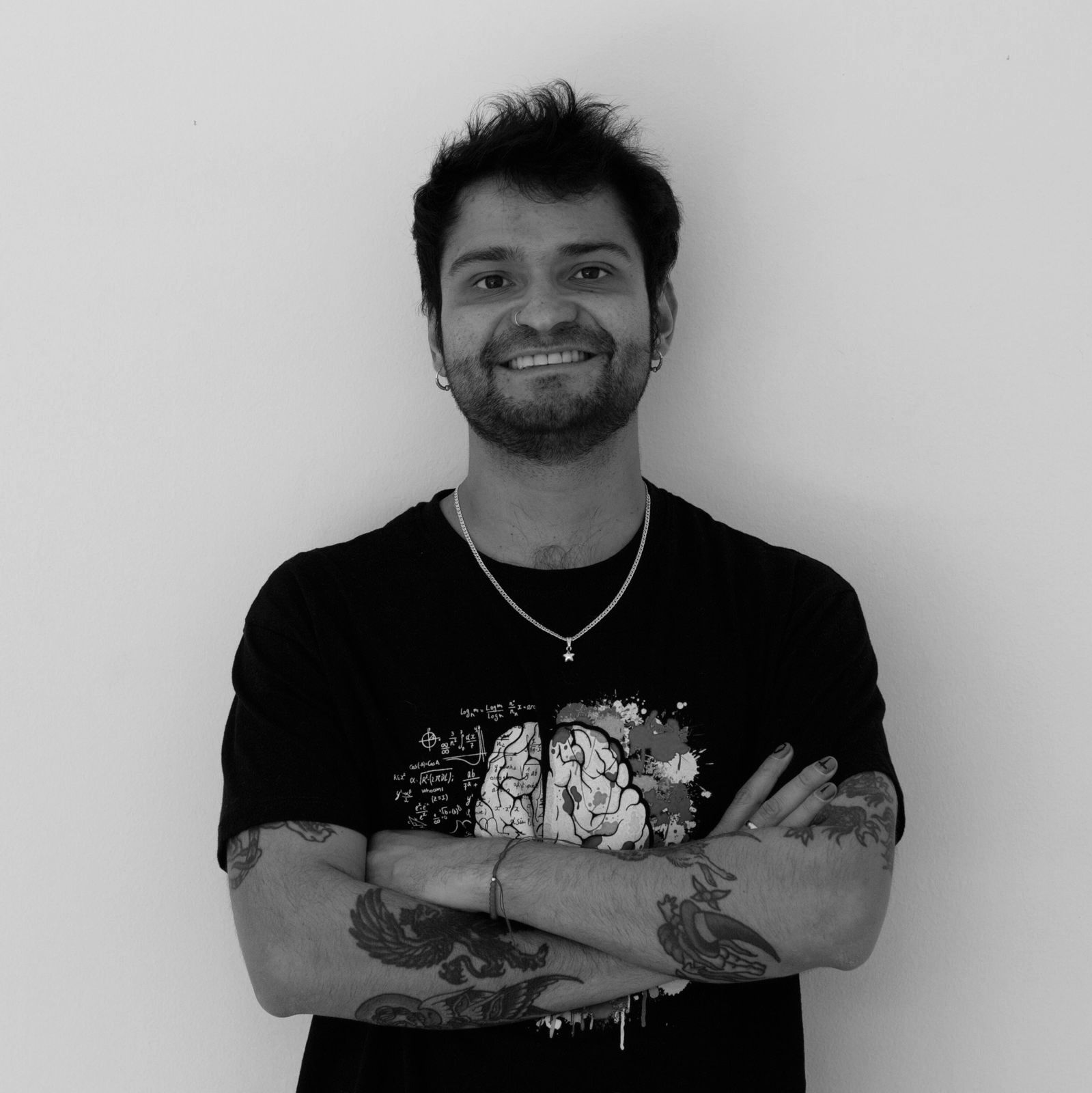}}]{Manuel Pérez-Carrasco}
received the Master's degree in Computer Science from the University of Concepción, Chile. He works on machine learning models for remote sensing and environmental monitoring.
\end{IEEEbiography}

\vskip -2\baselineskip plus -1fil
\begin{IEEEbiography}[{\includegraphics[width=1in,height=1.25in,clip,keepaspectratio]{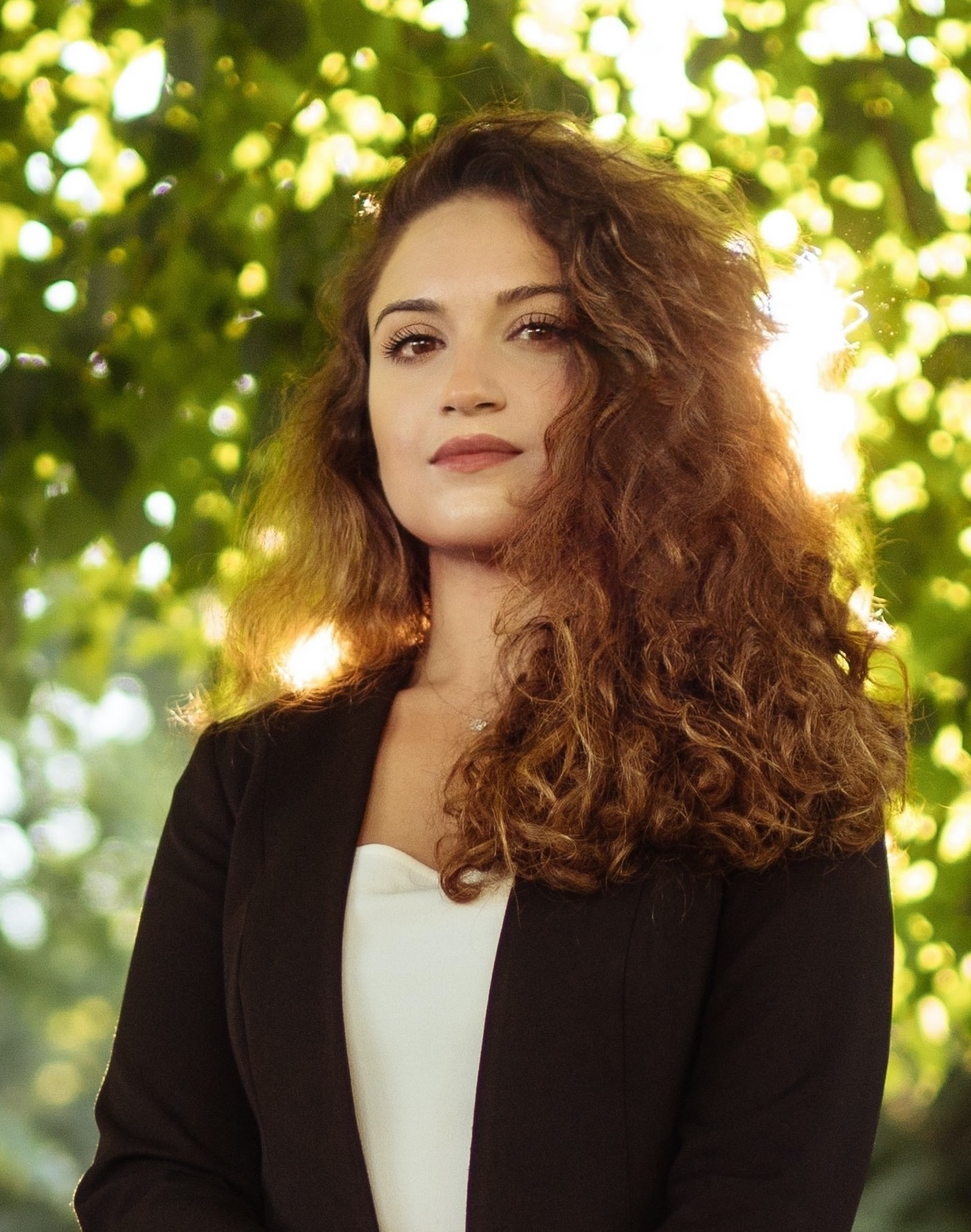}}]{Maya Nasr}
is a Science Engineer at Environmental Defense Fund and Harvard University, where she leads in-orbit lunar calibrations and the team developing machine learning models for the MethaneSAT mission. She brings expertise in space operations, mission planning, systems engineering, technology strategy, and space law and policy. She previously worked on space projects including NASA's Mars 2020 Perseverance rover mission, Cassini's mission activity on Titan, and the OneWeb satellites network. She holds Bachelor's, Master's, and Ph.D. degrees in Aerospace Engineering from MIT.
\end{IEEEbiography}

\vskip -2\baselineskip plus -1fil
\begin{IEEEbiography}[{\includegraphics[width=1in,height=1.25in,clip,keepaspectratio]{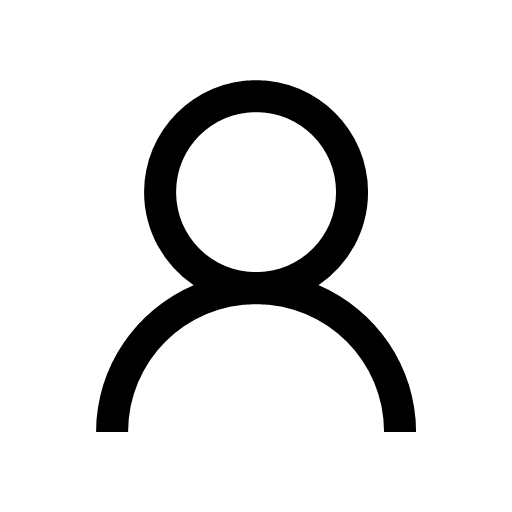}}]{Sébastien Roche}
is a Scientist at the Environmental Defense Fund and Harvard University where he contributes to level2 algorithm development and data analysis for MethaneSAT and MethaneAIR. His research focuses on improving greenhouse gas measurements from ground-, aircraft-, and spacecraft-based instruments with contributions to TCCON, MethaneAIR, MethaneSAT and the proposed Arctic Observing Mission.
\end{IEEEbiography}

\vskip -2\baselineskip plus -1fil
\begin{IEEEbiography}[{\includegraphics[width=1in,height=1.25in,clip,keepaspectratio]{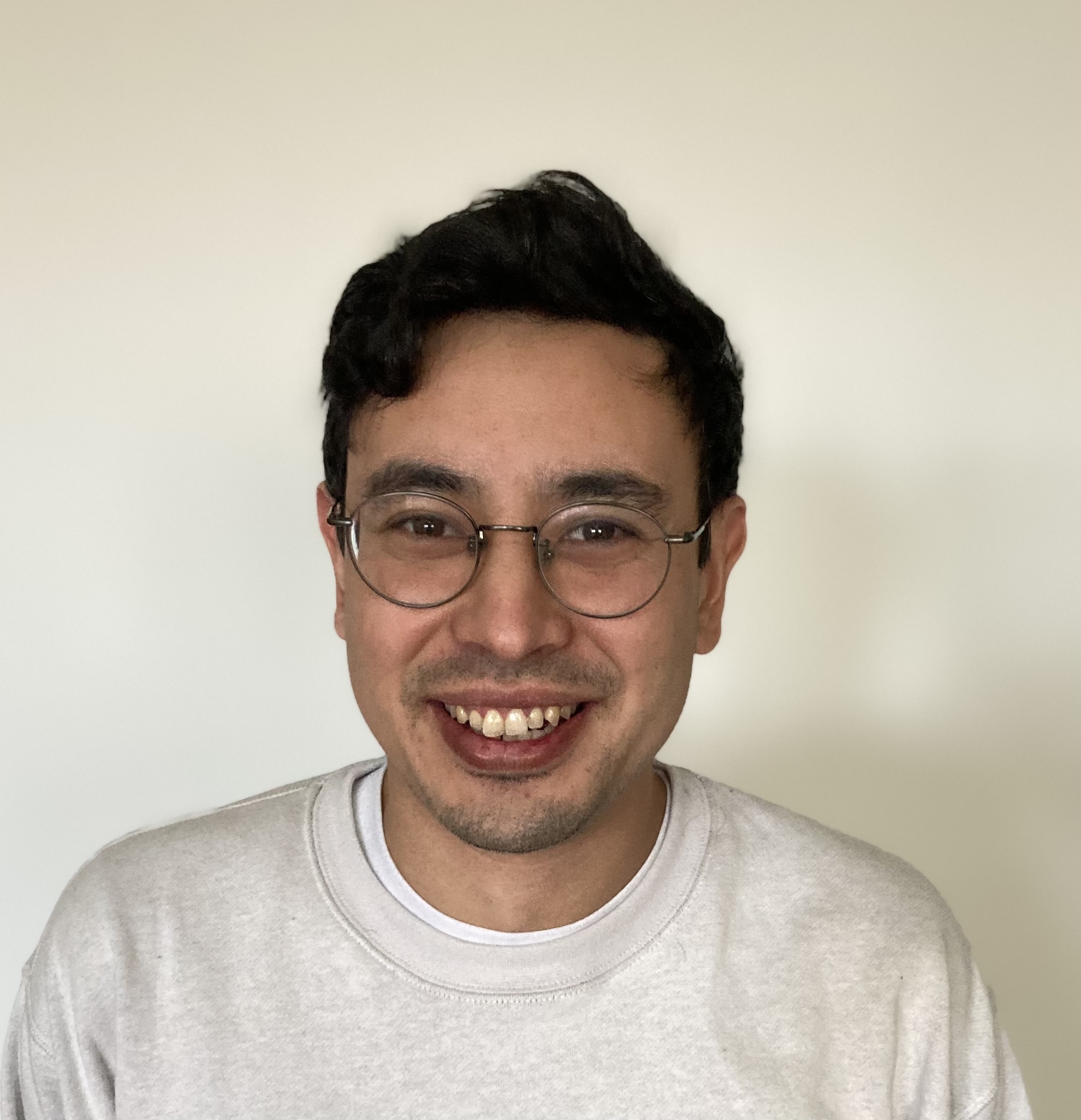}}]{Chris Chan Miller}
Christopher Chan Miller received the Ph.D. in Earth and Planetary Sciences from Harvard University in 2016. 
He is a Senior Scientist at the Environmental Defense Fund, where he leads level2 algorithm development for the MethaneSAT mission. 
His research focuses on atmospheric chemistry, satellite remote sensing, and radiative transfer, with contributions to MethaneSAT, TEMPO, and OMI satellite missions.
\end{IEEEbiography}

\vskip -2\baselineskip plus -1fil
\begin{IEEEbiography}[{\includegraphics[width=1in,height=1.25in,clip,keepaspectratio]{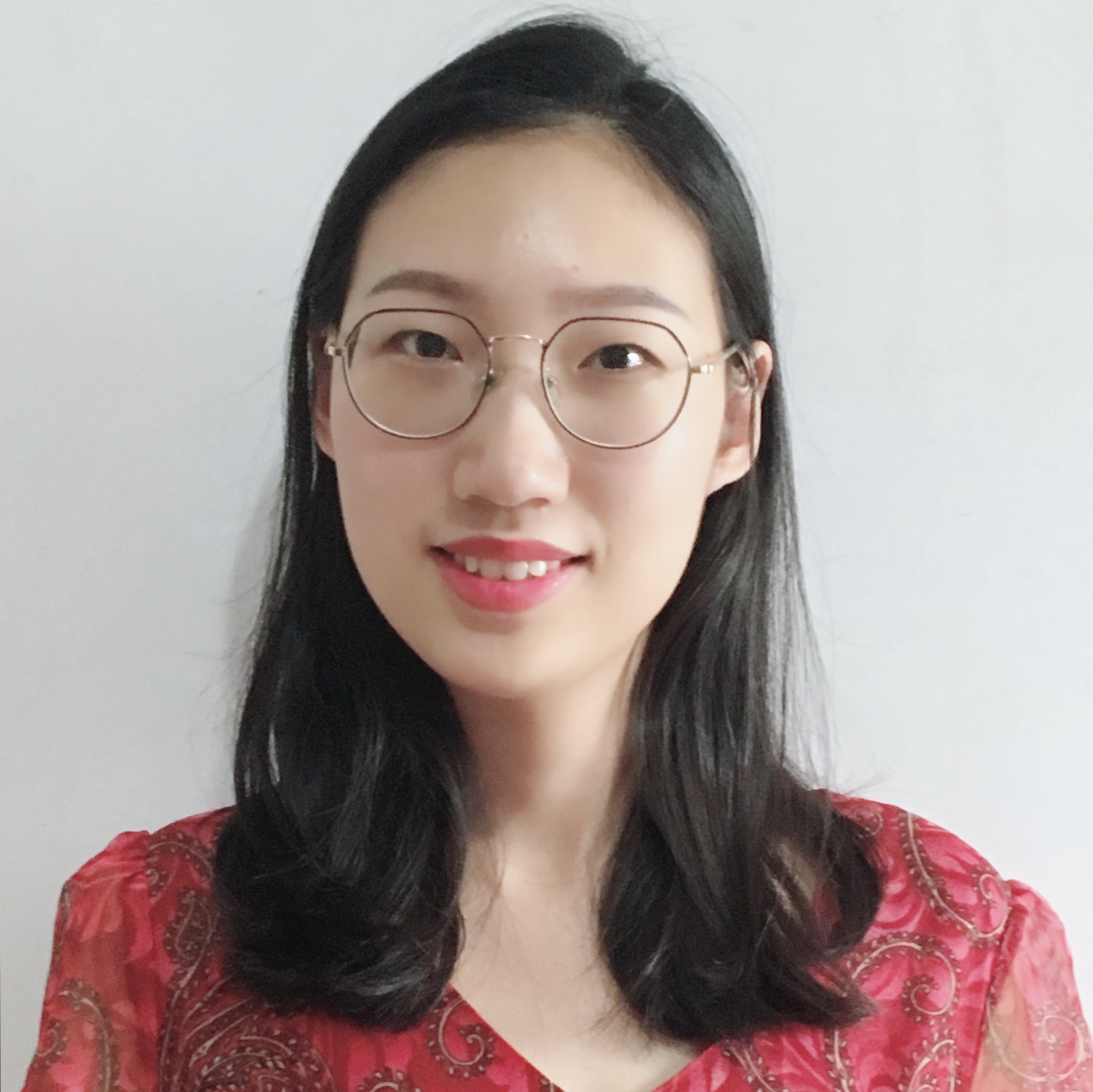}}]{Zhan Zhang}
received the B.S. degree in Environmental Science from Beijing Normal University, China, in 2016, and the M.S. degree in Ecology from Chinese Academy of Sciences, China, in 2019, and the Ph.D. degree in Energy Resources Engineering from Stanford University, USA, in 2024.
From 2024 to 2025, she served as the postdoctoral research fellow in Harvard University. Since 2025, she works as the Scientist, MethaneSAT Science in Environmental Defense Fund, where she mainly works on remote sensing methane emissions detection using MethaneSAT and MethaneAIR. Her research interest lies in remote sensing greenhouse gas emissions detection based on various satellite and airborne platforms.
Dr. Zhang has authored and coauthored more than 10 publications in high-impact international journals, including Geophysical Research Letters and Nature. She also serves as a reviewer for high-ranking journals including Remote Sensing of Environment and Environmental Science \& Technology.
\end{IEEEbiography}

\vskip -2\baselineskip plus -1fil
\begin{IEEEbiography}[{\includegraphics[width=1in,height=1.25in,clip,keepaspectratio]{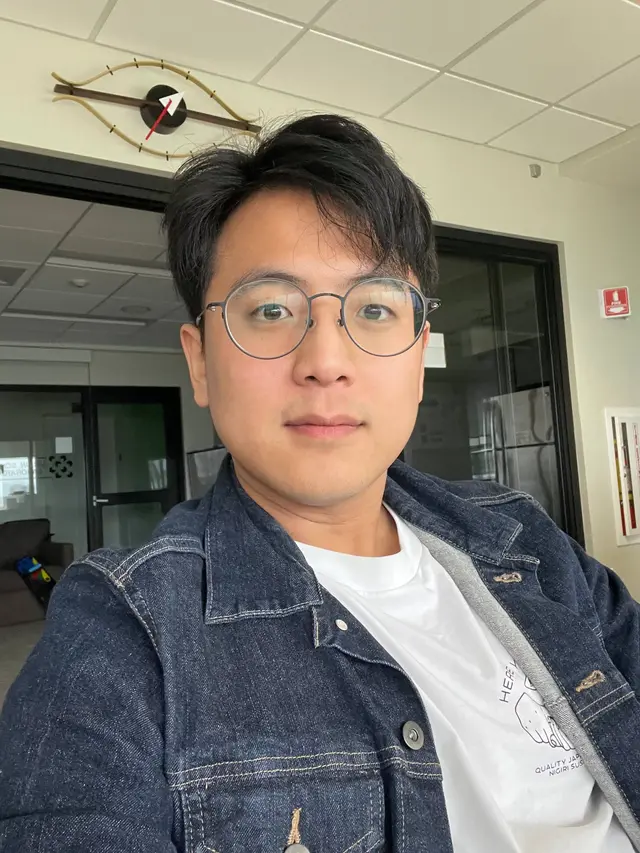}}]{Core Francisco Park}
received the Ph.D. degree in Physics from Harvard University, Cambridge, MA, USA. He has worked on AI for science and the science of AI, and his research interests include artificial intelligence, open-endedness, and human–AI interaction.
\end{IEEEbiography}

\vskip -2\baselineskip plus -1fil
\begin{IEEEbiography}[{\includegraphics[width=1in,height=1.25in,clip,keepaspectratio]{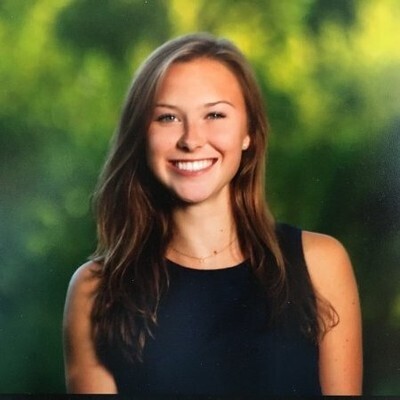}}]{Eleanor Walker}
received the B.S. degree in chemical engineering from the California Institute Of Technology in 2020. In 2022 she received the M.Sc in Atmosphere/Energy in the department of Civil and Environmental Engineering at Stanford University, where she was involved in research on estimating methane emissions from drainage canals in tropical peatlands. She is currently pursuing the Ph.D. degree in Environmental Science and Engineering at Harvard University, focusing on improving the retrieval and quantification algorithms for methane and carbon dioxide with MethaneSAT.
\end{IEEEbiography}

\vskip -2\baselineskip plus -1fil
\begin{IEEEbiography}[{\includegraphics[width=1in,height=1.25in,clip,keepaspectratio]{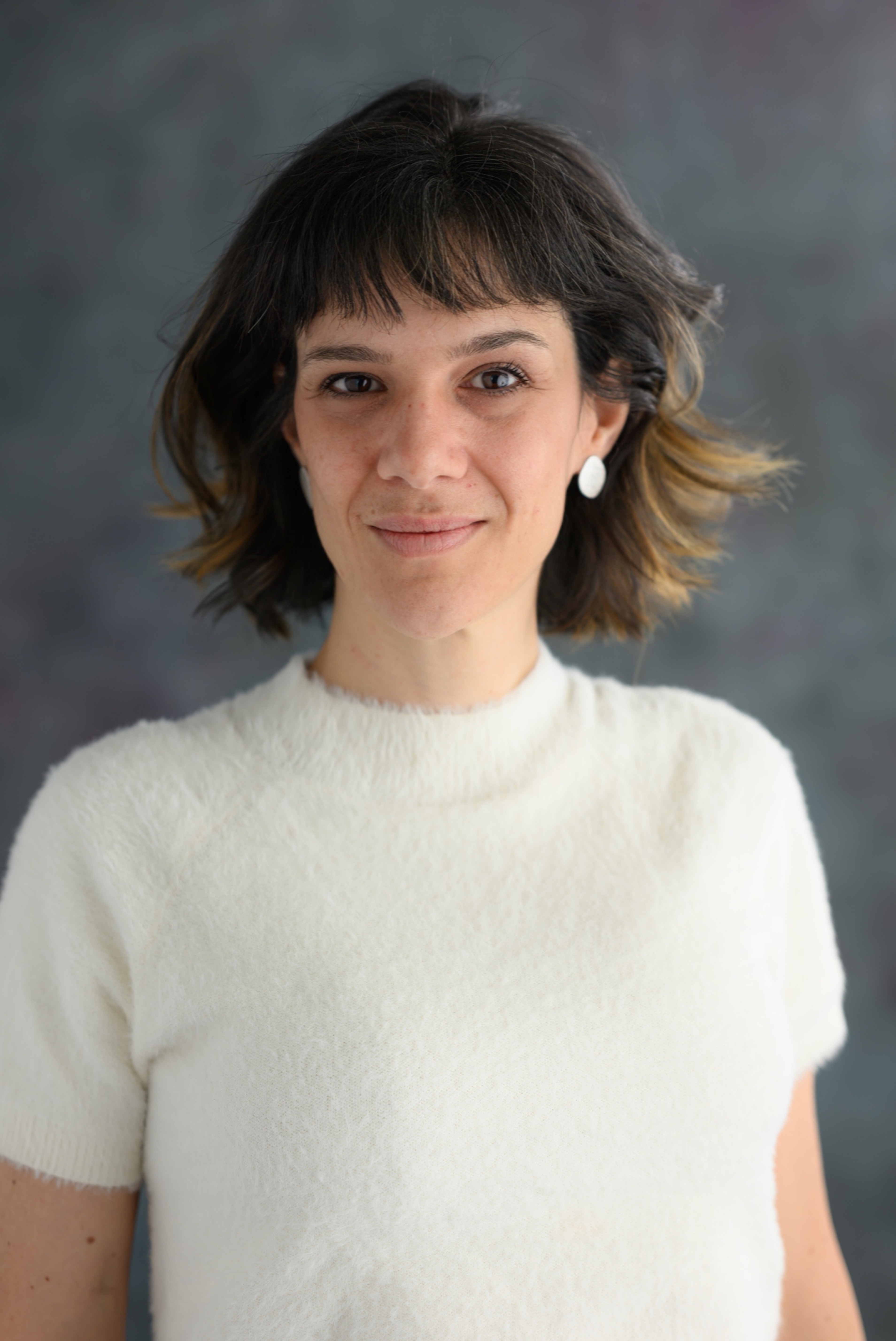}}]{Cecilia Garraffo}
Dr. Cecilia Garraffo is the founder and director of the AstroAI and the EarthAI Institutes at the Center for Astrophysics $\vert$ Harvard \& Smithsonian. She received the Presidential Early Career Award for Scientists and Engineers (PECASE) in 2024, the highest U.S. honor for early-career scientists. Originally from Argentina, she earned an MS in Astronomy, a PhD in Physics, and completed postdoctoral fellowships in AI and astrophysics. Her research spans AI-driven cosmology and the search for life in exoplanet atmospheres.
\end{IEEEbiography}

\vskip -2\baselineskip plus -1fil
\begin{IEEEbiography}[{\includegraphics[width=1in,height=1.25in,clip,keepaspectratio]{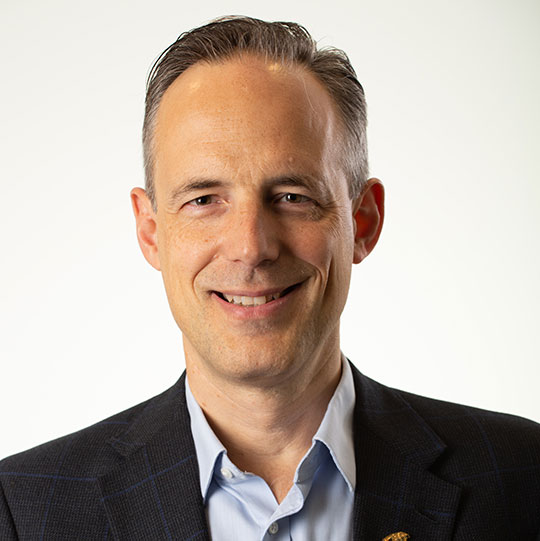}}]{Douglas Finkbeiner}
received the B.S. degree in physics and German from the University of Michigan, Ann Arbor, in 1994, and Ph.D. in physics from the University of California, Berkeley in 1999. From 1999 to 2001, he was a NASA ADP Postdoctoral Researcher at UC Berkeley, and from 2001 to 2006, he was a Hubble Fellow and Russell–Cotsen Fellow in the Department of Astrophysical Sciences at Princeton University. Since 2006, he has been with Harvard University, where he is a Professor with joint appointments in the Departments of Astronomy and Physics. His research spans Galactic structure and interstellar dust, large-scale imaging and spectroscopic surveys, high-energy astrophysics, and machine learning methods for astronomical data, with applications across multiwavelength remote sensing of the Milky Way. Dr. Finkbeiner is a Fellow of the American Physical Society and the American Astronomical Society, and a co-recipient of the 2014 Bruno Rossi Prize. 
\end{IEEEbiography}

\vskip -2\baselineskip plus -1fil
\begin{IEEEbiography}[{\includegraphics[width=1in,height=1.25in,clip,keepaspectratio]{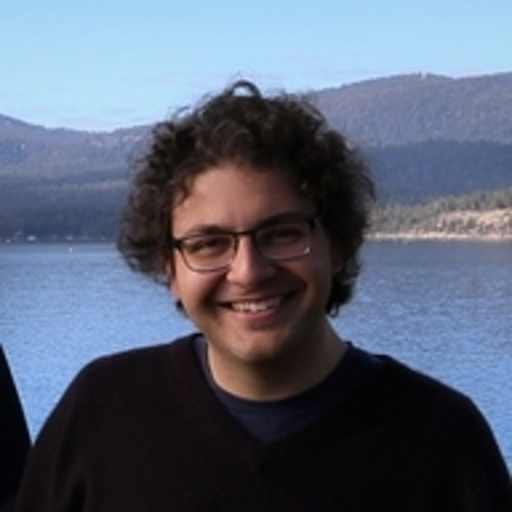}}]{Sasha Ayvazov} is a senior algorithm engineer at MethaneSAT.
\end{IEEEbiography}

\vskip -2\baselineskip plus -1fil
\begin{IEEEbiography}[{\includegraphics[width=1in,height=1.2in,clip,keepaspectratio]{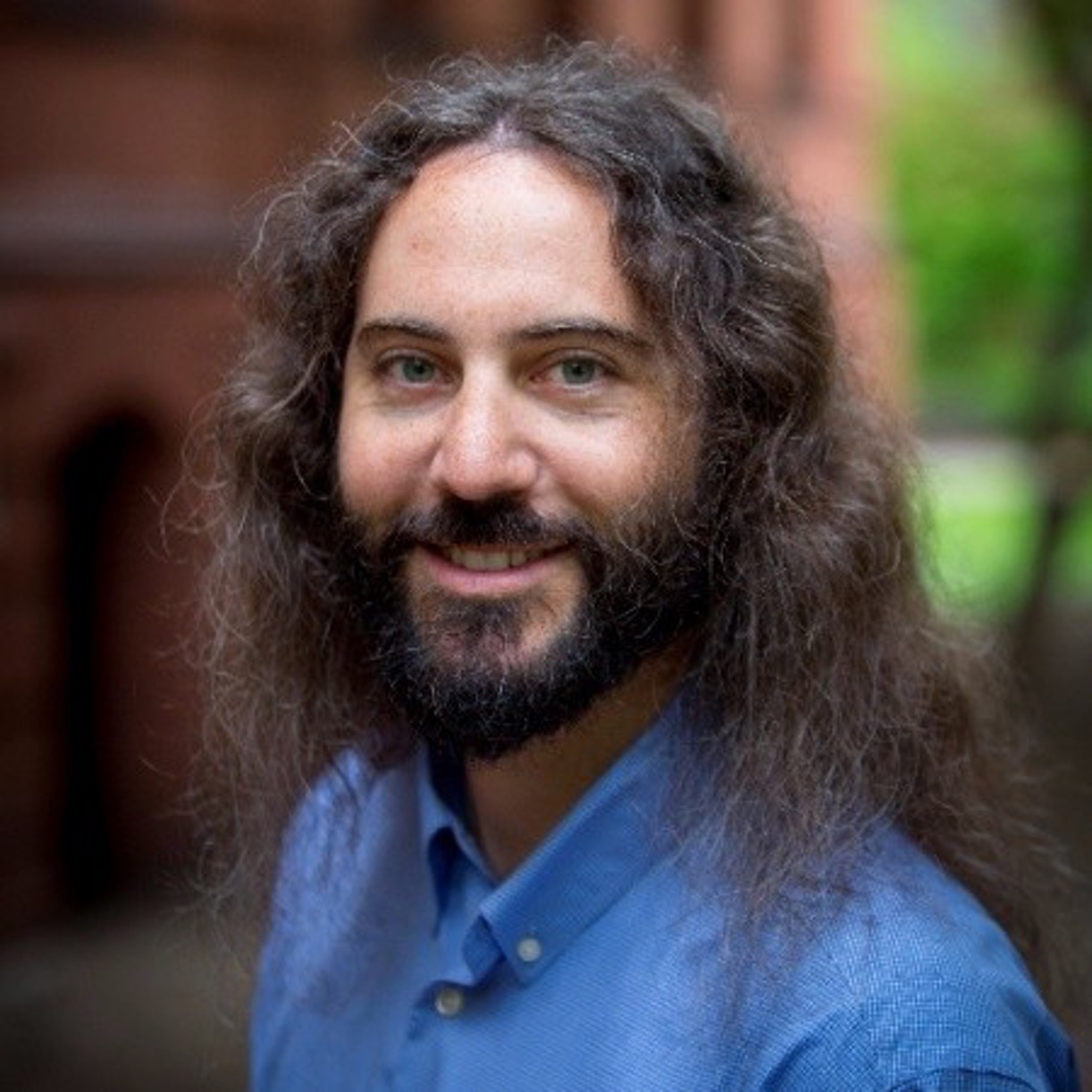}}]{Jonathan E. Franklin} (BS, Marlboro C., 2003; MSc, U. Massachusetts Amherst, 2005; PhD, Dalhousie U., 2015) is a Senior Project Scientist at Harvard University. His scientific work focuses on remote sensing measurements of greenhouse gases from ground, airborne, and satellite platforms. He is the calibration lead for the MethaneSAT and MethaneAIR missions.
\end{IEEEbiography}

\vskip -2\baselineskip plus -1fil
\begin{IEEEbiography}[{\includegraphics[width=1in,height=1.2in,clip,keepaspectratio]{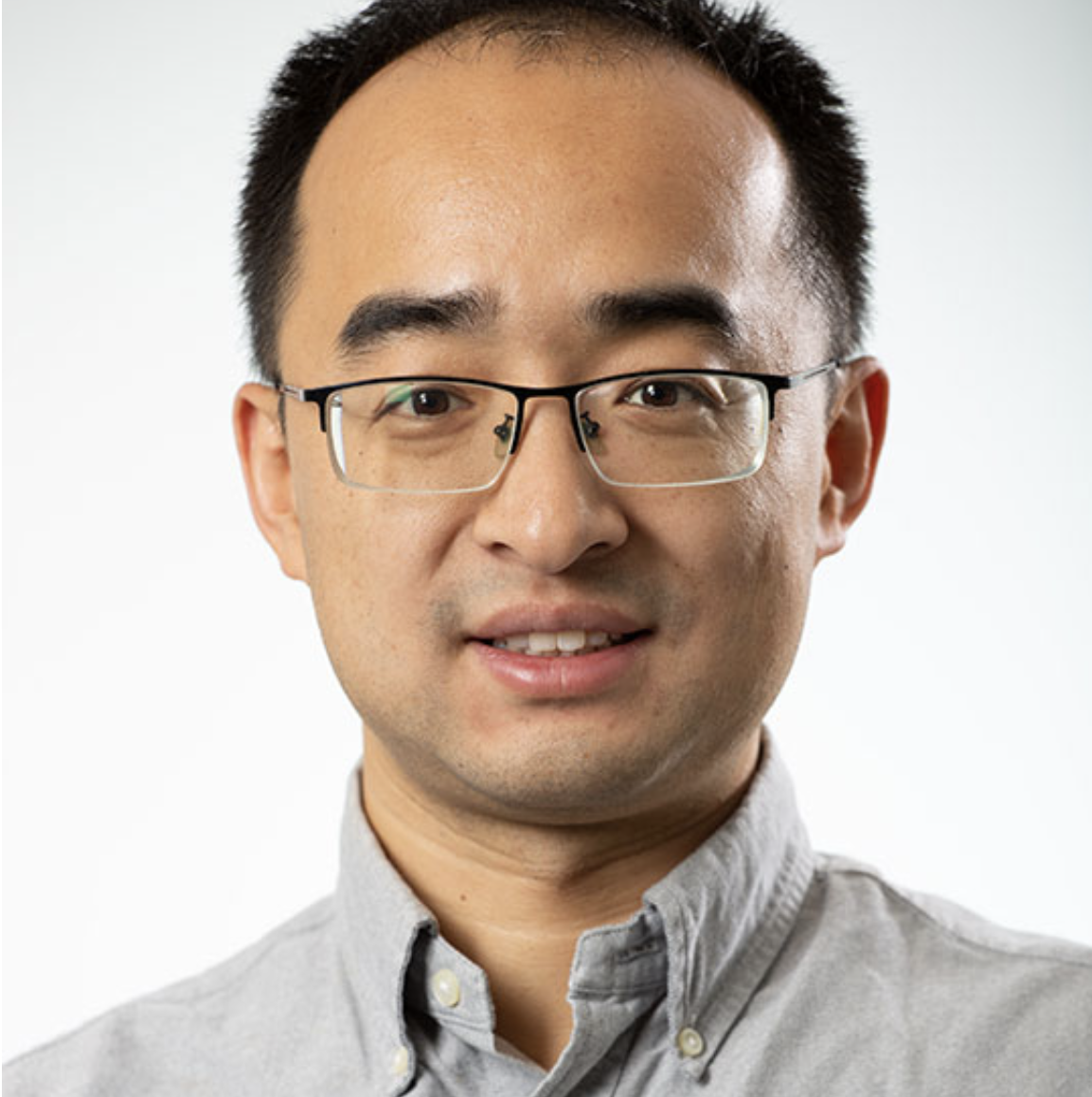}}]{Bingkun Luo}
 is with the Atomic and Molecular Physics Division, Center for Astrophysics $\vert$ Harvard \& Smithsonian, Cambridge, MA, USA. He received the Ph.D. degree from the University of Miami, and was a Postdoctoral Research Scientist with the Lamont–Doherty Earth Observatory, Columbia University. His research spans radiative transfer modeling and satellite remote sensing of the sea surface temperature, surface radiative budget, and cryosphere. His recent work focuses on developing calibration methods, improving L0–L1 processing, and executing data processing and QA/QC for MethaneSAT and MethaneAIR.
\end{IEEEbiography}

\vskip -2\baselineskip plus -1fil
\begin{IEEEbiography}[{\includegraphics[width=1in,height=1.2in,clip,keepaspectratio]{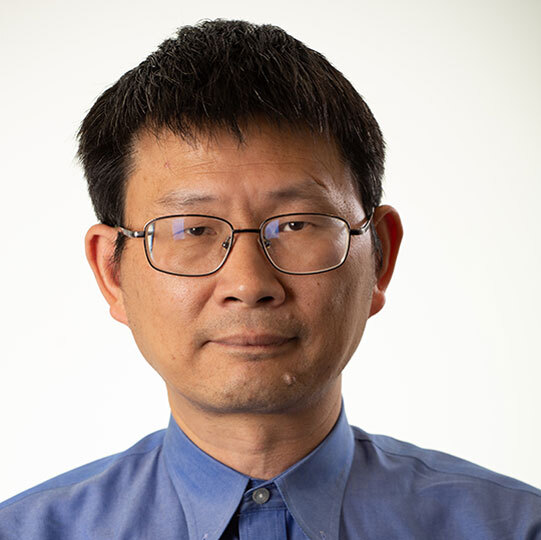}}]{Xiong Liu} received the B.S. degree in environmental chemistry from Nankai University in Tianjin, China, in 1995, and the M.S. degree in environmental chemistry from Research Center for Eco-Environmental Sciences, Chinese Academy of Sciences, Beijing, China, in 1998, and the M.S. degree in computer science and Ph.D. degree in atmospheric science from The University of Alabama in Huntsville, Huntsville, AL, USA, in 2002.
He is currently a Senior Physicist with the Center for Astrophysics $\vert$ Harvard \& Smithsonian, Cambridge, MA, USA. He is the Principal Investigator of the Tropospheric Emissions: Monitoring of Pollution (TEMPO) Project and leads the Atmospheric Measurements Group at CfA. His research interests include the remote sensing of atmospheric trace gases, aerosols, and clouds, satellite mission development, and instrument calibration.
Dr. Liu is a member of American Geophysical Union and American Meteorological Society. He received TEMPO Group Achievement Awards from NASA in 2013, 2020, and 2024, William T. Pecora Award to OMI International Science Team in 2018, AMS Special Award to OMI International Science Team in 2020, and Advances in Atmospheric Science Outstanding Editor Award in 2023 and 2024. 
\end{IEEEbiography}

\vskip -2\baselineskip plus -1fil
\begin{IEEEbiography}[{\includegraphics[width=1in,height=1.2in,clip,keepaspectratio]{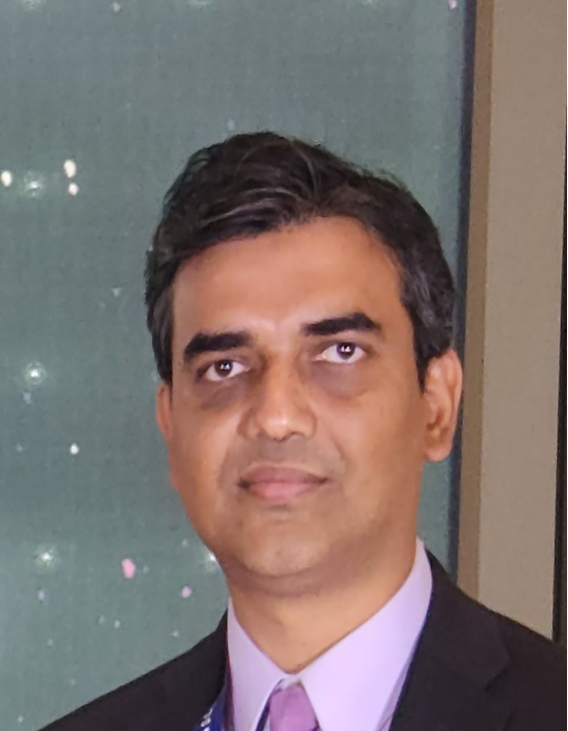}}]{Ritesh Gautam}
Dr. Ritesh Gautam is a Lead Senior Scientist on the MethaneSAT and MethaneAIR projects at Environmental Defense Fund (EDF). Since 2016, he has been overseeing remote sensing and science efforts at EDF in order to produce actionable data products and insights for quantifying and mitigating methane emissions, leveraging the broader multi-satellite ecosystem. Prior to EDF, he served as a tenured faculty member at the Indian Institute of Technology (IIT) Bombay, and previously was a Research Scientist at NASA Goddard Space Flight Center with USRA. His research interests over the last two decades have included fundamentals and applications of remote sensing of methane, aerosols, clouds and cryosphere, as well as developing improved understanding of pollution effects on air quality, climate and monsoon using Earth Observation approaches. He has published over 70 peer-reviewed scientific articles.
\end{IEEEbiography}

\vskip -2\baselineskip plus -1fil
\begin{IEEEbiography}[{\includegraphics[width=1in,height=1.25in,clip,keepaspectratio]{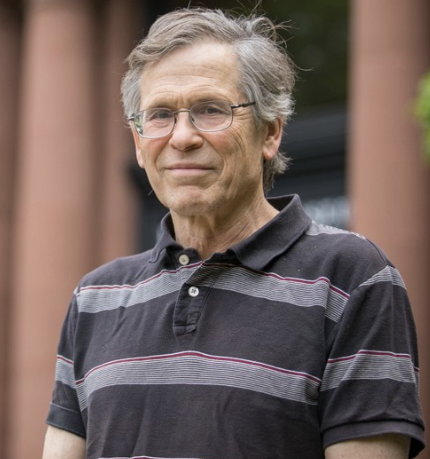}}]{Steven C. Wofsy}
(B.S., University of Chicago, 1966; Ph.D., Harvard, 1971) is the Abbott Lawrence Rotch Professor of Atmospheric and Environmental Chemistry in the John A. Paulson School of Engineering and Applied Science at Harvard University. His scientific work includes measurements and inverse modeling of fluxes of carbon dioxide and methane from ground based, aircraft, and remote sensing measurements. He is the science lead for the MethaneSAT and MethaneAIR imaging spectrometers to measure methane emissions worldwide.
\end{IEEEbiography}

\newpage 
\newpage 

\appendix

\subsection{Hyperparameter Selection} \label{appA}

For hyperparameter optimization, we employed a systematic approach using a 3-fold cross-validation strategy on a randomly selected 10\% subsample of the training data. 

For each model, we evaluated performance across learning rates $\{1 \times 10^{-4}, 5 \times 10^{-4}, 1 \times 10^{-3}, 5 \times 10^{-3}, 1 \times 10^{-2}\}$, using 3-fold cross-validation on the training set. Table \ref{tab:methane_lr} shows the chosen learning rates for each model architecture for MethaneAIR and MethaneSAT.

For all models, we employed the Adam optimizer with default $\beta_1 = 0.9$ and $\beta_2 = 0.999$ parameters. We applied class weighting to address class imbalance, with weights computed as the inverse of the class frequencies in the training set. No specific normalization was applied beyond the preprocessing steps described in Section \ref{sec:prep}.

The remaining hyperparameters were set according to the architectural specifications detailed in the methodology section. For the MLP, we used hidden layer dimensions of [20, 20]. The U-Net architecture maintained the channel dimensions specified in Section 3.2.3, while the SCAN model used a reduction ratio of 16 for the attention mechanism. The Combined MLP employed hidden dimensions of [256, 128] with a dropout rate of 0.2, and the Combined CNN used channel dimensions of [64, 32, 16] with the same dropout rate.

These hyperparameter configurations were consistently applied across all experimental evaluations to ensure fair comparison between the different architectures.

\begin{table}[h!]
\centering
\begin{tabular}{lcc}
\toprule
\textbf{Model} & \textbf{MethaneAIR} & \textbf{MethaneSAT} \\
\midrule
ILR& $1 \times 10^{-2}$ & $1 \times 10^{-2}$ \\
MLP & $1 \times 50^{-3}$ & $1 \times 10^{-2}$\\
SCAN & $1 \times 10^{-3}$ & $1 \times 10^{-3}$ \\
U-Net & $1 \times 10^{-3}$ & $5 \times 10^{-3}$ \\
Combined MLP & $1 \times 10^{-2}$ & $5 \times 10^{-4}$ \\
Combined CNN & $1 \times 10^{-2}$ & $5 \times 10^{-4}$ \\
\bottomrule
\end{tabular}
\caption{Best learning rates for each data source and  model architecture.}
\label{tab:methane_lr}
\end{table}

\begin{figure*}[t]
    \centering
    \begin{subfigure}[b]{0.49\textwidth}
        \centering
        \includegraphics[width=\textwidth]{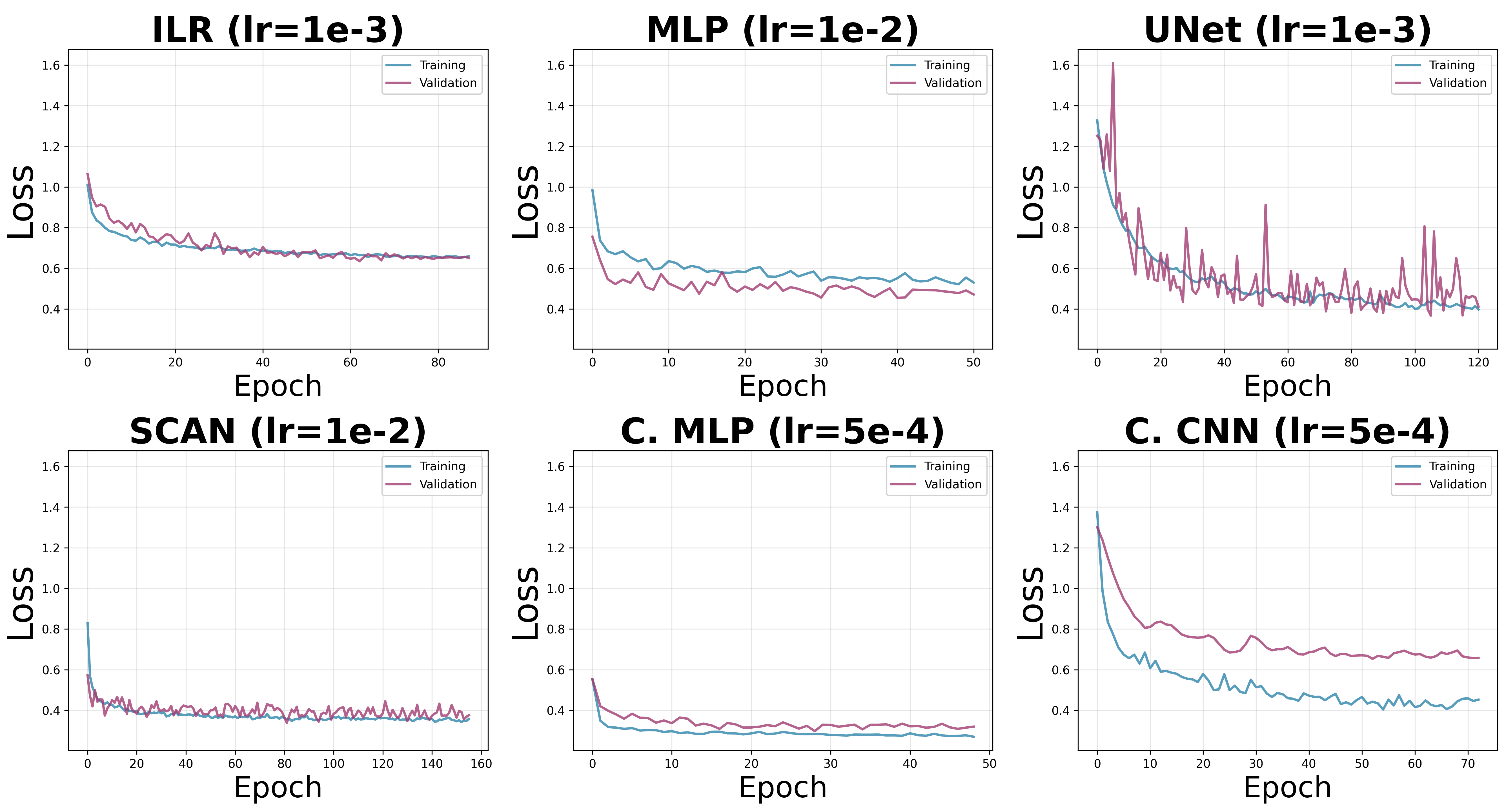}
        \caption{MethaneAIR dataset}
        \label{fig:learning_curves_mair}
    \end{subfigure}
    \hfill
    \begin{subfigure}[b]{0.49\textwidth}
        \centering
        \includegraphics[width=\textwidth]{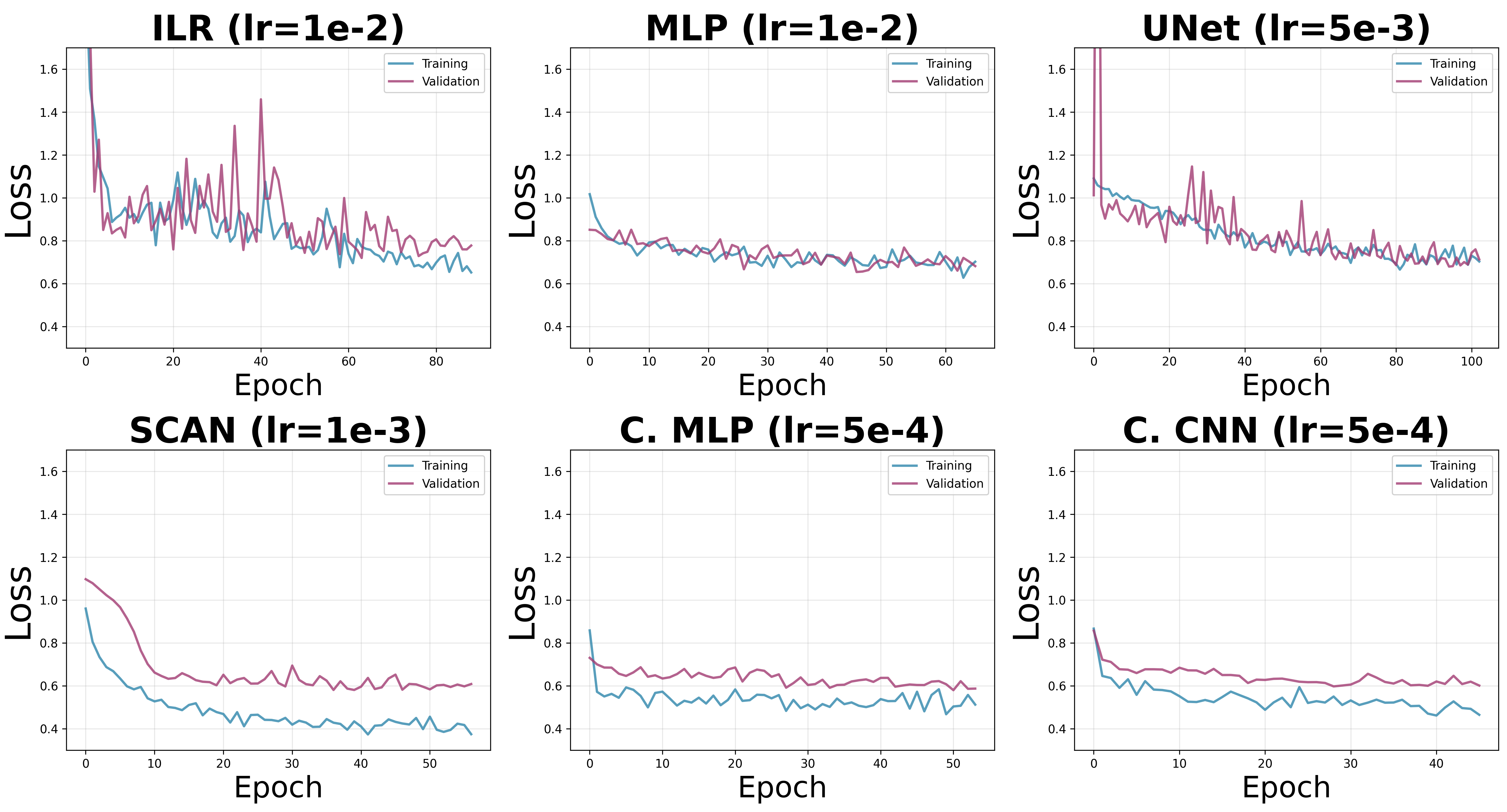}
        \caption{MethaneSAT dataset}
        \label{fig:learning_curves_msat}
    \end{subfigure}
    
    \caption{Training (blue) and validation (pink) loss evolution curves for all models on both datasets. Each subplot displays a specific architecture with its tuned learning rate. All models exhibit stable convergence with minimal overfitting, validating the robustness of the training setup.}
    \label{fig:learning_curves_all}
\end{figure*}

\mpc{Figure \ref{fig:learning_curves_all} presents the training and validation loss curves for all models evaluated in this study. Each subplot shows the loss evolution over training epochs for a different model architecture: Individual Linear Regression (ILR), Multi-Layer Perceptron (MLP), UNet, Spectral Channel Attention Network (SCAN), Combined MLP (C. MLP), and Combined CNN (C. CNN).
The training curves (blue) and validation curves (pink) demonstrate the convergence behavior of each model. All models were trained using the same experimental setup with fold 0 of the cross-validation splits. Weighted loss was applied to address class imbalance in the cloud and shadow detection tasks.}

\subsection{\mpc{Spectral Channel Attention Network Details}} \label{appB}

\subsubsection{Architecture Diagram}

\begin{figure*}[b]
\centering
\includegraphics[width=0.8\textwidth]{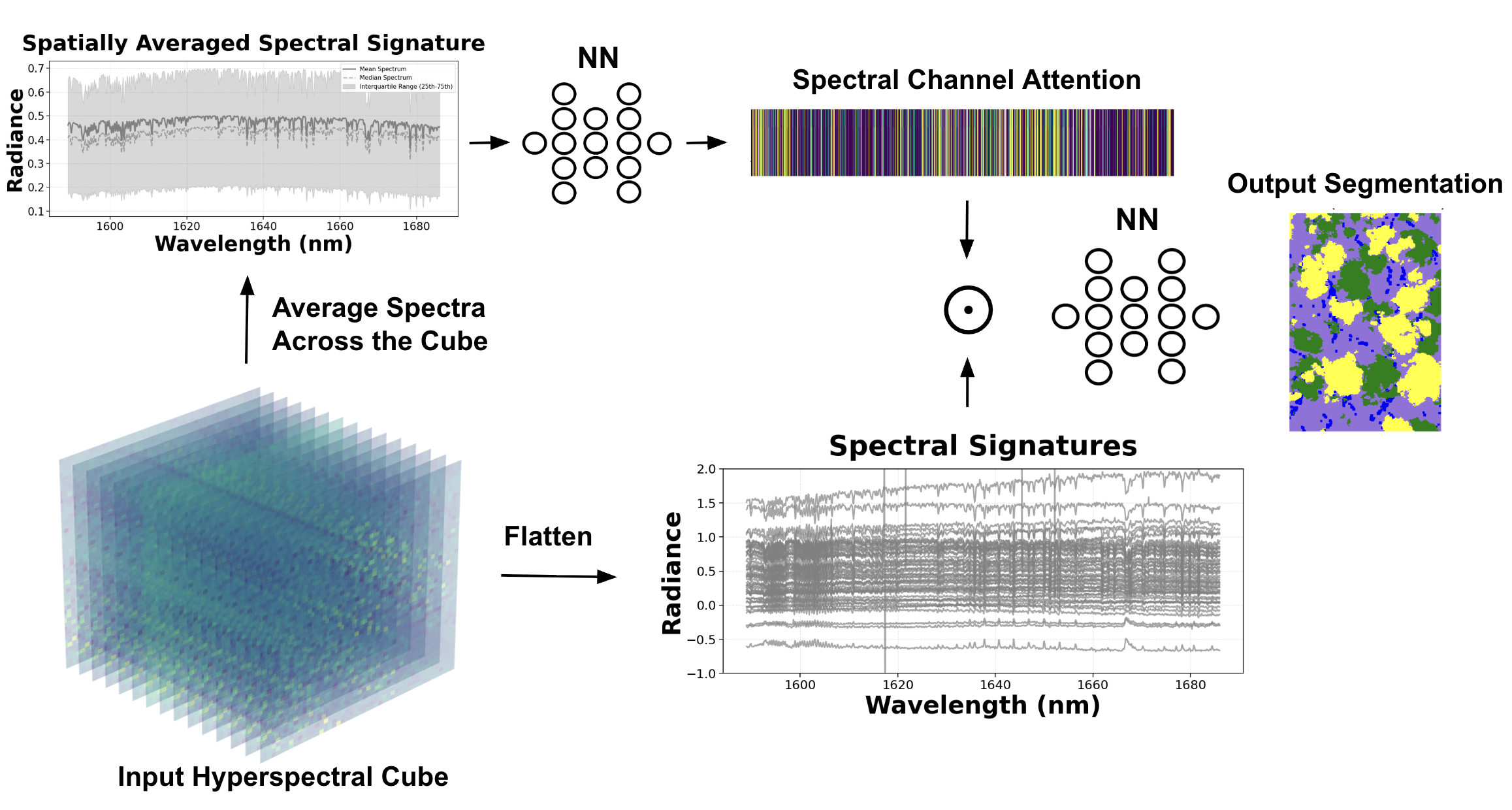}
\caption{SCAN architecture. Starting from an input hyperspectral cube, the data is flattened and processed to derive spectral signatures for each spatial sounding. The spectral channel attention module applies adaptive weighting to these signatures, amplifying informative wavelengths and reducing the influence of less discriminative spectral regions. Finally, a classification network uses these enhanced features to generate segmentation outputs for background (purple), clouds (yellow), and cloud shadows (green).}
\label{fig:scan_architecture}
\end{figure*}

\mpc{Figure~\ref{fig:scan_architecture} illustrates the Spectral Channel Attention Network (SCAN), which processes hyperspectral data through three key stages: spatial flattening to extract pixel-wise spectral signatures, adaptive channel weighting via the attention mechanism, and classification for semantic segmentation. The spectral channel attention module represents the critical innovation of this architecture, learning to emphasize wavelength regions that carry discriminative information for distinguishing clouds, shadows, and clear-sky observations while suppressing less informative spectral bands.}

\subsubsection{Comparison with Attention Mechanisms}

\mpc{To rigorously validate SCAN's design choices and demonstrate its suitability for spectral feature extraction, we conducted a systematic comparison against state-of-the-art attention mechanisms on MethaneSAT data. This comparison encompasses both channel attention approaches and transformer-based self-attention methods, providing context for understanding SCAN's performance.}

\mpc{SCAN's spectral attention module draws inspiration from established channel attention mechanisms in computer vision, particularly Squeeze-and-Excitation Networks (SE-Net) \citep{Hu_2018}, Efficient Channel Attention (ECA-Net) \citep{Wang_2020}, and Convolutional Block Attention Module (CBAM) \citep{Woo_2018}. These methods have demonstrated substantial improvements in image classification tasks by learning to emphasize informative feature channels while suppressing less useful ones.}

\mpc{Our approach shares the core architectural pattern of these methods: global average pooling followed by a bottleneck multilayer perceptron to generate channel-wise importance weights. However, there exists a fundamental distinction in application context. SE-Net, ECA-Net, and CBAM operate on spatial feature maps produced by convolutional layers, where channels represent learned abstract features. In contrast, SCAN operates directly on the spectral dimension of hyperspectral data cubes, where channels correspond to physical wavelengths with inherent spectroscopic meaning.}

\mpc{While spatial channel attention learns which learned features are discriminative, spectral channel attention learns which physical wavelengths are most informative for atmospheric phenomena. The latter problem is more constrained and interpretable, as the discriminative power of wavelengths is inspired by atmospheric physics (absorption and scattering properties) rather than learned representations. Our results demonstrate that this application of channel attention yields substantial performance gains for hyperspectral cloud and shadow detection.}

\mpc{We implemented three state-of-the-art attention-based architectures for systematic comparison on MethaneSAT data:}

\mpc{\textbf{SE-UNet}: We augmented the U-Net architecture (Section~\ref{subsubsec:unet}) with Squeeze-and-Excitation blocks \citep{Hu_2018} inserted after each convolutional stage in both encoder and decoder paths. Each SE block applies channel attention to spatial feature maps.  This implementation enables direct comparison of spatial channel attention (SE-UNet) versus spectral channel attention (SCAN).}

\mpc{\textbf{ViT-SegFormer}: We implemented a Vision Transformer \citep{Dosovitskiy_2020} backbone with SegFormer segmentation head \citep{xie_2021}. The input hyperspectral cube $\mathbf{X} \in \mathbb{R}^{H \times W \times C}$ is divided into non-overlapping patches of size $16 \times 16$ spatial soundings. Each patch is flattened (combining spatial and spectral dimensions) and linearly projected to embedding dimension 768.}

\mpc{Standard transformer layers with multi-head self-attention process the patch embeddings, learning long-range dependencies across the spatial domain. The SegFormer head upsamples the transformer output to generate pixel-wise predictions. This architecture represents the current state-of-the-art in vision transformers for dense prediction.}

\mpc{\textbf{SpectralFormer}: Following \cite{Hong_2021}, we implemented a transformer-based approach specifically designed for hyperspectral data. Rather than operating on spatial patches, SpectralFormer groups neighboring spectral bands into patches (30 bands per patch for our 1080-channel MethaneSAT data), resulting in 36 spectral patches. Each spectral patch is linearly projected to a fixed embedding dimension and processed through transformer layers with a learnable class token for pixel-wise classification.}

\mpc{This approach applies self-attention across the spectral dimension, enabling the model to learn complex inter-band relationships. A class token aggregates information across spectral patches to produce per-pixel class predictions.}

\mpc{All baseline models were trained using identical protocols describes in Section \ref{subsec:training}: weighted cross-entropy loss, Adam optimizer, same data augmentation strategies, and early stopping with patience of 20 epochs. Hyperparameters (learning rates, hidden dimensions) were optimized via 3-fold cross-validation on a 10\% validation subset to ensure fair comparison.}

\mpc{Table~\ref{tab:attention_comparison} presents comprehensive performance metrics for all attention-based methods on MethaneSAT test data. SCAN substantially outperforms all baselines across all metrics, achieving 71.53±0.75\% F1-score compared to SE-UNet (61.10±14.2\%), ViT-SegFormer (60.97±3.4\%), and SpectralFormer (62.65±1.69\%).}

\begin{table}[ht]
\centering
\caption{Performance comparison of SCAN against state-of-the-art attention mechanisms on MethaneSAT data. All methods evaluated using identical train/test splits and preprocessing. Results report mean ± standard deviation over 3 cross-validation folds. SCAN achieves substantial improvements (8.8 percentage points F1-score) over all baselines, validating the effectiveness of spectral channel attention for hyperspectral atmospheric artifact detection.}
\label{tab:attention_comparison}
\begin{tabular}{lcc}
\toprule
\textbf{Method} & \textbf{Accuracy (\%)} & \textbf{F1-Score (\%)} \\
\midrule
SE-UNet & 72.81$\pm$8.24 & 61.10$\pm$14.2 \\
ViT-SegFormer & 68.84$\pm$2.15 & 60.97$\pm$3.42 \\
SpectralFormer & 70.41$\pm$2.94 & 62.65$\pm$1.69 \\
\midrule
\textbf{SCAN} & \textbf{80.33±3.43} & \textbf{71.53±0.75}  \\
\bottomrule
\end{tabular}
\end{table}

\mpc{The superior performance of SCAN over transformer-based methods suggests that for hyperspectral remote sensing tasks with limited training data (hundreds to low thousands of scenes), simple attention mechanisms outperform complex self-attention architectures. This finding is directly applicable to missions other missions, which face similar dataset size constraints and atmospheric preprocessing challenges.}

\subsection{Appendix C: Detailed Model Predictions} \label{appC}

This appendix provides a comprehensive visual comparison of predictions all evaluated models across both the MethaneAIR and MethaneSAT datasets. These supplementary results validate and expand upon the analyses presented in Section \ref{sec:results}.

\subsubsection{MethaneAIR Predictions}

\begin{figure*}[h!]
\centering
\includegraphics[width=.9\linewidth]{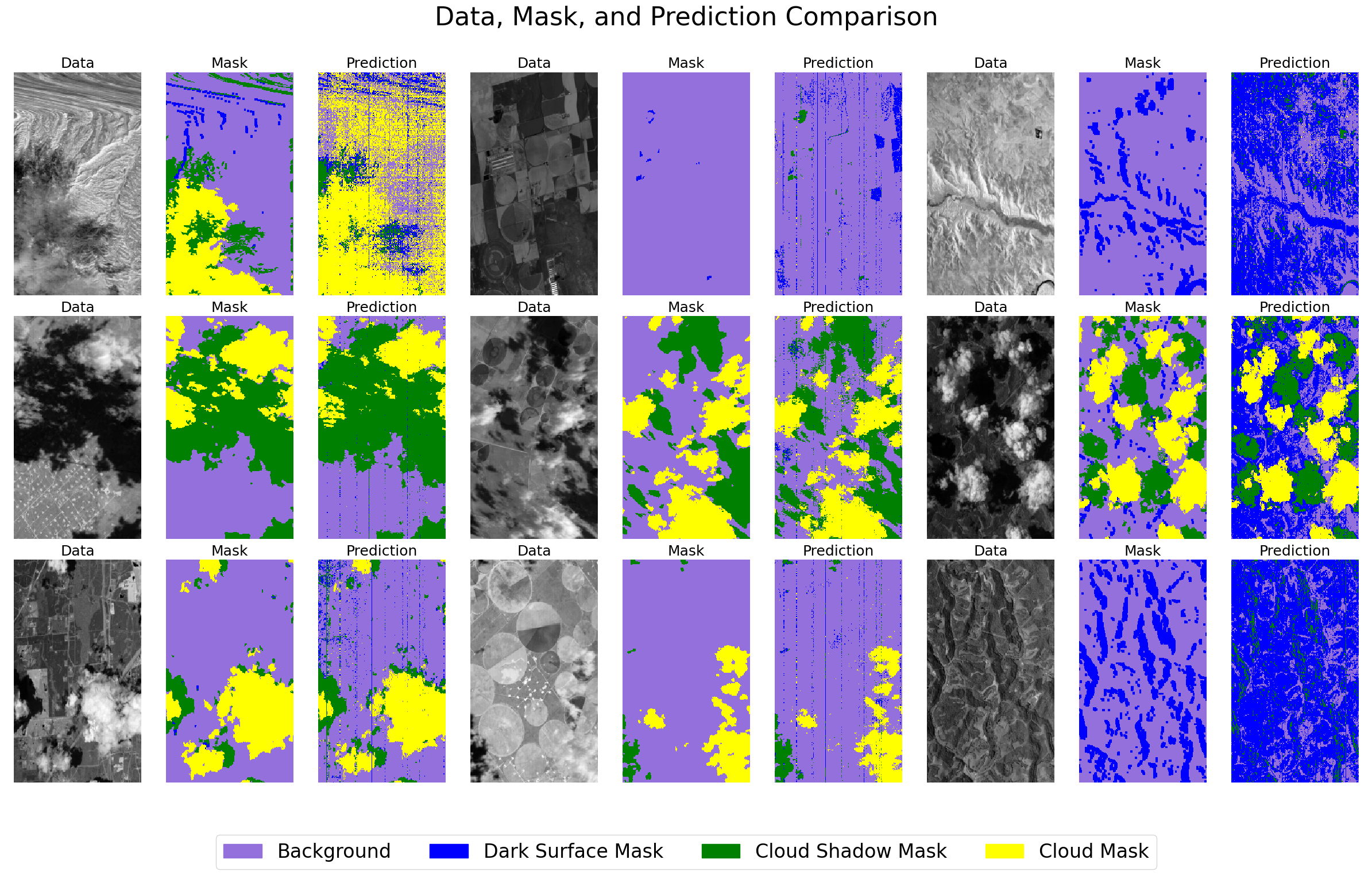}
\caption{Radiance at 1592nm, ground truth labels, and predictions of the ILR model for MethaneAIR data. The model exhibits significant noise and fragmentation, particularly in regions with complex terrain features.}
\label{fig:preds_ilr}
\end{figure*}

\begin{figure*}[h!]
\centering
\includegraphics[width=.9\linewidth]{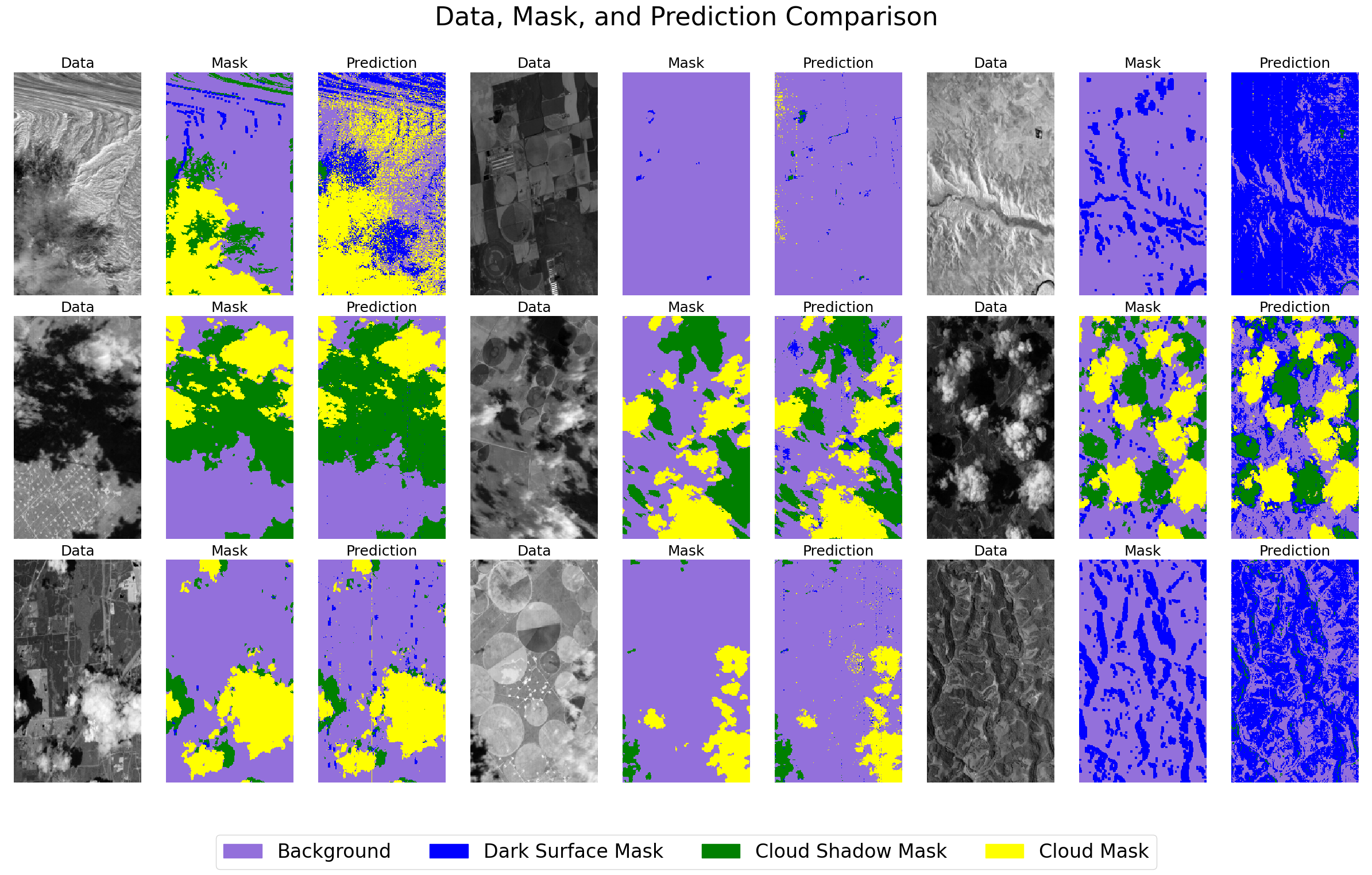}
\caption{Radiance at 1592nm, ground truth labels, and predictions of the MLP model for MethaneAIR data. While showing improved coherence compared to ILR, the model still produces speckled predictions.}
\label{fig:preds_mlp}
\end{figure*}

\begin{figure*}[h!]
\centering
\includegraphics[width=.9\linewidth]{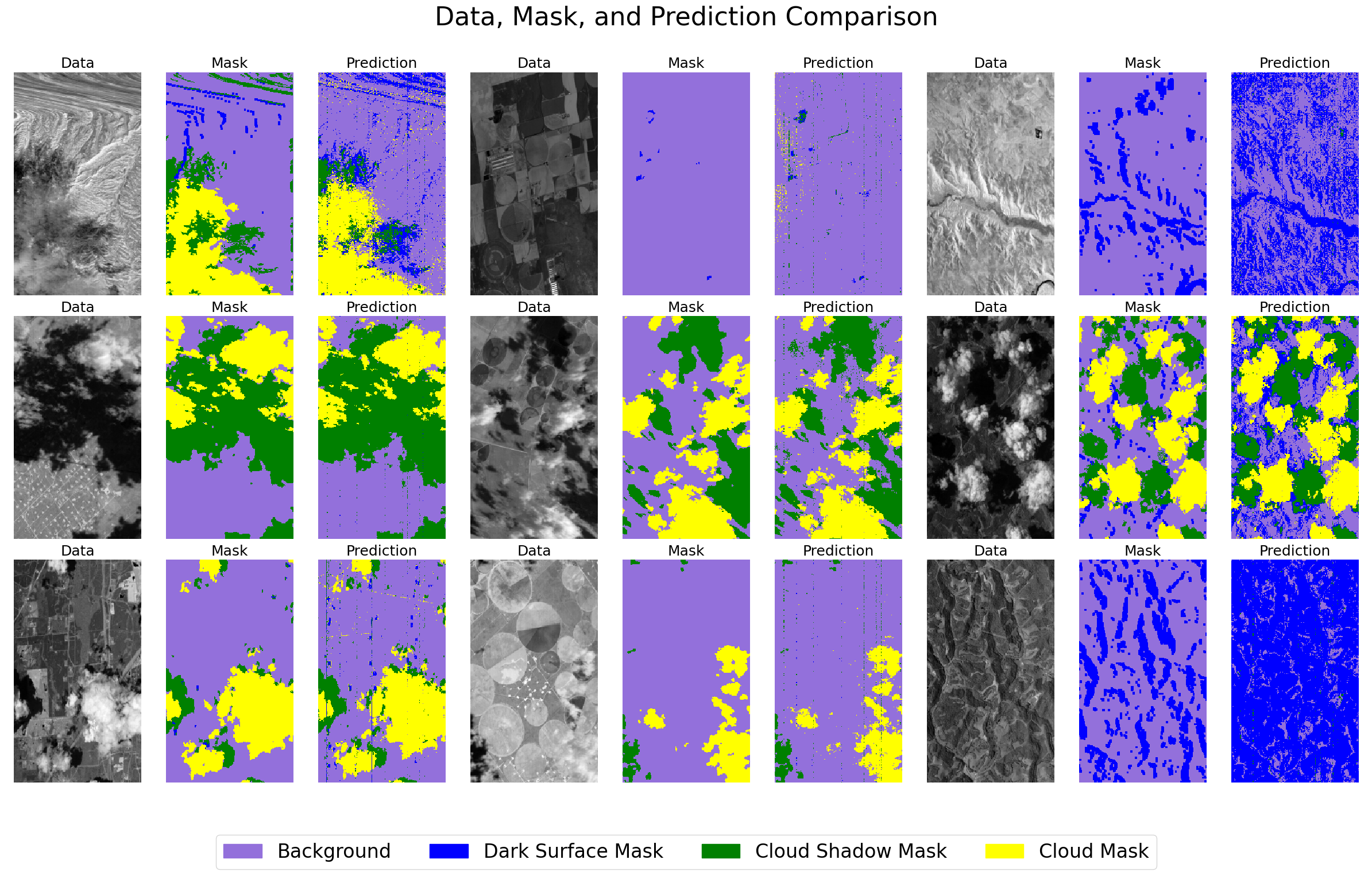}
\caption{Radiance at 1592nm, ground truth labels, and predictions of the SCAN model for MethaneAIR data. The model shows enhanced boundary detection capabilities compared to U-Net, but with some residual noise in spectrally complex regions.}
\label{fig:preds_san}
\end{figure*}

\begin{figure*}[h!]
\centering
\includegraphics[width=.9\linewidth]{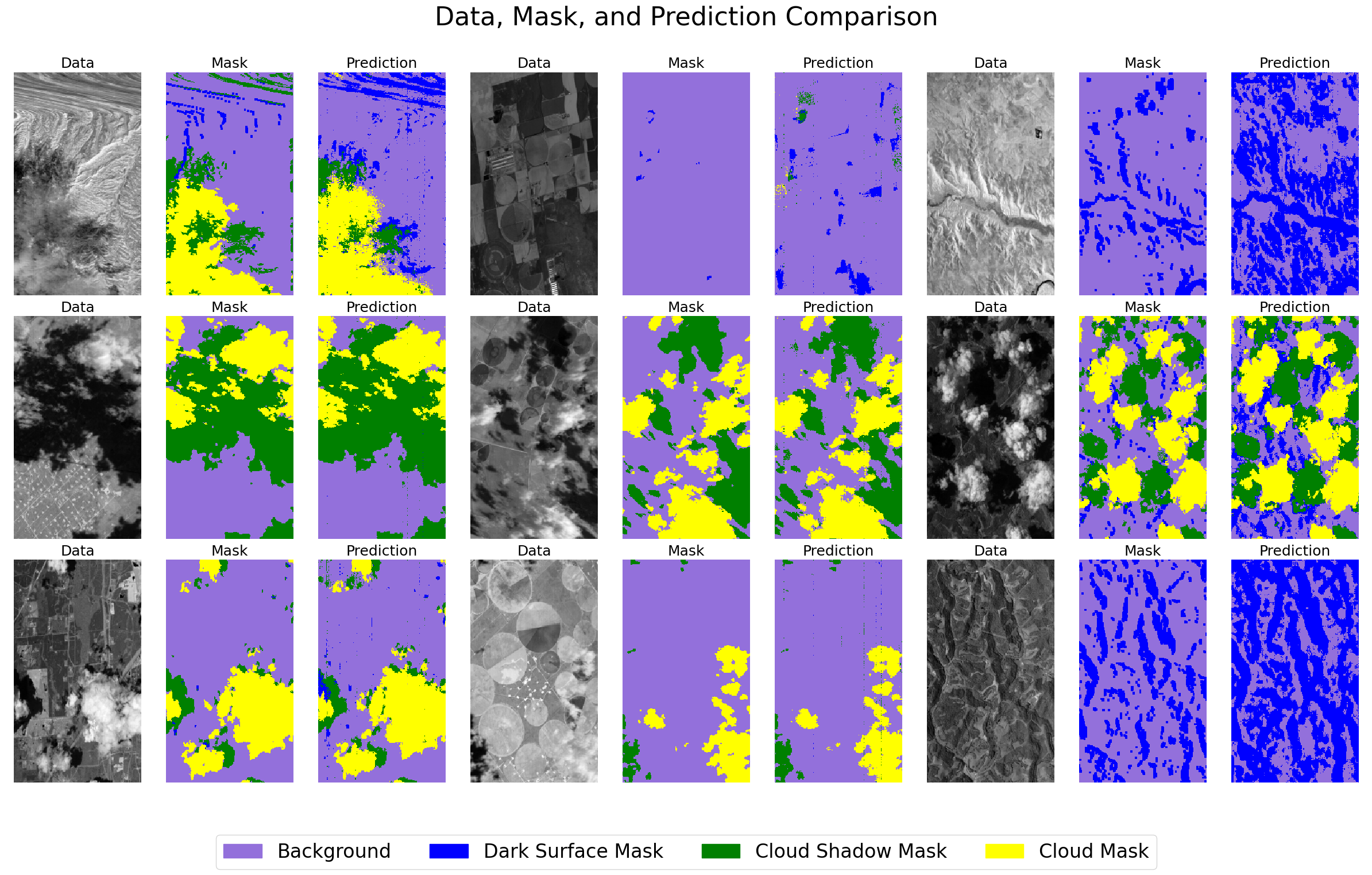}
\caption{Radiance at 1592nm, ground truth labels, and predictions of the Combined MLP model for MethaneAIR data. The fusion approach demonstrates improved performance over individual models.}
\label{fig:preds_combined_mlp}
\end{figure*}

Figures \ref{fig:preds_ilr} through \ref{fig:preds_combined_mlp} present detailed prediction results for all evaluated models on the MethaneAIR dataset. These visualizations offer insight into the specific capabilities and limitations of each approach when applied to identical input data.

The ILR predictions (Figure \ref{fig:preds_ilr}) exhibit significant fragmentation and noise, particularly evident in regions with complex terrain features. This visual noise corresponds to the high misclassification rates shown in the confusion matrix presented in the main results section, confirming the limitations of purely spectral approaches when spatial context is not considered.

The MLP model results (Figure \ref{fig:preds_mlp}) show improved coherence compared to ILR but still produce speckled predictions characteristic of pixel-wise classification approaches. While the model better captures the general distribution of clouds and shadows, it struggles with consistent classification, especially in areas with varying surface reflectance.

U-Net predictions (Figure \ref{fig:preds_unet}) demonstrate a substantial improvement in spatial coherence, with smoother, more contiguous regions for each class. However, as noted in the main results, the model's convolutional architecture tends to produce over-smoothed boundaries that do not precisely capture the intricate edges of cloud formations and their shadows.

The SCAN model results (Figure \ref{fig:preds_san}) illustrate its enhanced boundary detection capabilities compared to U-Net, with more precise delineation of cloud and shadow edges. This improved boundary precision comes with some residual noise in spectrally complex regions, representing the trade-off between spatial coherence and spectral fidelity.

The Combined MLP approach (Figure \ref{fig:preds_combined_mlp}) demonstrates the advantages of ensemble learning, with predictions that balance the strengths of individual models. By adaptively weighting the contributions of both U-Net and SCAN, this fusion approach reduces both the over-smoothing tendency of U-Net and the noise artifacts present in SCAN predictions.

Finally, the Combined CNN results (Figure \ref{fig:preds_combined_cnn}) achieve the most balanced and accurate predictions across all scenes. The model successfully integrates the precise boundary detection of SCAN with the spatial coherence of U-Net, validating the quantitative superiority demonstrated in the confusion matrices presented in the main results section.

\subsubsection{MethaneSAT Predictions}

\begin{figure*}[h]
\centering
\includegraphics[width=.85\linewidth]{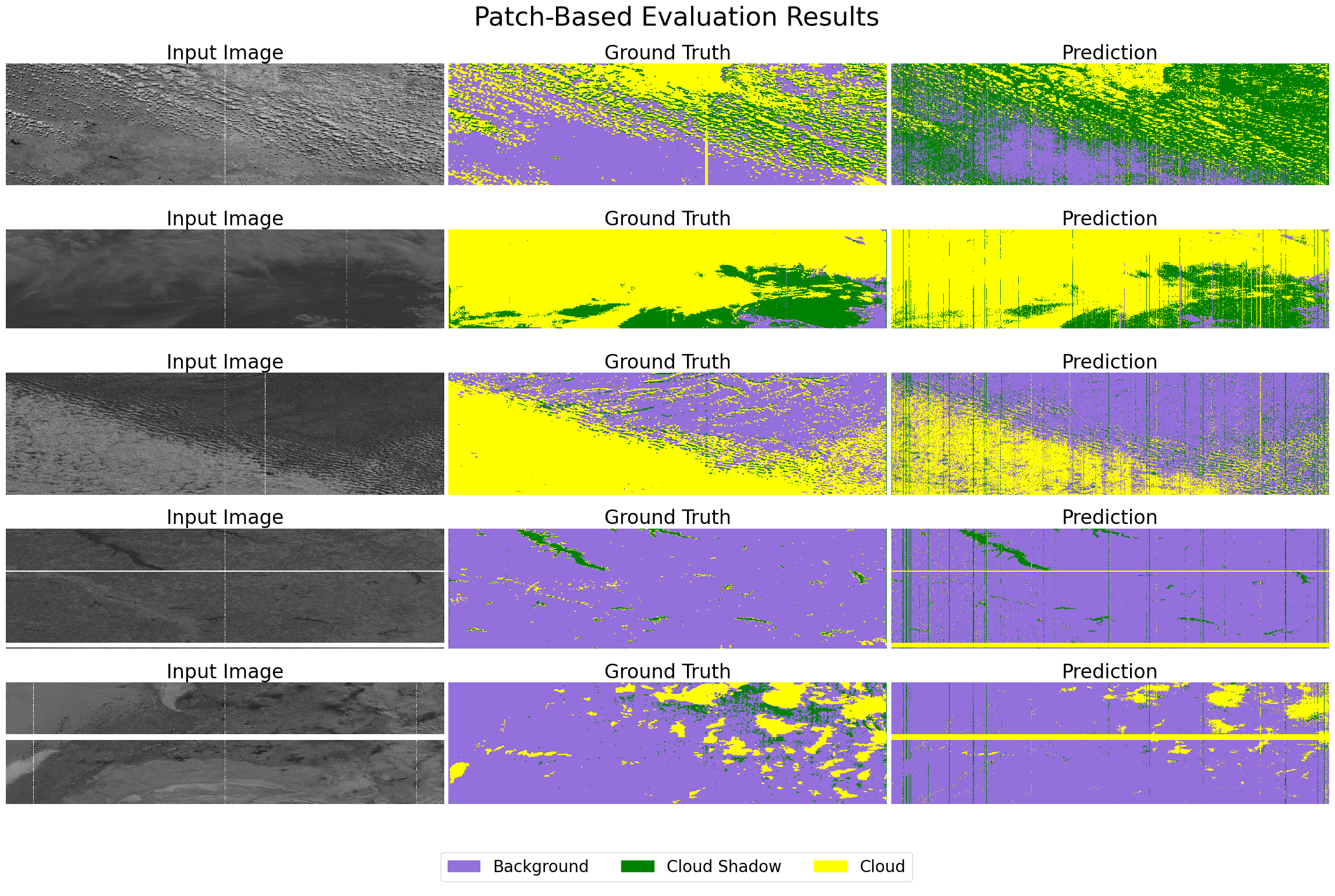}
\caption{Data, ground truth labels, and predictions of the ILR model for MethaneSAT data. The predictions exhibit significant artifacts and striping patterns characteristic of sensor-specific noise.}
\label{fig:preds_ilr_msat}
\end{figure*}

\begin{figure*}[h]
\centering
\includegraphics[width=.85\linewidth]{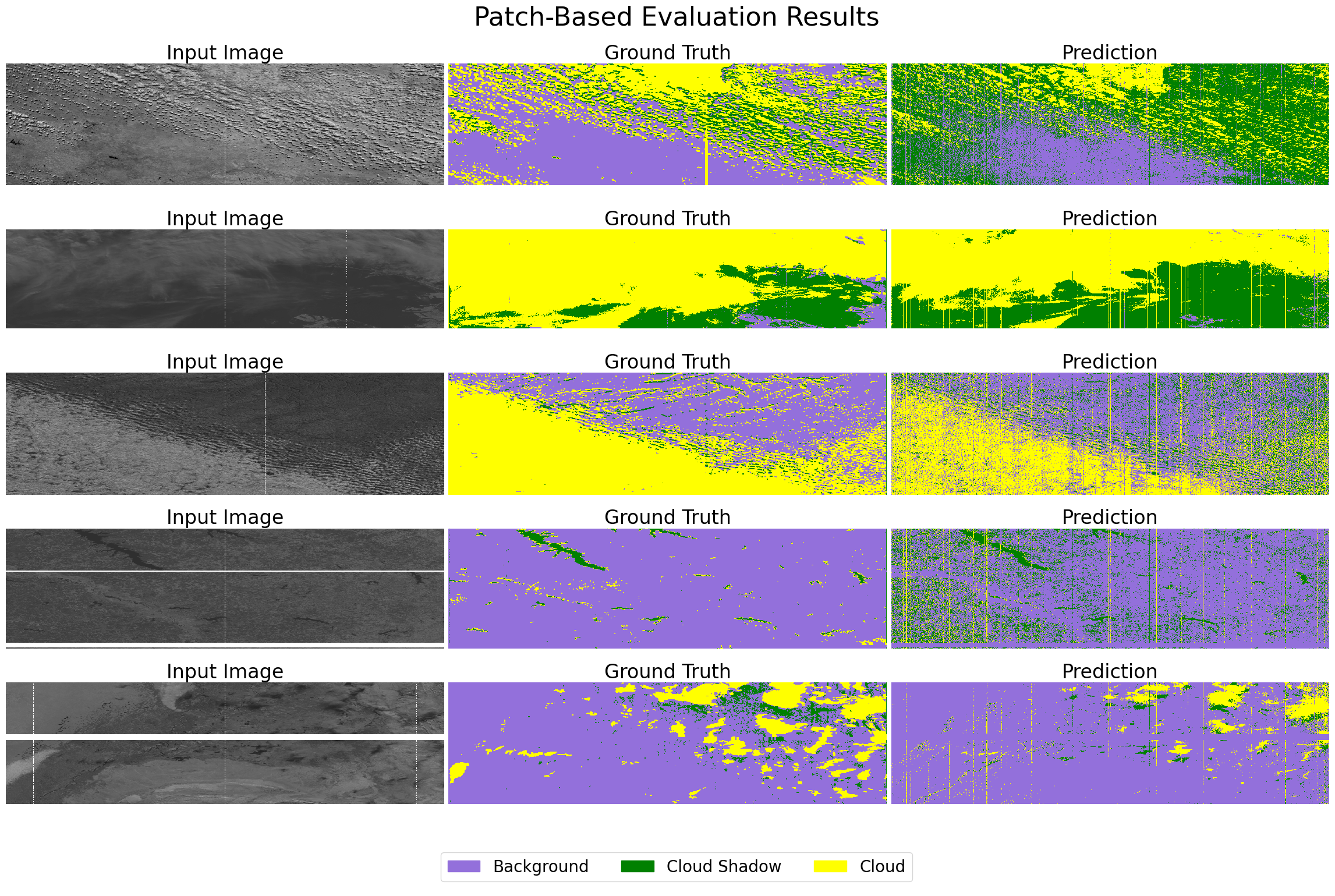}
\caption{Data, ground truth labels, and predictions of the MLP model for MethaneSAT data. The model shows improved performance over ILR but retains horizontal striping artifacts.}
\label{fig:preds_mlp_msat}
\end{figure*}

\begin{figure*}[h]
\centering
\includegraphics[width=.85\linewidth]{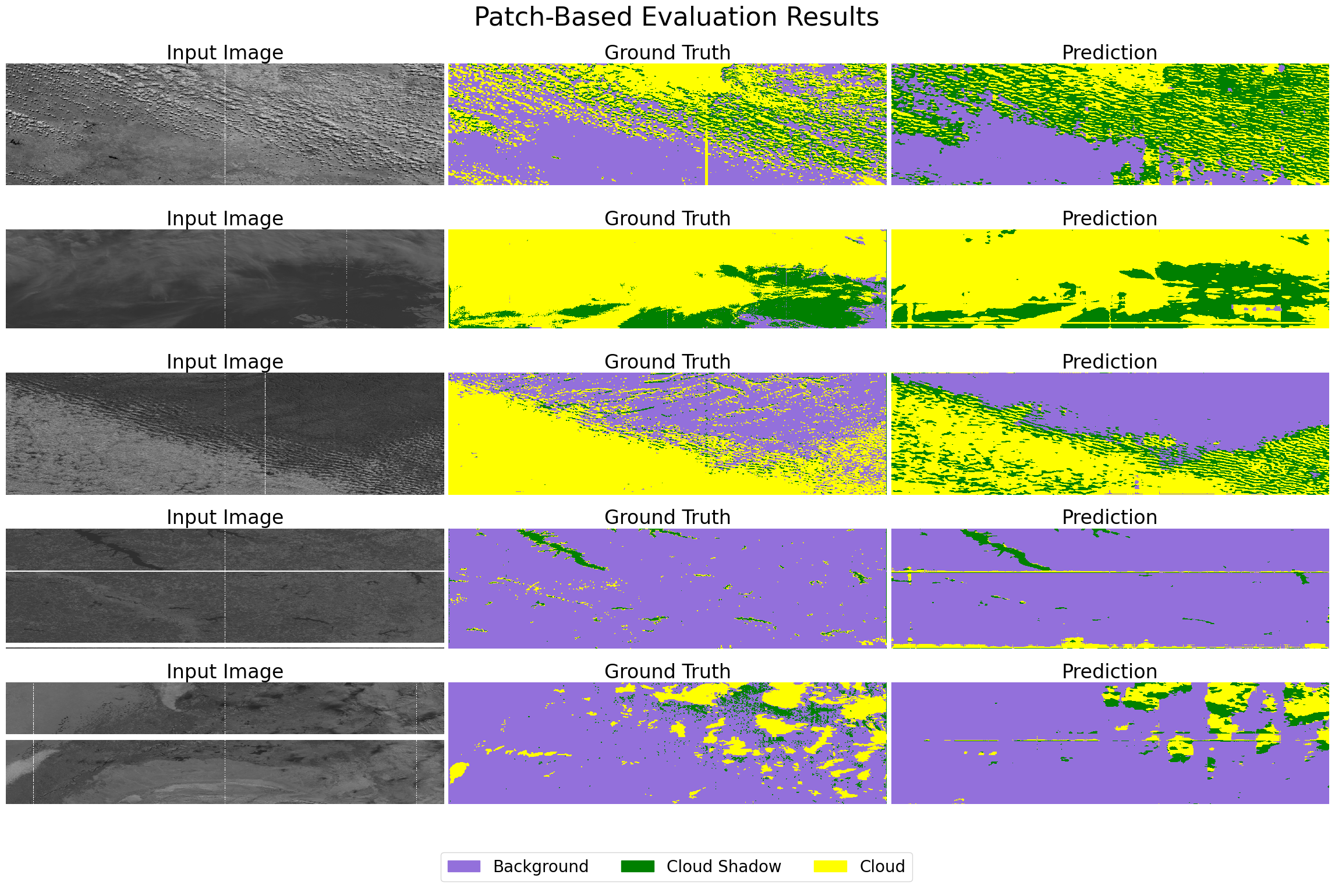}
\caption{Data, ground truth labels, and predictions of the U-Net model for MethaneSAT data. The model produces spatially coherent predictions but tends to undershoot the extent of cloud formations in certain scenes.}
\label{fig:preds_unet_msat}
\end{figure*}

\begin{figure*}[ht!]
\centering
\includegraphics[width=.85\linewidth]{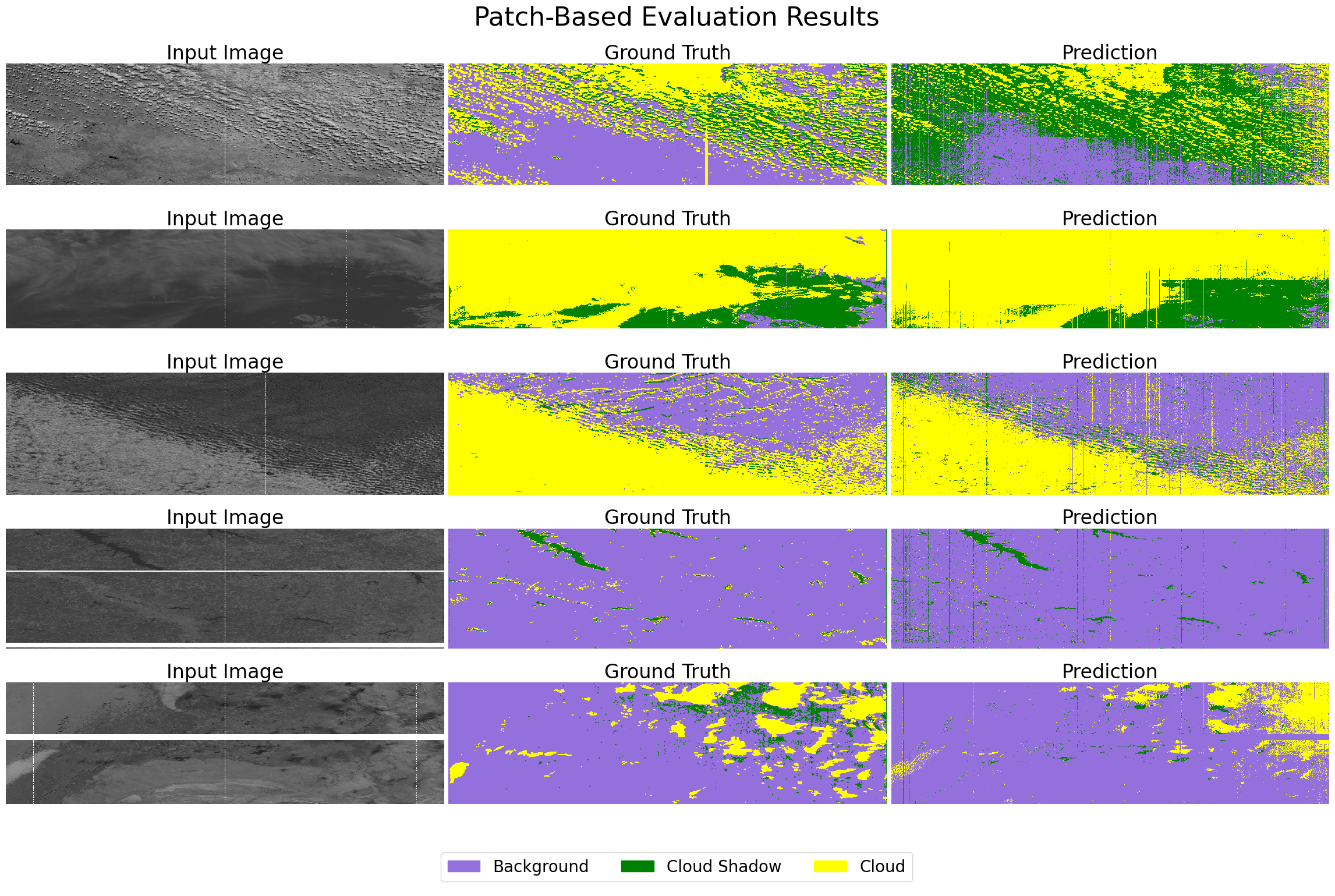}
\caption{Data, ground truth labels, and predictions of the SCAN model for MethaneSAT data. The model demonstrates improved performance compared to U-Net, particularly in capturing the boundaries of cloud and shadow regions.}
\label{fig:preds_san_msat}
\end{figure*}

\begin{figure*}[!t]
\centering
\includegraphics[width=.85\linewidth]{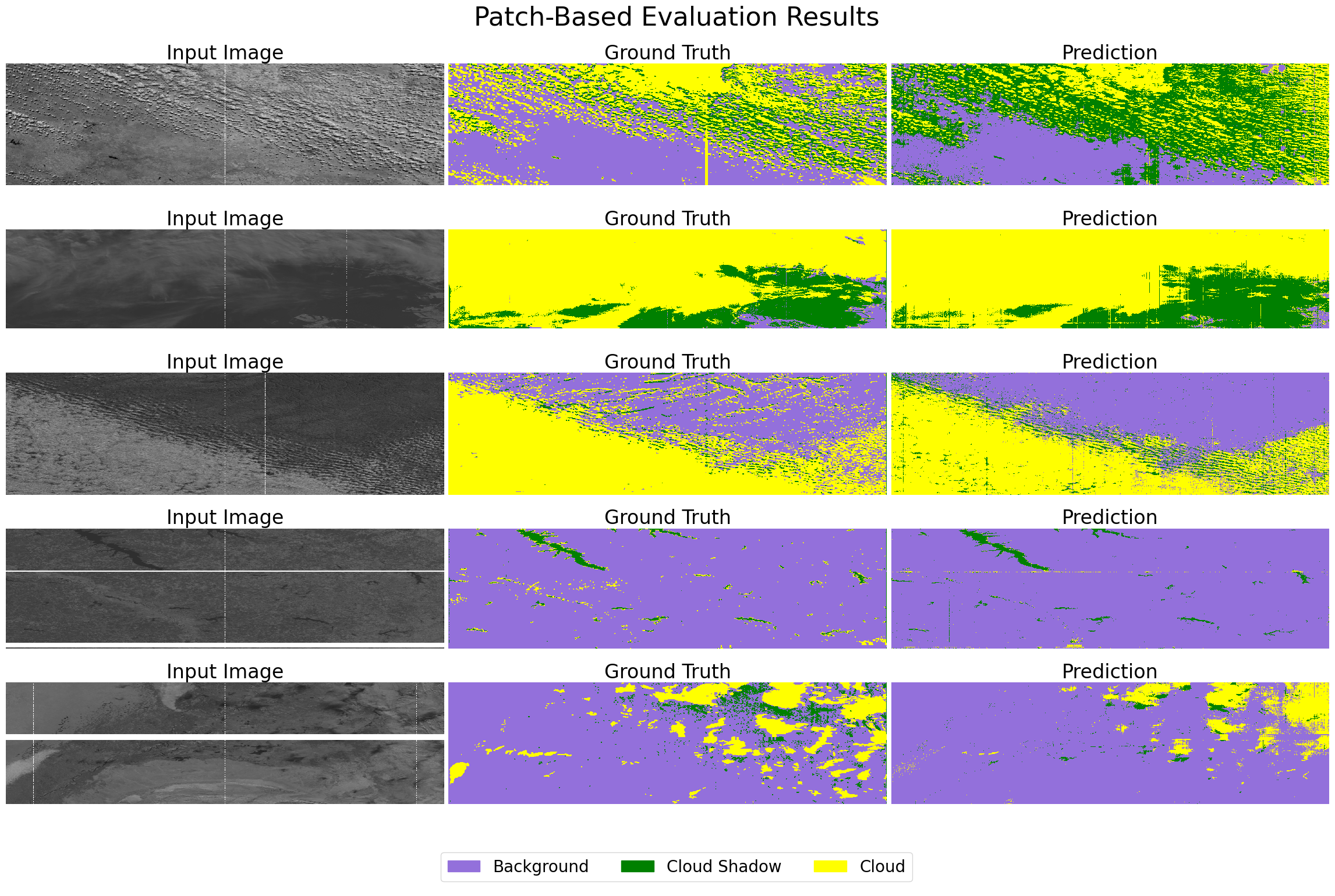}
\caption{Data, ground truth labels, and predictions of the Combined MLP model for MethaneSAT data. The model shows enhanced performance through fusion, particularly in challenging scenes with varying surface reflectance.}
\label{fig:preds_combined_mlp_msat}
\end{figure*}

Figures \ref{fig:preds_ilr_msat} through \ref{fig:preds_combined_mlp_msat} present detailed prediction results for all evaluated models on the MethaneSAT dataset. These visualizations complement the confusion matrices presented in Section \ref{sec:results} and demonstrate the performance of each model across diverse terrain and atmospheric conditions.



U-Net results (Figure \ref{fig:preds_unet_msat}) reveal significantly enhanced spatial coherence with the elimination of most noise artifacts. However, as noted in the main results, the model tends to undershoot the extent of cloud formations in certain scenes and produces overly simplified shadow boundaries, limiting its overall accuracy.

The SCAN predictions (Figure \ref{fig:preds_san_msat}) demonstrate notably improved performance compared to the U-Net model for MethaneSAT data, particularly in capturing the boundaries of cloud and shadow regions. This visual observation corresponds to the quantitative results presented in the main section, where SCAN outperformed U-Net on MethaneSAT data despite showing slightly lower performance on MethaneAIR.

The Combined MLP approach (Figure \ref{fig:preds_combined_mlp_msat}) shows enhanced performance through fusion, particularly in challenging scenes with varying surface reflectance. The model effectively reduces misclassification between spectrally similar classes while preserving the detailed structures of cloud formations.

Finally, the Combined CNN results (Figure \ref{fig:preds_combined_cnn_msat}) achieve the most accurate and balanced predictions across all MethaneSAT scenes. The model's ability to preserve spatial context during fusion enables it to effectively handle complex patterns of cloud and shadow formations while minimizing noise and misclassification, validating its superiority as demonstrated in the confusion matrices presented in the main results section.

\vfill
\end{document}